\pdfoutput=1  
\PassOptionsToPackage{table}{xcolor}

\documentclass[manuscript, screen, nonacm]{jair}   

\setcopyright{none}
\acmMonth{9}
\acmYear{2026}

\usepackage[utf8]{inputenc} 

\RequirePackage[
  datamodel=acmdatamodel,
  style=acmauthoryear,
  backend=biber,
  giveninits=true,
  uniquename=init,
  ]{biblatex}

\usepackage{makecell}
\usepackage{tabularx}
\usepackage{array}
\newcolumntype{P}[1]{>{\raggedright\arraybackslash}p{#1}}
\providecommand{\mathds}{\mathbf}
\usepackage{mathtools}
\mathtoolsset{showonlyrefs}
\DeclareMathOperator*{\argmax}{arg\,max}
\usepackage{subcaption}
\definecolor{dark-blue}{RGB}{0,0,191}
\hypersetup{colorlinks=true,citecolor=dark-blue,linkcolor=dark-blue,urlcolor=dark-blue}
\usepackage[capitalize,noabbrev]{cleveref}

\usepackage{xurl}
\usepackage[most]{tcolorbox}

\tcbset{
  chatturn/.style={
    boxrule=0.4pt,
    arc=2pt,
    left=6pt,
    right=6pt,
    top=4pt,
    bottom=4pt,
    fonttitle=\small\bfseries,
  },
  chatblue/.style={colback=blue!4!white,colframe=blue!55!black},
  chatgreen/.style={colback=green!4!white,colframe=green!45!black},
  chatgrey/.style={colback=gray!8,colframe=gray!60},
  chatgray/.style={chatgrey},
}

\NewTColorBox{chatmsg}{O{blue} m O{}}{
  chatturn,
  chat#1,
  title={\textbf{#2}},
  #3
}

\newcommand{\act}[1]{\texttt{[#1]}}

\AtBeginDocument{%
  \fancyfoot{}%
  \fancypagestyle{firstpagestyle}{\fancyhf{}}%
}

\begin{document}

\title[Learning to Sell]{Learning to Sell: Reinforcement Learning for Strategic Large Language Model Agents in Multi-Product Markets}

\author{Shuze Daniel Liu}
\email{shuzel@mit.edu}
\affiliation{%
  \institution{Massachusetts Institute of Technology}
  \city{Cambridge}
  \state{Massachusetts}
  \country{USA}
}
\affiliation{%
  \institution{Purdue University}
  \city{West Lafayette}
  \state{Indiana}
  \country{USA}
}

\author{Claire Chen}
\email{clairechen@caltech.edu}
\affiliation{%
  \institution{California Institute of Technology}
  \city{Pasadena}
  \state{California}
  \country{USA}
}

\author{Jiuqi Wang}
\email{jiuqi@email.virginia.edu}
\affiliation{%
  \institution{University of Virginia}
  \city{Charlottesville}
  \state{Virginia}
  \country{USA}
}
\author{David Simchi-Levi}
\email{dslevi@mit.edu}
\affiliation{%
  \institution{Massachusetts Institute of Technology}
  \city{Cambridge}
  \state{Massachusetts}
  \country{USA}
}
\affiliation{%
  \institution{Purdue University}
  \city{West Lafayette}
  \state{Indiana}
  \country{USA}
}
\author{Thorsten Joachims}
\email{thorsten.joachims@cornell.edu}
\affiliation{%
  \institution{Cornell University}
  \city{Ithaca}
  \state{New York}
  \country{USA}
}

\renewcommand{\shortauthors}{Liu et al.}

\begin{abstract}
Autonomous large language model (LLM) agents operating in multi-product markets must make sequential decisions under information asymmetry and resource constraints. We develop a machine learning approach for training such agents to act effectively as sellers in a multi-item bargaining environment, where a seller concurrently negotiates a catalog of substitutable assets across a pool of independent buyers. Buyers hold private, heterogeneous valuations across products, and each can purchase at most one item. Facing limits on total communication turns, the seller must dynamically match buyers with the most profitable products considering their private valuations, while strategically allocating its limited interaction budget toward combinations of greater potential value. We formalize this problem as a Partially Observable Markov Decision Process using a structured, four-part message protocol that maps natural language into a parsable and regulated decision space. Using this formalization, we design a post-training method using Reinforcement Learning from Verifiable Rewards (RLVR). To evaluate this framework, we construct a multidimensional metric suite that quantifies constraint adherence, seller surplus extraction, and allocation quality. Our trained seller agent learns to match limited inventory to buyers more effectively, matching or outperforming trillion-parameter frontier models in both seller surplus extraction and buyer-product allocation quality. Finally, these learned strategies generalize robustly to unseen market structures, correlated valuation distributions, and price ranges not encountered during training.

\end{abstract}

\maketitle

\section{Introduction}\label{sec:intro}
When autonomous language agents manage multiple concurrent interactions, they must make sequential decisions under incomplete information while deciding how to allocate limited communication opportunities across these interactions. Multi-product bargaining with private customer valuations provides a concrete setting in which to study this challenge. In many commercial settings, such as corporate procurement, customized supply chain contracting, and business-to-business sourcing, price setting does not occur through static posted schedules \citep{simchi2005logic}. Instead, transactions are finalized through sequential, multi-round bargaining interactions where counterparties systematically withhold their true reservation values to protect their individual economic surplus \citep{chatterjee1983bargaining, myerson1983efficient}. Under these asymmetric conditions, a seller cannot naively accept initial bids. To maximize total profit, the seller must explore the counterparty pool sufficiently to gather information and estimate hidden valuations \citep{fudenberg1983sequential}. 
However, this information-gathering process must be balanced against the delay and haggling costs associated with prolonged negotiation \citep{desai2004let, rubinstein1982perfect}. In a concurrent marketplace, these frictions create an additional need to allocate limited time and attention across competing negotiations.

We investigate a specific instance of this challenge in which a single seller concurrently negotiates with multiple independent buyers over a portfolio of functionally substitutable products through separate dialogue channels. Each buyer can purchase at most one product. This restriction creates a coupled matching problem: once an agreement is reached, the matched buyer and item leave the market, changing the opportunities available for the remaining buyer-item matches. A global communication budget further limits the total number of dialogue turns across the marketplace, preventing the seller from negotiating extensively with every buyer. At each turn, the seller must jointly decide which buyer to engage and what prices to offer across products to screen that buyer's preferences. To maximize total profit, the seller must use its limited communication opportunities to elicit valuation information, infer private multi-item valuations from the resulting dialogue histories, and coordinate buyer-item matching across the catalog.

Large language models (LLMs) are increasingly being evaluated as autonomous strategic agents in multi-agent commercial interactions \citep{fu2023improving, davidson2024evaluating}. However, recent benchmarks have identified systematic weaknesses in the negotiation and strategic decision-making capabilities of general-purpose language models \citep{bianchi2024well, xia2024measuring}. Such weaknesses can become particularly consequential in our setting, where locally attractive agreements may conflict with the long-horizon objective of maximizing portfolio-wide profit. For example, accepting an early offer that is profitable in isolation can remove both the buyer and the item from future consideration. This early agreement may therefore foreclose the possibility of a more profitable match, as that buyer might have held a significantly higher valuation for a different product in the catalog, ultimately leading to an inefficient allocation of inventory and lower overall profit.

The paper addresses these limitations by introducing a framework that aligns natural language interaction with multi-product portfolio surplus optimization. We make three primary contributions to autonomous AI agents and multi-agent sequential decision-making:

\begin{enumerate}
\item \textbf{A Structural Formalization for Concurrent Bargaining over Substitutable Product Portfolios:} We formalize the dynamic allocation of the substitutable product portfolios across multiple independent buyers as a finite-horizon Partially Observable Markov Decision Process (POMDP). To bridge natural language generation with verifiable sequential decision-making, we introduce a regularized, multi-part protocol (\texttt{Thought}, \texttt{Target}, \texttt{Talk}, \texttt{Action}). This structural protocol maps high-dimensional text dynamics into a parsable, rule-bounded decision space, providing a systematic framework for controlling agent behavior, enforcing hard action constraints, and coordinating decisions under a limited interaction budget.

\item \textbf{An End-to-End Post-Training Method and Emergent Strategic Behaviors:} We develop a post-training method using Reinforcement Learning from Verifiable Rewards (RLVR) to optimize the seller model against strategic counterparties. Our optimization loop uses a terminal, verifiable economic reward reflecting portfolio-wide surplus extraction, without explicit behavioral heuristics or predefined matching logic. We show that buyer-item matching behaviors and allocation precision improve as the agent maximizes surplus. As these mechanics are never explicitly rewarded, their emergence demonstrates that verifiable economic rewards can induce coordinated information-gathering and allocation behaviors. More broadly, negotiation is one instance of a broader class of partially observable sequential decision problems requiring language agents to gather information, allocate limited resources, and coordinate actions toward persistent objectives. Our results provide evidence that end-to-end training with verifiable rewards can induce the strategic behaviors needed to solve such problems without hand-crafted decision rules or behavioral supervision.

\item \textbf{A Multidimensional Evaluation Suite:} We introduce a multidimensional metric suite designed to evaluate bargaining and routing efficacy beyond raw deal volume or seller surplus extraction. This framework systematically quantifies distinct operational dimensions: protocol constraint adherence (\textit{Instruction Violation Rate}), margin optimization efficiency (\textit{Seller Surplus Extraction Ratio}), and matching precision (\textit{Deal on Top Buyer Rate} and \textit{Optimal-Pair Offer Coverage Rate}). We validate this framework by constructing an empirical evaluation dataset comprised of $6{,}500$ functionally substitutable products drawn from real-world marketplace listings, using it to benchmark our trained agent against a diverse panel of leading frontier models.

\end{enumerate}

The remainder of this paper is organized as follows. \Cref{sec:related} reviews the relevant literature across large language models as autonomous strategic agents, strategic negotiation, and sequential allocation under resource constraints. \Cref{sec:methodology} formalizes the concurrent multi-product bargaining problem as a finite-horizon Partially Observable Markov Decision Process and specifies the verifiable reward formulation. \Cref{sec:dataset} details the curation and characteristics of our empirical substitutable product dataset. \Cref{sec:experimental_setup} outlines the simulation environment, marketplace parameters, and agent configurations. \Cref{sec:evaluation} introduces our multidimensional evaluation framework and metrics. \Cref{sec:benchmark} presents our main in-distribution benchmarking results against a diverse panel of leading frontier models. \Cref{sec:training_dynamics} analyzes the empirical learning trajectories and emergent strategic behaviors of the learning seller agent. \Cref{sec:ood} subjects the trained seller agent to diverse out-of-distribution stress testing across scale expansions, structural imbalances, correlated valuation structures, and unseen price distributions. Finally, \Cref{sec:discussion} examines the broader implications of our findings for autonomous language agents, and \Cref{sec:conclusion} concludes the paper.

\section{Related Work}
\label{sec:related}

This work builds on three strands of literature: large language models as autonomous strategic agents, sequential bargaining and information elicitation under asymmetric information, and sequential allocation and matching under resource constraints. We connect these strands by studying a concurrent marketplace in which buyer valuation screening through natural-language interaction must be coordinated with buyer-item matching under a global communication budget.

\subsection{Large Language Models as Autonomous Strategic Agents}
\label{subsec:lit_llm_agents}

Autonomous learning agents have traditionally relied on tabular or deep reinforcement learning operating over structured state-action spaces 
\citep{gijsbrechts2022can, boute2022deep}. 
These decision-theoretic frameworks typically operate on structured observations and actions rather than the open-ended natural-language interactions that characterize many strategic settings, including negotiation, persuasion, and commercial exchange \citep{desai2004let, backus2020sequential}. To capture these high-dimensional text dynamics directly, recent literature has evaluated Large Language Models (LLMs) as autonomous decision-makers in interactive environments \citep{yao2022react, liu2024agentbench, wang2024survey, zeng2024agenttuning}.
However, systematic benchmarks demonstrate that general-purpose language agents often exhibit shortsightedness; standard instruction-tuned or reasoning models frequently prioritize linguistic agreeableness over economic optimality, and struggle to navigate the tension between concession and value extraction under counterparty pressure \citep{fu2023improving, chawla2023selfish, xia2024measuring, davidson2024evaluating, abdelnabi2024cooperation}.

One possible contributor to these behaviors is the objective mismatch between general-purpose post-training and task-specific strategic optimization. Standard training methods like Reinforcement Learning from Human Feedback (RLHF) typically optimize for broadly desirable conversational behavior, such as helpfulness and cooperation, rather than for a specific long-horizon strategic objective \citep{ziegler2019fine, ouyang2022training}. Prior work has also documented overly agreeable behavior in language models \citep{perez2023discovering, shah2026too}. Reinforcement Learning from Verifiable Rewards (RLVR) provides an alternative way to align model behavior with task-specific objectives. By optimizing against programmatically verifiable outcomes rather than subjective human labels, RLVR and closely related verifier-based methods have substantially improved mathematical reasoning \citep{cobbe2021training, wang2024math, shao2024deepseekmath, guo2025deepseek} and, in some work, code generation \citep{ma2023let, guo2025deepseek}. We use verifiable-reward post-training to study a distinct agent-learning problem: training an end-to-end strategic agent to coordinate information gathering, negotiation, and allocation decisions toward a unified economic objective of maximizing seller surplus. Unlike modular negotiation systems that decouple strategy from language generation \citep{he2018decoupling, yarats2018hierarchical}, our pipeline unifies persuasive communication and concurrent multi-buyer market search into a single optimization loop, training the agent to actively discover hidden valuation boundaries without access to market priors.

\subsection{Negotiation and Price Discovery under Asymmetric Information}
\label{subsec:lit_negotiation_asymmetric}

Negotiation provides a natural setting in which strategic agents can elicit private information about counterparties, such as hidden costs or valuations.  Classical work on bargaining, pricing, and supply-chain contracting studies how structured pricing menus, quantity discounts, and contractual mechanisms can manage these informational frictions \citep{cachon2001contracting, leider2016bargaining, liu2021optimal}. Related AI research examines bilateral multi-issue negotiation under deadlines
\citep{fatima2006multi} and resource reallocation through sequences of negotiated
deals \citep{dunne2005extremal}.
However, when the allocation process is governed by a finite interaction budget, bargaining choices directly impact revenue management. Under these operational capacity constraints, a seller must strategically manage sequential offers across channels to protect profit margins and optimize surplus extraction when interacting with strategic counterparties. This dynamic requires integrating sequential bargaining directly into resource allocation, emphasizing the need to actively elicit buyer valuations through interaction to maximize revenue \citep{bhandari2011revenue}.

Classical models typically formalize multi-turn negotiation as an information-elicitation process where players infer private valuations through sequential pricing choices under incomplete information \citep{harsanyi1967games, chatterjee1983bargaining}. Beyond rigid numerical strategies, empirical and experimental research highlights that real-world bargainers often rely on unstructured communication and behavioral heuristics to discover surplus boundaries and gather market data \citep{srivastava2000price, guo2023gathering}. While emerging learning-based frameworks leverage large language models to automate these dialogue-driven interactions \citep{bergemann2026training, liu2026instructing, liu2026strategic}, they restrict their focus to isolated, bilateral settings. Our work expands this literature by evaluating a concurrent marketplace governed by a communication limit. Under this formulation, the seller must simultaneously execute conversational valuation screening and buyer-item matching across multiple private channels, maximizing overall seller surplus without prior knowledge of the buyers' hidden valuations.

\subsection{Sequential Allocation and Matching under Resource Constraints}
\label{subsec:lit_inventory_matching}
Sequential allocation and matching problems study how limited, heterogeneous resources should be assigned across multiple decision opportunities under capacity constraints. Classical work in online matching and revenue management establishes algorithms, competitive guarantees, and approximation frameworks for making such allocation decisions under uncertainty \citep{gallego1994optimal, talluri2006theory, karp1990optimal, mehta2007adwords, manshadi2012online}. These approaches typically rely on centralized optimization frameworks—such as dynamic programming decompositions or multi-commodity linear programs—and operate on structured numerical signals, often assuming access to demand distributions or market priors \citep{gallego1997multiproduct, bitran2003overview, jaillet2014online}.

When customer preferences are unknown a priori, the allocation problem introduces an exploration-exploitation tradeoff under hard capacity limits. This tension has been addressed through online learning and adaptive allocation methods that learn demand patterns, balance inventory, and manage structural market imbalances across channels \citep{ferreira2018online, ma2020algorithms, cheung2022inventory, gong2022online, chen2024assortment, feng2025batching, simchi2025greedy, liu2026or}. While these models provide strong performance guarantees, they typically represent customer preferences, requests, and allocation opportunities through structured numerical signals. In contrast, our framework models a concurrent marketplace where a single agent must coordinate matching decisions across parallel communication channels, while preferences are discovered through multi-turn natural-language dialogue.

\section{Problem Formulation and POMDP Framework}
\label{sec:methodology}

\begin{figure}[t]
\centering
\includegraphics[width=\textwidth]{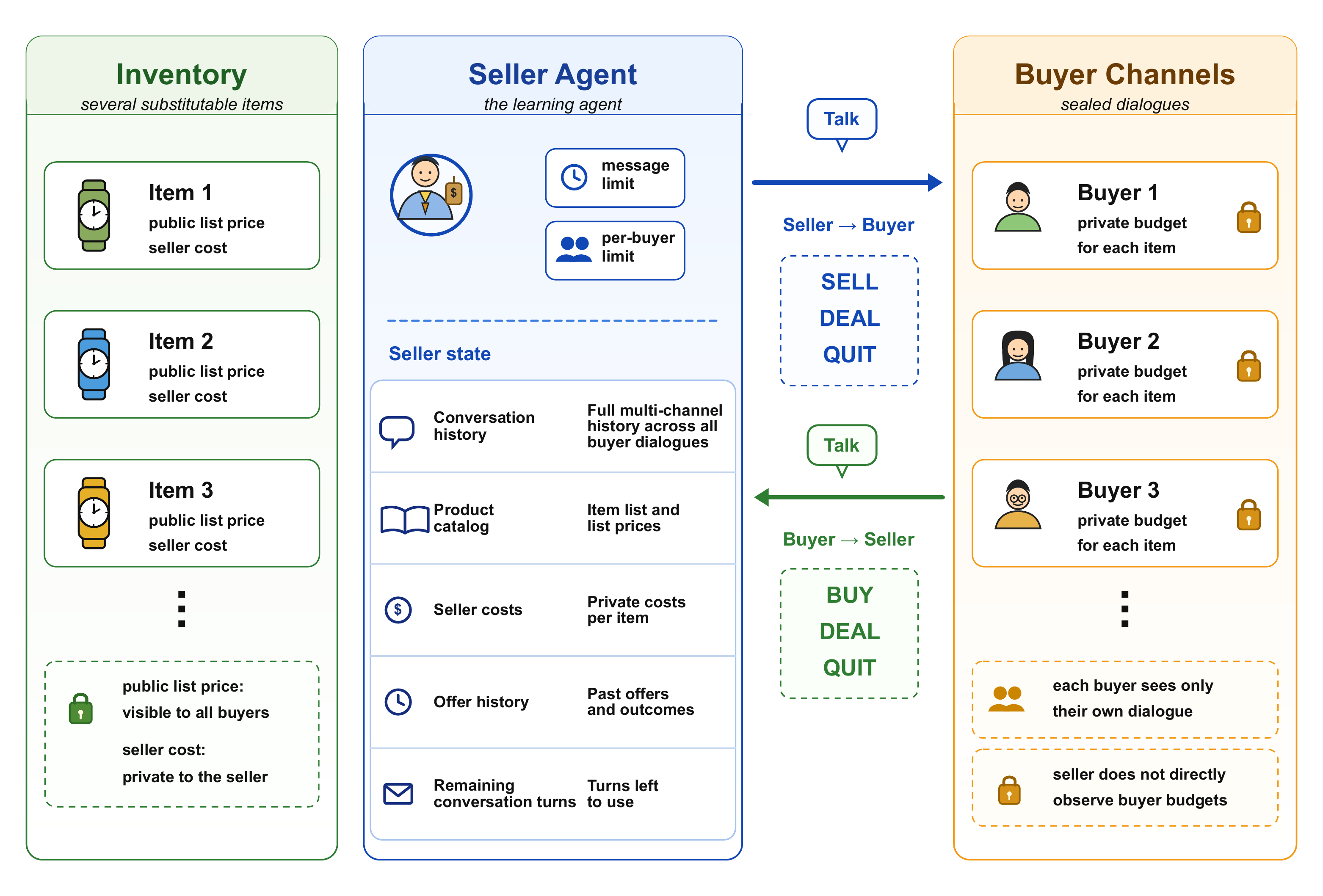}
\caption{
Structural design of the bargaining environment for substitutable product portfolios. The seller agent manages inventory across independent, sealed buyer channels subject to global and local communication constraints.
}
\label{fig:market_framework}
\Description{A schematic of the market: one seller at the centre holds a catalog of substitutable items and private per-item costs, connected by separate sealed channels to several buyers, each holding private per-item valuations. Labels on the channels mark the per-buyer turn limit $K$, and a label on the seller marks the global budget $T$, with $T$ smaller than the number of buyers times $K$.}
\end{figure}

We formalize the multi-product seller negotiation problem as a sequential decision process where a single seller (the learning agent) is responsible for allocating a finite catalog of $n$ substitutable items across $m$ heterogeneous buyers. The $n$ items are functionally substitutable variants drawn from a single product type, and each buyer requires at most one item from the catalog. The seller holds a private per-item cost vector $\{\mathcal{C}_j\}_{j=1}^{n}$ and is matched against $m$ independent buyers indexed by $i \in \{1, \ldots, m\}$. Each buyer $i$ possesses a private per-item valuation vector $\{\mathcal{B}_{i,j}\}_{j=1}^{n}$, forming a hidden valuation (budget) matrix $\mathcal{B} \in \mathbb{R}^{m \times n}$ generated according to the environment-specific valuation process (parameterized in Section~\ref{subsec:market_settings}; see Appendix~\ref{subsec:app_additional_stats} and Appendix~\ref{subsec:ood_setup} for full details).

All buyers can view the public catalog parameters—including product descriptions, metadata, and per-item list prices $\{P_{\text{list},j}\}_{j=1}^{n}$—but the seller's private costs remain hidden. Besides, the communication channels are partitioned: buyer $i$ only observes messages directed to them and has no visibility into the concurrent dialogues or private budgets of the other $m-1$ participants. Furthermore, the seller must strategically allocate a finite global communication budget of $T$ total dialogue turns across the $m$ available buyer slots, subject to a per-buyer turn limit $K$, where $T < mK$. Consequently, the seller cannot afford to fully exhaust every conversation and must dynamically choose which buyers to engage. The items for each episode are drawn from a dataset of substitutable products (see \Cref{sec:dataset} for details about the curated dataset).

\subsection{Multi-Item Seller Negotiation as a POMDP}

We formalize the multi-product seller negotiation problem from the seller's perspective as a finite-horizon Partially Observable Markov Decision Process (POMDP) defined by the tuple $(\mathcal{X}, \mathcal{A}, \mathcal{P}, \mathcal{R}, \mathcal{O}, \Omega, T)$, where $\mathcal{X}$ is the environment state space, $\mathcal{A}$ is the action space, $\mathcal{P}$ is the state transition probability function, $\mathcal{R}$ is the reward function, $\mathcal{O}$ is the observation space, $\Omega$ is the observation mapping, and $T$ is the finite game horizon. 

Concretely, the environment state at turn $t$ is
\[
  x_t \;=\; (\mathcal{B},\, \Phi,\, \mathcal{C},\, \mathcal{H}_t,\, \mathcal{N}_t,\, T_t) \;\in\; \mathcal{X},
\]
comprising the hidden buyer valuation matrix $\mathcal{B}$, the public catalog parameters $\Phi$, the seller's private cost vector $\mathcal{C}$, the dialogue history $\mathcal{H}_t$, the structured negotiation record $\mathcal{N}_t$, and the remaining global communication budget $T_t$. The last three are taken at their values immediately before the seller chooses its action at turn $t$. The episode-level components $\mathcal{B}$, $\Phi$, and $\mathcal{C}$ are fixed throughout an episode, whereas $\mathcal{H}_t$, $\mathcal{N}_t$, and $T_t$ evolve after each seller--buyer exchange. The seller's observation is the seller-visible projection of this state,
\[
  o_t \;=\; \Omega(x_t) \;=\; (\Phi,\, \mathcal{C},\, \mathcal{H}_t,\, \mathcal{N}_t,\, T_t) \;\in\; \mathcal{O},
\]
so that $\mathcal{O}$ is the space of seller-visible configurations consisting of the public catalog, the seller's own cost information, the accumulated dialogue history, the structured negotiation record, and the remaining communication budget. The observation mapping $\Omega$ is the projection that withholds the valuation matrix: $\mathcal{B} \notin o_t$, and it remains hidden for the entire episode. Because $\mathcal{H}_t$ and $\mathcal{N}_t$ already accumulate the full interaction record, $o_t$ is itself the seller's complete information state, and the seller chooses its action directly from $o_t$ \citep{sutton2018reinforcement}. We index the seller's turns by $t \in \{1, 2, \ldots\}$. Each episode begins with the buyers simultaneously presenting opening offers; turn $t=1$ thus represents the seller's first tactical response. The complete definitions of $o_t$ alongside other components of the POMDP are detailed sequentially below.

\textbf{Seller Observation ($\mathcal{O}$).} The observation $o_t$ at turn $t$ aggregates all operational and linguistic information available to the seller. $o_t$ consists of:
\begin{enumerate}
    \item[(i)] The full dialogue history $\mathcal{H}_t$ available before seller turn $t$, including the buyers' opening offers when $t=1$.
    \item[(ii)] The public product catalog parameters $\Phi$, comprising per-item textual metadata, free-text descriptions, and list prices $\{P_{\text{list},j}\}_{j=1}^{n}$.
    \item[(iii)] The seller's private cost vector $\{\mathcal{C}_j\}_{j=1}^n$.
    \item[(iv)] A structured negotiation history $\mathcal{N}_t$ that tracks each buyer slot's current operational status (\textit{active}, \textit{satisfied}, \textit{quit}, or \textit{capped}), the number of messages already exchanged with that buyer ($u_{i,t}$, where $u_{i,t} \le K$), and the complete numerical sequence of all standing and historical price offers exchanged per buyer across all $n$ items.
    \item[(v)] The remaining global communication budget $T_t = T - \sum_{i=1}^{m} u_{i,t}$, representing the total remaining messages available to the seller.
\end{enumerate}
At each turn, the seller selects its structured action based on \(o_t\).
Information remains asymmetric: the seller never directly observes any entry of the valuation matrix \(\mathcal{B}\) and must infer buyer valuations from the observed interaction history.

\textbf{Action Space ($\mathcal{A}$).} At each turn $t$, the seller emits a structured, four-part message conforming to the following protocol:
\begin{quote}
   \texttt{Thought:} Internal reasoning where the seller plans its strategy (hidden from all buyers).\\
\texttt{Target:} The specific buyer $i \in \{1, \ldots, m\}$ that the seller chooses to talk to on this turn.\\
    \texttt{Talk:} The natural language dialogue sent exclusively to the targeted buyer.\\
    \texttt{Action:} An explicit command chosen from:
    \begin{itemize}
        \item \act{SELL}~$[\langle\texttt{Item-}j_1,\ \$O_1\rangle, \ldots, \langle\texttt{Item-}j_k,\ \$O_k\rangle]$: A variable-length price menu where the seller quotes a price $O_j \ge 0$ for any subset of items $j$.
        \item \act{DEAL}~\texttt{Item-}$j$~\$$O_j$: A singleton action accepting a standing buyer offer on item $j$.
        \item \act{QUIT}: Immediately terminates the entire multi-buyer episode.
    \end{itemize}
\end{quote}
The \texttt{Target} field determines which sealed buyer conversation the seller advances; non-targeted buyers remain silent and consume no turns. The variable-length format of the \act{SELL} action allows the seller to pitch multiple items simultaneously within a single turn. To avoid rule violations, the seller's action must target an active buyer, use valid items from the catalog, and exclude items that have already been sold.

\textbf{Transition Function ($\mathcal{P}$).} Upon parsing the seller's action at turn $t$, the environment sends the seller's message to the chosen buyer $i$. Based on the buyer's private valuations, this buyer replies with one of three actions: a list of counter-offers \act{BUY} for any subset of items, an acceptance \act{DEAL} for a single item, or a \act{QUIT} command to end the conversation.

The environment then updates the evolving components of $x_t$: it appends the exchange to the dialogue history $\mathcal{H}_{t+1}$, updates the structured negotiation history to $\mathcal{N}_{t+1}$, decrements the remaining global communication budget ($T_{t+1} = T_t - 1$), and increments the number of messages exchanged with that buyer ($u_{i, t+1} = u_{i,t} + 1$). The hidden valuation matrix $\mathcal{B}$, the catalog $\Phi$, and the cost vector $\mathcal{C}$ are fixed throughout the episode and are never modified by the transition. Writing the update as $x_{t+1} \sim \mathcal{P}(\cdot \mid x_t, a_t)$, the seller then receives $o_{t+1} = \Omega(x_{t+1})$ and the cycle repeats. The targeted buyer's status transitions to \emph{quit} if they walk away, \emph{capped} if $u_{i,t+1} = K$, or \emph{satisfied} if a deal is successfully closed. Inactive buyers are removed from the legal target space for the remainder of the episode.

\textbf{Termination.} An episode terminates immediately when any of the following conditions are met: (i) all $n$ items have been successfully sold; (ii) all buyer slots become inactive; (iii) the remaining global communication budget is fully depleted ($T_t = 0$); or (iv) the seller explicitly invokes \act{QUIT}. An episode also terminates immediately if (v) the seller commits an instruction violation, in which case the terminal reward is $\mathcal{R} = -1$ (\Cref{subsec:reward}).
 Because a buyer transitions to \emph{satisfied} and exits the market upon closing a single transaction, the episode yields at most $\min(n, m)$ deals, ensuring each buyer absorbs at most one item across the entire time horizon.

To ground this sequential decision framework in a concrete market setting, Figure~\ref{fig:dialogue-excerpt-first-two-turns} illustrates the first two turns of a trained agent negotiating a three-item video-recorder portfolio. The example demonstrates the seller's ability to navigate asymmetric information: the agent correctly rejects below-cost opening bids, anchors the buyer toward higher-margin assets  (\texttt{Target} and \texttt{Talk}), and secures a profitable deal while conserving communication budget. The full episode transcript and a comparative analysis against an untrained baseline are provided in Appendix~\ref{app:transcripts}.

\newcommand{\rendbuyer}[3]{}
\newcommand{\rendseller}[5]{}
\newcommand{\rendturndiv}{}

\newcommand{\runtranscript}{%
  \rendbuyer{Buyer 1 --- turn 1}%
            {How about \$40 for video\_recorders\_3?}%
            {\act{BUY} video\_recorders\_3 \$40.00}%
  \rendbuyer{Buyer 2 --- turn 1}%
            {I'd like to propose \$15 for video\_recorders\_3 — is that acceptable?}%
            {\act{BUY} video\_recorders\_3 \$15.00}%
  \rendbuyer{Buyer 3 --- turn 1}%
            {Let’s start with video\_recorders\_3. I propose \$35. Is that acceptable?}%
            {\act{BUY} video\_recorders\_3 \$35.00}%
  \rendseller{Seller --- turn 1}%
             {Buyer 1's offer of \$40 for item 3 is below my cost (\$42.07) — violation if accepted. Buyer 2's \$15 is also below cost — invalid. Buyer 3's \$35 is still below cost. I should target Buyer 1 first — they were first to act and may be serious. I’ll pitch video\_recorders\_1 at \$160 (close to list) and video\_recorders\_2 at \$230 (also near list), to test their willingness. I’ll start with Buyer 1 to maximize early leverage.}%
             {Buyer 1}%
             {Thank you for your opening. I cannot accept your \$40 offer for video\_recorders\_3 as it is below my cost. However, I am willing to negotiate on video\_recorders\_1 and video\_recorders\_2. I propose \$160 for video\_recorders\_1 and \$230 for video\_recorders\_2. Please let me know if you can meet either price.}%
             {\act{SELL} video\_recorders\_1 \$160.00, video\_recorders\_2 \$230.00}%
  \rendturndiv
  \rendbuyer{Buyer 1 --- turn 2}%
            {How about \$140 for video\_recorders\_1?}%
            {\act{BUY} video\_recorders\_1 \$140.00}%
  \rendseller{Seller --- turn 2}%
             {Buyer 1 has now submitted a new [BUY] for video\_recorders\_1 at \$140, which is above my cost (\$55.99) but below my proposed \$160. I must consider opportunity cost: if I reject this and try to push for a higher price, I risk losing the sale entirely, especially since Buyer 1 may be the only one interested in item 1. Accepting \$140 now gives me certainty and frees up my message budget for targeting Buyers 2 and 3 on the remaining items.}%
             {Buyer 1}%
             {I accept your offer of \$140 for video\_recorders\_1. The deal is closed. Item 1 has been sold, and I will no longer pitch it to you or others.}%
             {\act{DEAL} video\_recorders\_1 \$140.00}%
}

\begin{figure}[!tp]
\captionsetup{font=footnotesize}
\begingroup
\scriptsize
\setlength{\parskip}{2pt}
\renewcommand{\rendbuyer}[3]{%
  \noindent\begin{minipage}{0.75\textwidth}%
  \begin{tcolorbox}[enhanced,frame hidden,colback=blue!3!white,
    borderline west={2.4pt}{0pt}{blue!60!black},
    left=10pt,right=4pt,top=3pt,bottom=3pt,arc=0pt]
    \textbf{\textcolor{blue!60!black}{#1}}\\
    \textbf{Talk:} #2\\
    \textbf{Action:} #3
  \end{tcolorbox}%
  \end{minipage}\par\smallskip%
}
\renewcommand{\rendseller}[5]{%
  \hspace*{0.25\textwidth}\begin{minipage}{0.75\textwidth}%
  \begin{tcolorbox}[enhanced,frame hidden,colback=green!3!white,
    borderline east={2.4pt}{0pt}{green!50!black},
    left=4pt,right=10pt,top=3pt,bottom=3pt,arc=0pt]
    \hfill\textbf{\textcolor{green!50!black}{#1}}\\
    \textbf{Thought:} #2\\
    \textbf{Target:} #3\\
    \textbf{Talk:} #4\\
    \textbf{Action:} #5
  \end{tcolorbox}%
  \end{minipage}\par\smallskip%
}
\renewcommand{\rendturndiv}{\smallskip}
\begin{center}\footnotesize
{\bfseries\color{blue!60!black} Item catalog and hidden buyer valuations $\mathcal{B}_{i,j}$}\\[0.4em]
  \begin{tabular}{@{}lccc@{}}
    \toprule
     & video recorder 1 & video recorder 2 & video recorder 3 \\
     & {\scriptsize (ANNKE security-camera recorder)} & {\scriptsize (Lorex 4K camera recorder)} & {\scriptsize (ZOSI security-camera recorder)} \\
    \midrule
    List price & \$169.99 & \$249.99 & \$55.99 \\
    Seller cost & \$55.99 & \$100.00 & \$42.07 \\
    \midrule
    Buyer 1 valuation & \$147.60 & \$193.57 & \textbf{\$45.66} \\
    Buyer 2 valuation & \textbf{\$193.18} & \textbf{\$252.12} & \$22.60 \\
    Buyer 3 valuation & \$122.97 & \$149.76 & \$43.99 \\
    \bottomrule
  \end{tabular}
\end{center}
\smallskip
\runtranscript
\caption{\textbf{First two seller turns of a trained-seller episode on a three-item video-recorder catalog}. The table above the transcript gives the episode's catalog (list price, seller cost) and the hidden per-item buyer valuations (bold marks each item's top valuation). All three buyers open on the cheapest item with bids below even that item's cost; the seller rejects Buyer~1's opening bid, quotes near-list prices on the two high-margin items, and closes Item~1 one turn later at \$140. \emph{Talk} and \emph{Action} lines are verbatim; buyer \emph{Thought} fields are omitted, and the seller \emph{Thought} is abridged. The full transcript and the corresponding untrained-base-model transcript appear in Appendix~\ref{app:transcripts}.}
\label{fig:dialogue-excerpt-first-two-turns}
\Description{A table of the three-item catalog with list prices, seller costs and the hidden buyer valuation matrix, above a transcript of the first two seller turns. Each turn is shown as a labelled message box with the seller's Thought, Target, Talk and Action fields, alternating with the addressed buyer's reply.}
\endgroup
\end{figure}

\subsection{Verifiable Reward Formulation}
\label{subsec:reward}

The reward is a single terminal value computed directly from the finalized negotiation transcript. Given a catalog where public list prices exceed seller costs ($P_{\text{list},j} > \mathcal{C}_j, \forall j$), the \emph{catalog-wide list-price margin} is $\sum_{j=1}^{n} (P_{\text{list},j} - \mathcal{C}_j)$, the total margin that would result from selling every catalog item at its list price. This quantity is used to normalize the reward. The terminal reward $\mathcal{R}$ is defined as:

\begingroup
\fontsize{10.5pt}{12pt}\selectfont
\begin{equation}
\mathcal{R} =
    \begin{cases}
        \displaystyle \frac{\sum_{j=1}^{n} \min\big(P_{\text{deal},j} - \mathcal{C}_j,\, P_{\text{list},j} - \mathcal{C}_j\big) \cdot \mathds{1}[\text{item } j \text{ sold}]}{\sum_{j=1}^{n} \big(P_{\text{list},j} - \mathcal{C}_j\big)} & \text{if the seller commits no instruction violation,} \\[1.2em]
        -1 & \text{if the seller commits an instruction violation,}
    \end{cases}
    \label{def:multiitem_reward}
\end{equation}
\endgroup

\noindent where $P_{\text{deal},j}$ is the finalized transaction price, $\mathcal{C}_j$ is the seller's private cost, and $P_{\text{list},j}$ is the public list price for item $j$. This formulation caps individual item contributions to their list-level surplus, bounding the non-violation reward to $[0, 1]$.

\paragraph{Successful Deals and Surplus Extraction.} For episodes without instruction violations, $\mathcal{R}$ measures the \emph{Seller Surplus Extraction Ratio}—the percentage of the catalog-wide list-price margin that the seller successfully captures. This reward structure encourages three key strategic behaviors: (1) unsold items hurt the final reward because they contribute to the denominator without adding to the numerator, pushing the seller to close more deals when profitable; (2) varying item values naturally incentivize the seller to focus on more profitable products; and (3) since each buyer can purchase at most one item, accepting a poor offer early carries a high opportunity cost by permanently removing that buyer from the market.

\paragraph{No-Deal Terminal.} If the global communication budget is exhausted or all channels close without a transaction, the reward defaults to $\mathcal{R} = 0$. This ensures that walking away is always preferred to selling at a loss. Specifically, if a buyer's underlying valuation is lower than the seller's cost and no positive surplus can be reached, walking away protects the agent from an instruction violation penalty while still encouraging it to find profitable deals when they are possible.

\paragraph{Instruction Violation.} A penalty of $\mathcal{R} = -1$ is  applied if the seller breaks the core rules. This penalty triggers if the seller quotes a price below its cost ($O_j < \mathcal{C}_j$), tries to target an inactive buyer, includes an item that has already been sold, or fails to follow the required message format. Putting these penalties directly into the reward function ensures that the seller learns to follow the text formatting rules and never sells at a loss.

\section{The Substitutable-Product Dataset}
\label{sec:dataset}

\begin{figure}[h]
  \centering
  \includegraphics[width=0.65\linewidth]{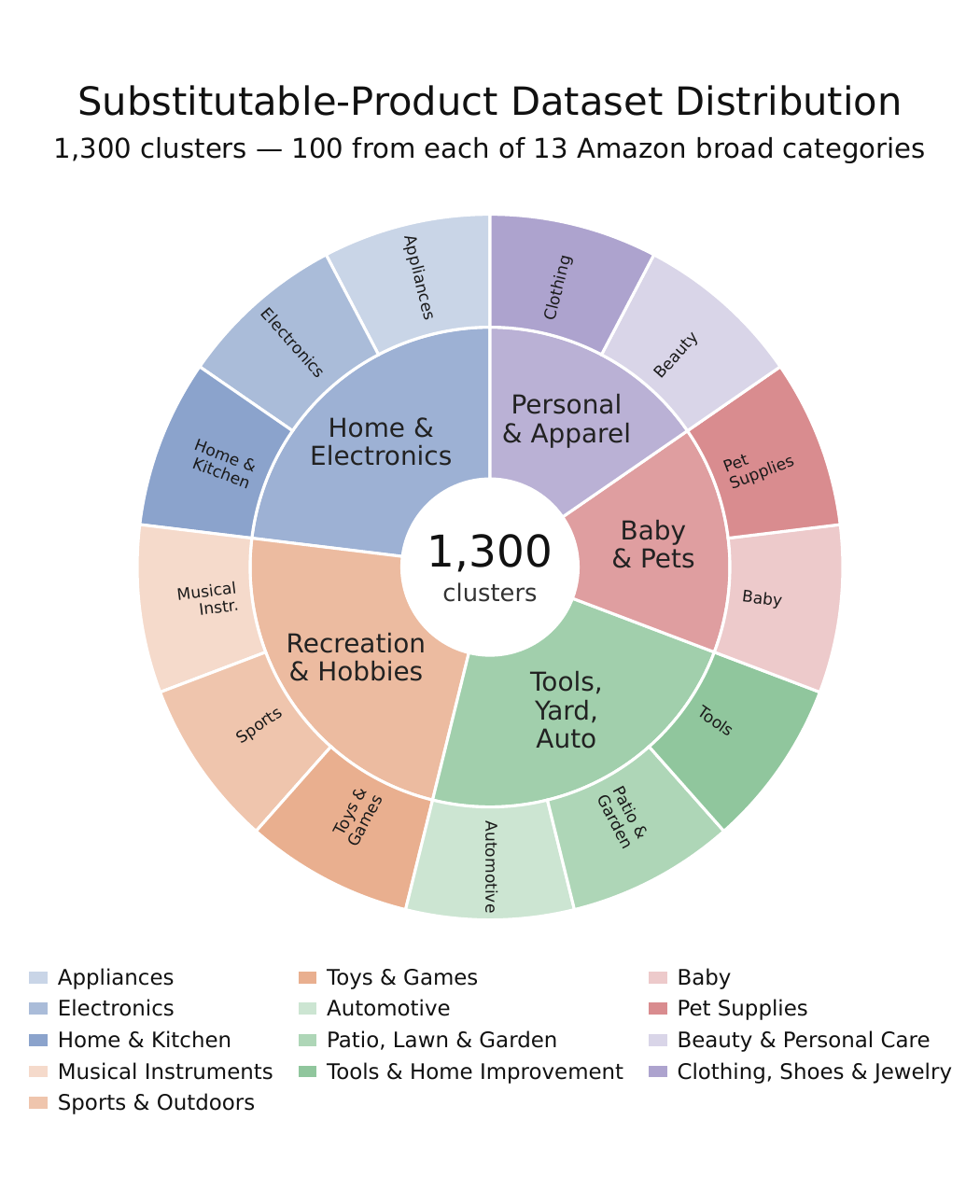}
  \caption{Composition of the substitutable-product dataset. The outer ring shows the $13$ Amazon broad categories ($100$ clusters each); the inner ring groups them into five themes; the center reports the $1{,}300$-cluster total.}
  \label{fig:dataset_pie}
  \Description{A two-level ring chart. The outer ring is divided into 13 equal segments, one per Amazon broad category, each holding 100 clusters; the inner ring groups those segments into five themes of unequal size; the centre reports the 1,300-cluster total.}
\end{figure}

To ground every negotiation episode in realistic product economics, we curate a dataset of substitutable Amazon products. The dataset comprises $1{,}300$ clusters---$100$ from each of $13$ broad Amazon categories---and each cluster contains $5$ substitutable products of a single narrow product type (for example, electric guitars, over-ear headphones, or food processors), for a total of $6{,}500$ products. Every product carries its title, brand, public list price $P_{\text{list},j}$, private seller's cost $\mathcal{C}_j$, and free-text product features; every cluster additionally records its brand count, list-price range, and a written substitutability rationale. This substitutability ensures that every item in a cluster serves as a practical alternative for the buyer, requiring the seller to solve an active matching problem to optimize total seller surplus. Furthermore, the dataset guarantees $P_{\text{list},j} > \mathcal{C}_j$ for every product, so the reward in  \Cref{subsec:reward} is well-defined on every episode.

Across the dataset, the five items within any given product cluster exhibit high brand diversity, averaging 4.4 unique brands per group. Furthermore, to ensure the pricing environment is non-trivial, the median ratio between the highest and lowest public list price within a single cluster is 1.85. Comprehensive per-category descriptive statistics are deferred to Appendix~\ref{app:dataset}.

\subsection{Dataset Construction} The dataset is assembled from live Amazon catalog data accessed through Keepa \citep{Keepa2026}, a third-party data service that records Amazon's product catalog, sales rankings, and price history. Construction follows a three-step pipeline: first, a language model proposes concrete product types within each broad category; second, the best-selling Amazon items for those types are retrieved via the Keepa API; third, an LLM reviewer filters the results to ensure strict functional substitutability, retaining only items that serve identical use cases while generating a text-based justification for each finalized cluster. Following the dataset formulation in \citet{xia2024measuring}, each product's seller cost $\mathcal{C}_j$ is set to the lowest price at which that product was ever sold on Amazon, and its list price $P_{\text{list},j}$ is the publicly posted price at collection time.

\subsection{Dataset Splits}
For training and in-distribution evaluation, the 12 base categories are split 80/20 at the cluster level into 960 training clusters and 240 test clusters. This split ensures that evaluation always operates on product contexts unseen during training. The 13th category, Musical Instruments (100 clusters), is held out for the out-of-distribution evaluation described in \Cref{sec:ood}. Its list prices are distinct from the other 12 base categories, concentrating in a higher range (median \$165.49 versus \$44.10 for the other 12 categories); the complete price distribution comparison is detailed in Appendix~\ref{app:dataset}.
Because every valuation in an episode scales with the item's list price (\Cref{subsec:market_settings}), episodes drawn from this category place the seller in a price range it never encountered during training.

\section{Simulation Environment and Experimental Setup}
\label{sec:experimental_setup}

This section details the operational setup of our negotiation framework for substitutable product portfolios, the economic parameterization of the marketplace, and the algorithmic parameters used to optimize the seller agent. 

\subsection{Agent Configurations and Training Setup}
\label{subsec:agent_models}

We implement both the seller and buyer roles using the \texttt{Qwen3-30B-A3B-Instruct-2507} model \citep{qwen3technicalreport}. Because our primary objective is to study how an LLM seller can manage substitutable product portfolios under resource constraints, the buyer models remain fixed during the training loop while the seller agent actively updates its parameters. Specifically, the buyer slots during training are represented by an iteration-60 checkpoint optimized via Reinforcement Learning from Verifiable Rewards (RLVR) on a separate bilateral negotiation task \citep{liu2026instructing}. We optimize the learning seller agent using the CISPO objective \citep{chen2025minimax} without an explicit KL divergence penalty, executing the training sequence for a single epoch over the active in-distribution training split. To preserve empirical reproducibility, all granular optimization hyperparameters, batch sizes, and role-specific sampling temperatures are detailed in Appendix~\ref{subsec:hyperparams}.

Because the baseline buyer checkpoint was trained exclusively on single-item negotiations, we use system prompts to adapt it to multi-product settings. These prompts describe the product catalog, the formatting rules for offers, the private buyer valuations, and the rule that a buyer exits the market immediately after its first closed deal. To ensure the buyers behave rationally and adhere to the negotiation protocol, we employ a pre-training validation pass. We verify that all prompted buyer agents maintain a protocol violation rate below 1\% across a diverse range of market scenarios; prompt details are provided in Appendix \ref{app:prompts}.

\subsection{Market Settings and Information Structure}
\label{subsec:market_settings}

We partition the marketplace dataset (\Cref{sec:dataset}) at the cluster level into 960 training clusters and 240 held-out test clusters across 12 product categories. This split ensures that the model is evaluated on product contexts unseen during training, with buyer valuations freshly sampled. During each simulation episode, we sample a single cluster from the training split and uniformly at random select \(n=3\) of its five substitutable items without replacement, then match the seller concurrently against \(m=3\) independent buyers. Over the course of an episode, the seller operates under a global communication budget of $T = 7$ total dialogue turns, subject to a per-buyer turn limit of $K = 3$ turns.

Every entry of the hidden buyer valuation matrix $\mathcal{B} \in \mathbb{R}^{m \times n}$ is sampled stochastically at the initialization of each episode. The valuation for buyer $i$ evaluating item $j$ is formalized as:
\begin{equation}
    \mathcal{B}_{i,j} = r_{i,j} \cdot P_{\text{list},j}
    \label{eq:budget_gen}
\end{equation}
where $P_{\text{list},j}$ represents the public list price of the item, and the underlying valuation ratio is drawn independently and uniformly from the continuous interval $r_{i,j} \sim \mathcal{U}[0.40, 1.15]$. Because this interval spans widely above and below the public list price anchor ($1.0$), the seller cannot assume a fixed margin baseline and must use dialogue to actively probe for hidden values.

Prior work demonstrates that standard language models deployed as bargaining agents often exhibit a systematic bias, emitting initial bids that correlate with their hidden private valuations \citep{xia2024measuring, liu2026instructing}. To prevent the seller from exploiting these immediate first-turn heuristics, which directly reflect the counterparty's underlying preferences, we inject structural noise into the buyers' opening offers. At the start of an episode, each buyer freely selects an initial subset of items to target. With a probability of 0.5, the opening offer price on every selected item is completely decoupled from the buyer's true underlying valuation and resampled independently into the low-ball interval $[0.25, 0.30] \cdot P_{\text{list},j}$. Otherwise, the buyer generates its opening offer freely subject to its private valuation constraint, which tends to reflect its private valuation.. This design breaks the correlation between initial bids and buyers' private valuations, forcing the seller to leverage multi-turn dialogue to recover accurate preferences.

\section{Evaluation Metrics and Framework}
\label{sec:evaluation}

\begin{figure*}[t]
\centering
\includegraphics[width=\textwidth]{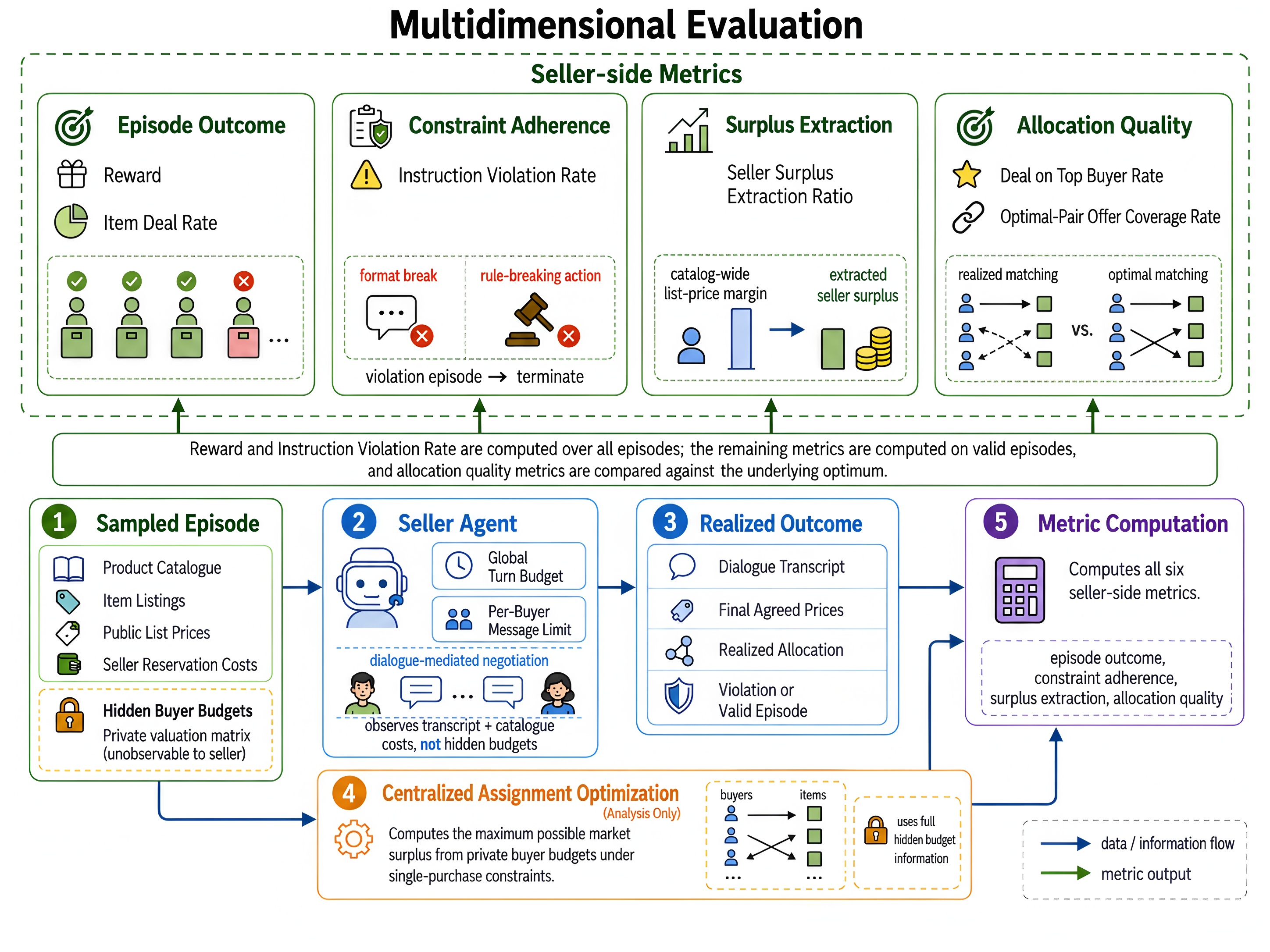}
\caption{Schematic Architecture of the Multidimensional Evaluation Framework. The evaluation pipeline computes performance metrics across five phases: (1) a Sampled Episode sets global and local turn constraints; (2) the Seller Agent negotiates with several buyer agents over the public catalog under asymmetric information; (3) a Realized Outcome captures the negotiation transcript, deal prices, and allocation; (4) a Centralized Assignment Optimization determines the maximum possible surplus from the private buyer valuations; and (5) the Metric Computation engine overlays realized results against this optimum to measure \textit{Episode Outcome}, \textit{Constraint Adherence}, \textit{Surplus Extraction}, and \textit{Allocation Quality}.}
\label{fig:evaluation_framework}
\Description{A pipeline diagram in five stages: a sampled episode fixing the global and per-buyer turn limits; the seller agent negotiating with several buyer agents over the public catalog; the resulting transcripts and closed deals; an oracle that computes the surplus-maximizing assignment; and a final block of metrics grouped into episode outcome, constraint adherence, surplus extraction, and allocation quality.}
\end{figure*}

This section introduces a multidimensional framework to evaluate the trained seller's matching and negotiation skills under information asymmetry and communication constraints. Because the seller concurrently manages substitutable assets across independent buyers under a strict communication budget, revenue alone provides an incomplete picture of overall performance. To this end, \Cref{fig:evaluation_framework} outlines our evaluation pipeline, which systematically quantifies performance across six metrics in four families as defined in the following subsections:  episode outcome (\textit{Reward}, \textit{Item Deal Rate}), constraint adherence (\textit{Instruction Violation Rate}), surplus extraction (\textit{Seller Surplus Extraction Ratio}), and allocation quality (\textit{Deal on Top Buyer Rate, Optimal-Pair Offer Coverage Rate}).

During evaluation, a seller violation---such as quoting a price below cost, targeting an inactive buyer, listing an item that has already been sold, or emitting an unparseable message---terminates the episode immediately at a penalty reward of $\mathcal{R} = -1$. \textit{Reward} and \textit{Instruction Violation Rate} are computed over all episodes. All other metrics are computed over valid episodes completed without a violation.
The oracle matching model underpinning our allocation metrics, together with their full formal definitions, is provided in Appendix~\ref{app:backup_metrics}.

\subsection{Episode Outcome}

\paragraph{Reward.} The terminal reward defined in equation~\eqref{def:multiitem_reward}; $\mathcal{R} = -1$ on a violation episode, and $\mathcal{R} \in [0, 1]$ on a valid episode, regardless of the total number of deals successfully closed across the catalog.

\paragraph{Item Deal Rate.} The fraction of (item, episode) pairs that close a deal, bounded within $[0,1]$. Here, $n$ is the number of items in the catalog, so the denominator represents the total opportunities to close a deal.
\[
  \text{Item Deal Rate} \;=\; \frac{\text{Total Items Sold}}{n \times \text{Total Valid Episodes}}.
\]

\subsection{Constraint Adherence}

\paragraph{Instruction Violation Rate.} The fraction of episodes that terminate in a seller violation, representing the share of all episodes on which the seller executes an invalid action and receives a penalty reward:
\[
  \text{Instruction Violation Rate} \;=\; \frac{\text{Total Violation Episodes}}{\text{Total Episodes}}.
\]
A value of zero corresponds to perfect protocol compliance. Because each violation incurs the full penalty reward of $-1$, this rate represents a driver of the reward discrepancy between tested models. 

\subsection{Surplus Extraction}

\paragraph{Seller Surplus Extraction Ratio.} The episode-level fraction of the catalog-wide list-price margin captured by the seller,
\[
  \text{Seller Surplus Extraction Ratio} \;=\; \frac{\sum_{j=1}^{n} (P_{\text{final},j} - \mathcal{C}_j) \cdot \mathds{1}[\text{item } j \text{ sold}]}{\sum_{j=1}^{n} (P_{\text{list},j} - \mathcal{C}_j)},
\]
where $P_{\text{final},j} := \min(P_{\text{deal},j}, P_{\text{list},j})$. This formulation matches the non-violation branch of the terminal reward introduced in \Cref{subsec:reward}. We aggregate this metric over all valid episodes, including those that close no deals: the catalog-wide denominator $\sum_j (P_{\text{list},j} - \mathcal{C}_j)$ is strictly positive on every episode, so a no-deal episode contributes a well-defined ratio of 0. Because the seller’s objective is total episode seller surplus rather than the favorability of any single trade, we define this ratio at the episode level rather than per item.

\subsection{Allocation Quality}
\label{subsec:allocation_metrics}

The allocation metrics compare the seller's realized buyer--item assignments against an \emph{oracle optimum}, which represents the matching of items to buyers that maximizes total seller surplus. This optimum follows the one-deal-per-buyer constraint, with each buyer--item pair valued by the nonnegative seller surplus attainable at the lower of the buyer's valuation and the item's list price. We denote the set of these surplus-maximizing matchings as $\mathcal{M}^\star$, and the seller's realized allocation as $A$, where $A(j)$ is the buyer who purchased item $j$ (or $\varnothing$ if unsold). The optimum is computed from the hidden buyer valuation matrix $\mathcal{B}$ for evaluation purposes only; it is never observed by the seller. Appendix~\ref{app:backup_metrics} provides the complete formal definitions of the oracle matching model and these allocation metrics.

\paragraph{Deal on Top Buyer Rate.} Because buyers hold distinct private valuations for the same asset, an optimal seller should use the multi-turn negotiation to discover which buyer values each item the most. To evaluate this routing capability, this metric measures how often a successfully sold item is allocated to its highest-valuation buyer, pooled across all valid episodes:
\[
  \text{Deal on Top Buyer Rate} \;=\; \frac{\sum_{\text{ep}} \sum_{j \in S} \mathds{1}\big[\, A(j) \in \argmax_i \mathcal{B}_{i,j} \,\big]}{\sum_{\text{ep}} |S|},
\]
where $S$ is the set of items sold in a given episode, $A(j)$ denotes the buyer assigned to item $j$, and $\mathcal{B}_{i,j}$ represents the valuation of buyer $i$ for item $j$. The term $\mathds{1}[\cdot]$ is the indicator function, which evaluates to $1$ if the statement inside the brackets holds true and $0$ otherwise.
This rate is calculated over sold items rather than averaged per episode so that each transaction contributes equally. 
Because a single buyer might hold the highest valuation for multiple items but can purchase at most one item under the matching constraint, a rate below 1.0 does not inherently imply routing suboptimality.

\paragraph{Optimal-Pair Offer Coverage Rate.} This metric measures how frequently the seller addresses the optimal buyer–item pairings identified by the oracle optimum in its negotiation conversation, through either a \act{SELL} or a \act{DEAL} action. Because the surplus-maximizing matching is not necessarily unique, we select the specific optimal matching $M \in \mathcal{M}^\star$  that yields the highest overlap with the seller's actual proposals. We then count how many of these optimal pairs the seller actually proposed during the negotiation—by targeting buyer $i$ with item $j$ in at least one \act{SELL} or \act{DEAL} action—regardless of whether a final deal was reached. The rate pools these counts across all valid episodes containing a non-empty optimal assignment:
\[
  \text{Optimal-Pair Offer Coverage Rate} \;=\; \frac{\text{Total Optimal Pairs Offered}}{\text{Total Pairs in Optimal Matching}}.
\]
Unlike the \textit{Deal on Top Buyer Rate}, this scores matching \emph{intent} rather than final execution, isolating strategic routing and buyer exploration from final price agreements.

\section{Performance Benchmarking}
\label{sec:benchmark}

This section evaluates the empirical performance of our trained seller agent against state-of-the-art frontier language models, encompassing leading closed-source systems and large-scale open-weight architectures. We evaluate how well these agents maximize seller surplus under strict communication limits and hidden buyer valuations, contrasting task-specific reinforcement learning with strong general-purpose models.

\begin{table}[h]
  \centering
  \small
  \setlength{\tabcolsep}{4pt}
  \resizebox{\linewidth}{!}{%
\begin{tabular}{lccccccc}
\toprule
Model & Params & Reward $\uparrow$ & \makecell{Seller Surplus\\Extraction\\Ratio} & \makecell{Item\\Deal\\Rate} & \makecell{Deal on\\Top Buyer\\Rate} & \makecell{Optimal-Pair\\Offer Coverage\\Rate} & \makecell{Instruction\\Violation\\Rate} \\
\midrule
\rowcolor{gray!10} Qwen3-30B-A3B-Instruct-2507-trained (\textbf{Ours}) & 30B & $\mathbf{+0.407}$ & $\mathbf{+0.425}$ & $64.7\%$ & $\mathbf{0.687}$ & $\mathbf{0.844}$ & $1.3\%$ \\
\rowcolor{gray!10} Qwen3-30B-A3B-Instruct-2507-untrained & 30B & $-0.329$ & $+0.184$ & $66.2\%$ & $0.544$ & $0.554$ & $43.3\%$ \\
\midrule
Kimi-K2.6-thinking & 1T & $\underline{+0.379}$ & $+0.379$ & $73.0\%$ & $0.566$ & $0.559$ & $\mathbf{0.0\%}$ \\
DeepSeek-V4-Pro-thinking & 1.6T & $+0.369$ & $\underline{+0.392}$ & $73.4\%$ & $0.612$ & $0.592$ & $1.7\%$ \\
DeepSeek-V4-Pro-nothink & 1.6T & $+0.361$ & $+0.378$ & $\underline{76.2\%}$ & $0.596$ & $0.640$ & $1.3\%$ \\
DeepSeek-V4-Flash-thinking & 284B & $+0.349$ & $+0.366$ & $71.0\%$ & $0.568$ & $0.566$ & $1.3\%$ \\
Kimi-K2.6-nothink & 1T & $+0.331$ & $+0.378$ & $75.0\%$ & $\underline{0.616}$ & $\underline{0.654}$ & $3.4\%$ \\
GPT-5.4-high-reasoning & closed-source & $+0.309$ & $+0.315$ & $\mathbf{78.1\%}$ & $0.560$ & $0.604$ & $\underline{0.4\%}$ \\
DeepSeek-V4-Flash-nothink & 284B & $+0.241$ & $+0.306$ & $73.4\%$ & $0.538$ & $0.543$ & $5.0\%$ \\
GPT-5.4-mini-high-reasoning & closed-source & $+0.111$ & $+0.212$ & $72.3\%$ & $0.465$ & $0.457$ & $8.4\%$ \\
\bottomrule
\end{tabular}
  }
  \caption{\textbf{In-distribution held-out evaluation} ($3$ buyers, $3$ items, $T=7$). The trained checkpoint (\textbf{Ours}) and the untrained base above the rule; frontier baselines below. Baselines sorted by \textit{Reward} (descending); best per column in bold, second-best underlined (\textit{Instruction Violation Rate} is lower-is-better). Metric definitions in \Cref{sec:evaluation}. Standard errors in \Cref{tab:indist3_appendix}.}
  \label{tab:indist3_main}
\end{table}

\subsection{Experimental Configuration and Baselines}
\label{subsec:benchmarking_setup}

We benchmark the trained \texttt{Qwen3-30B-A3B-Instruct-2507} seller against its untrained base model \citep{qwen3technicalreport} and a broad panel of baselines: \texttt{GPT-5.4} \citep{singh2025openai}, \texttt{GPT-5.4-mini} \citep{singh2025openai}, \texttt{DeepSeek-V4-Pro} \citep{deepseekai2026deepseekv4}, \texttt{DeepSeek-V4-Flash} \citep{deepseekai2026deepseekv4}, and \texttt{Kimi-K2.6} \citep{team2025kimi}. To measure the impact of extended reasoning, the open-weight models are evaluated in both their standard (\texttt{nothink}) and extended internal reasoning (\texttt{thinking}) configurations, while the closed-source models are evaluated at their maximum reasoning effort (full specifications are provided in \Cref{tab:model_sources_full}).
Notably, both our trained model and its untrained base architecture \texttt{Qwen3-30B-A3B-Instruct-2507} operate without a provider-native hidden reasoning mode. All sellers, including these two, still emit the protocol's \texttt{Thought} field, which is part of the structured action format rather than a hidden reasoning channel.

Every model plays the seller role within the interactive environment introduced in \Cref{sec:methodology} on the exact same negotiation scenarios. Performance is quantified across the primary evaluation dimensions established in \Cref{sec:evaluation}: episode outcome (\textit{Reward} and \textit{Item Deal Rate}), constraint adherence (\textit{Instruction Violation Rate}), economic margin (\textit{Seller Surplus Extraction Ratio}), and assignment precision (\textit{Deal on Top Buyer Rate} and \textit{Optimal-Pair Offer Coverage Rate}). The quantitative results on the in-distribution held-out test split ($3$ buyers, $3$ items, $T=7$) are summarized in \Cref{tab:indist3_main}, and the relative performance rankings across key metrics are visualized in Appendix~\ref{sec:indist_benchmark_bars}.

\subsection{Surplus Extraction and Economic Trade-offs}
\label{subsec:surplus_tradeoffs}

As shown across these evaluations, the trained seller model, though significantly smaller in parameter size compared to the tested baselines (see Table~\ref{tab:model_sources_full}), leads the benchmark panel across all primary economic dimensions. The trained seller achieves a top-performing mean \textit{Reward} of $+0.407$ and a \textit{Seller Surplus Extraction Ratio} of $+0.425$, outperforming trillion-parameter commercial models and high reasoning frontiers like \texttt{Kimi-K2.6-thinking} ($+0.379$ reward) and \texttt{DeepSeek-V4-Pro-thinking} ($+0.369$ reward).
Notably, our trained 30B model outperforms these heavy-reasoning models without using a provider-native hidden reasoning mode. This demonstrates that task-specific training can achieve high strategic performance at a fraction of the model size, without relying on a provider-native hidden reasoning mode.

This performance advantage stems from a fundamental divergence in negotiation strategy. The trained seller model successfully trades a slightly lower overall \textit{Item Deal Rate} ($64.7\%$) for a vastly higher realized seller surplus. In contrast, the generalist model baselines typically adopt the opposite approach, prioritizing raw transaction volume over deal quality. For example, \texttt{GPT-5.4-high-reasoning} maximizes transaction volume with a panel-leading \textit{Item Deal Rate} of $78.1\%$, yet yields a depressed surplus extraction ratio of $+0.315$. Across our benchmark, the evaluated general-purpose models achieve higher agreement volume but lower economic margin. This pattern is consistent with prior reports of cooperative and economically suboptimal behavior in general-purpose language agents \citep{chawla2023selfish, xia2024measuring}.  Conversely, our trained agent internalizes the opportunity cost of premature agreements imposed by the one-deal-per-buyer constraint; it leverages selective rejections, anchors firmly at list-price boundaries, and preserves asset availability for higher-value opportunities.

\subsection{Allocation Quality}
\label{subsec:allocation_precision}

The agent's ability to maximize surplus depends heavily on how well it allocates inventory when buyer preferences are unknown. Because each buyer only purchases a single item and holds private valuations across substitutable products, a seller cannot maximize seller surplus by making blind, random offers. As shown in Table~\ref{tab:indist3_main}, our model outperforms the baselines on these allocation metrics, achieving an \textit{Optimal-Pair Offer Coverage Rate} of $0.844$ and a \textit{Deal on Top Buyer Rate} of $0.687$.

These results demonstrate that the post-training method successfully sharpens the agent's ability to uncover hidden buyer valuations. Instead of making untargeted proposals across the buyer pool, the specialized 30B model uses the multi-turn dialogue to systematically learn what each buyer values. This allows the agent to accurately match items to high-valuation buyers, optimizing inventory allocation within the conversation turn limit.
While larger generalist models often sacrifice profit to reach an  agreement, our method instead teaches the agent to find the right buyer for each item. The emergence of this capability without explicit supervision validates the effectiveness of our post-training method.

\section{Empirical Training Dynamics}
\label{sec:training_dynamics}

This section analyzes the behavioral changes in the seller agent throughout post-training, which utilizes Reinforcement Learning from Verifiable Rewards (RLVR) paired with the CISPO objective to optimize the agent against frozen buyer counterparts. To evaluate the progression of the seller agent holistically, its performance is tracked across the multidimensional metrics defined in \Cref{sec:evaluation}.

\begin{figure}[tbp]
  \centering
  \begin{minipage}[t]{0.49\textwidth}
    \centering
    \includegraphics[width=\linewidth]{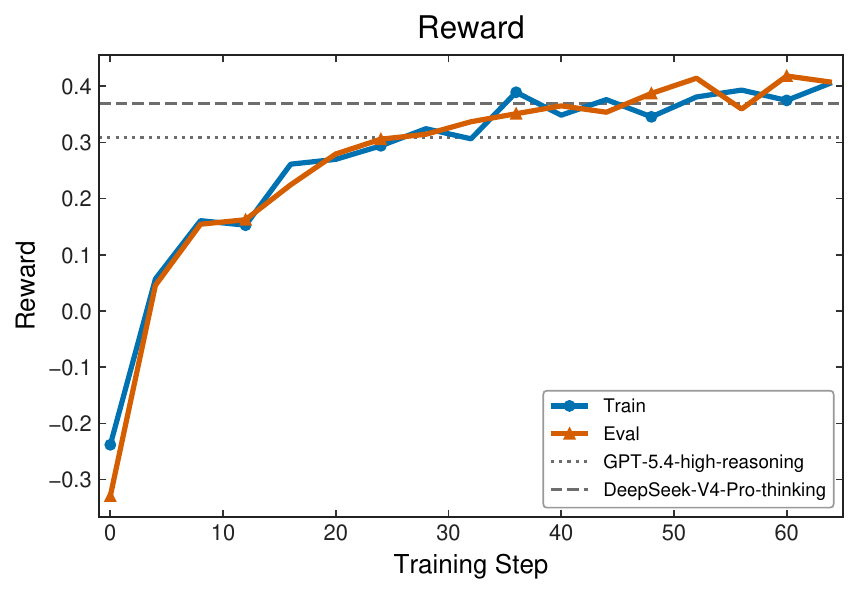}
    \caption{\textbf{Reward.} In-distribution curve across training steps. The blue series tracks the training dynamic, while the orange series marks performance on the held-out test split ($m=3$ buyers, $n=3$ items, global communication budget $T=7$). Horizontal reference lines show frontier model benchmarks.}
\label{fig:train_reward}
    \label{fig:training_curve_reward}
  \end{minipage}\hfill
    \begin{minipage}[t]{0.49\textwidth}
    \centering
    \includegraphics[width=\linewidth]{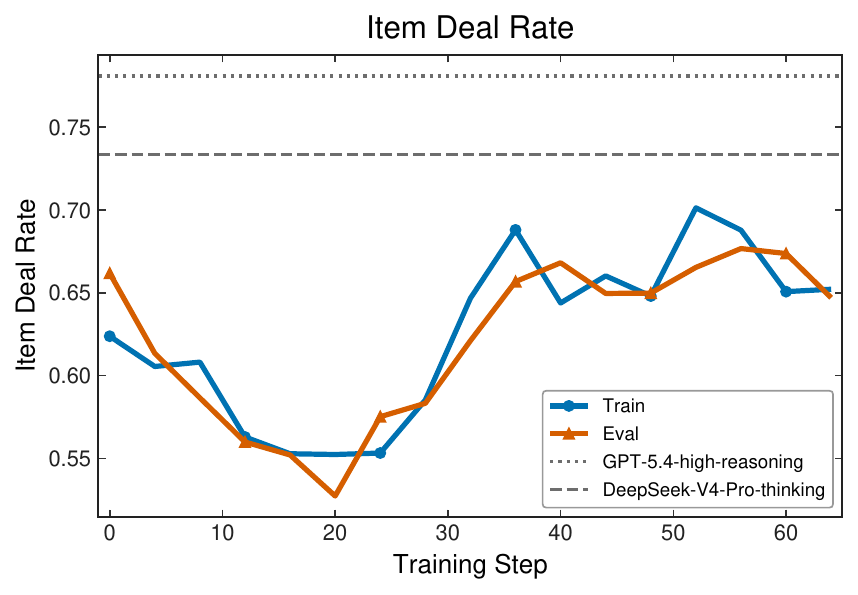}
    \caption{\textbf{Item Deal Rate.} In-distribution curve across training steps. Experimental setup, series definitions, and benchmarks match Figure~\ref{fig:train_reward}.}
\label{fig:train_deal_rate}
    \label{fig:training_curve_per_item_deal_rate}
  \Description{Two line charts against training step. Left, Reward: both the blue training series and the orange held-out test series rise steeply over roughly the first 20 steps and then flatten well above their starting levels. Right, Item Deal Rate: both series decline to a local minimum near step 20 and then recover; the training series ends above its starting level, while the held-out test series ends slightly below its starting level.}
  \end{minipage}\hfill
\end{figure}

\begin{figure}[tbp]
  \centering
   \begin{minipage}[t]{0.49\textwidth}
    \centering
    \includegraphics[width=\linewidth]{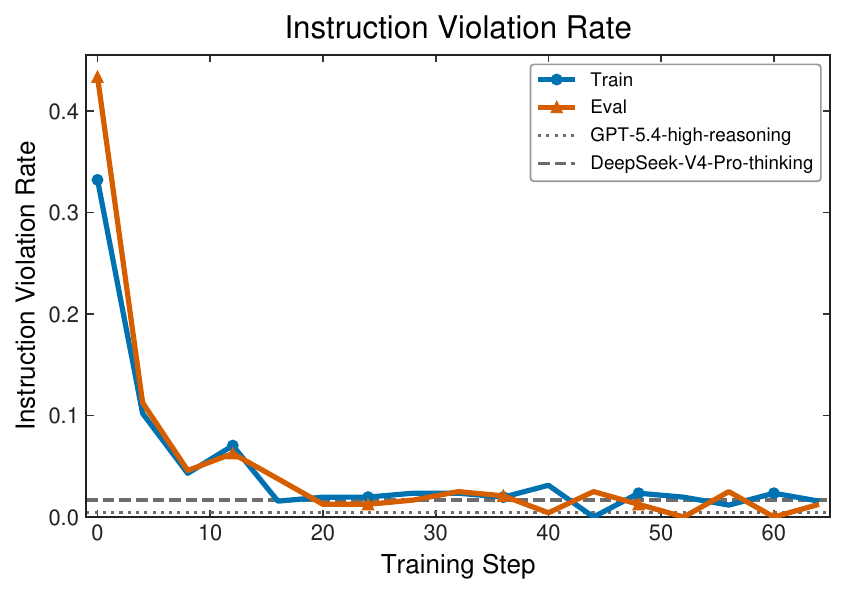}
    \caption{\textbf{Instruction Violation Rate.} In-distribution curve across training steps. Experimental setup, series definitions, and benchmarks match Figure~\ref{fig:train_reward}.}
    \label{fig:training_curve_instruction_violation_rate}
  \end{minipage}
 \begin{minipage}[t]{0.49\textwidth}
    \centering
    \includegraphics[width=\linewidth]{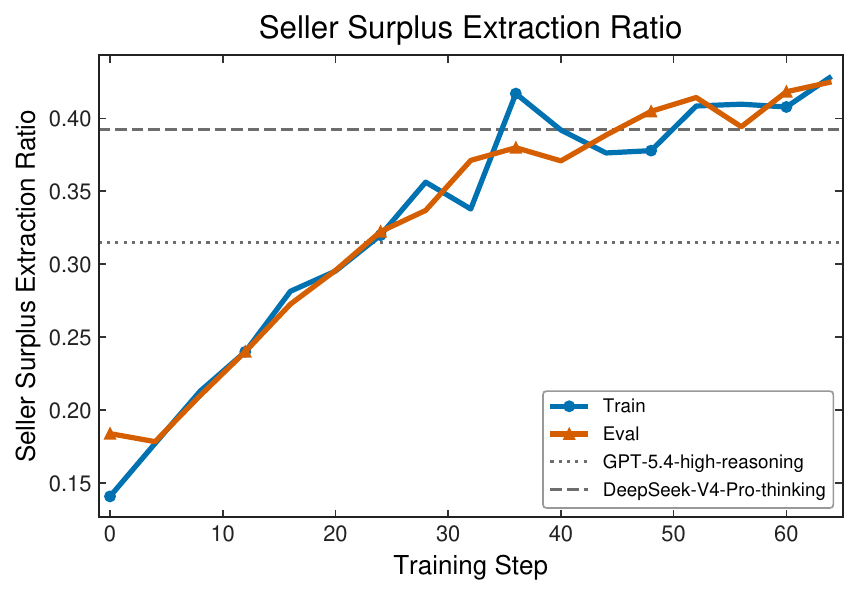}
    \caption{\textbf{Seller Surplus Extraction Ratio.} In-distribution curve across training steps. Experimental setup, series definitions, and benchmarks match Figure~\ref{fig:train_reward}.}
    \label{fig:training_curve_seller_surplus_extraction_ratio}
  \Description{Two line charts against training step. Left, Instruction Violation Rate: both series start high and fall sharply toward zero within the first steps, then stay flat near zero. Right, Seller Surplus Extraction Ratio: both series rise steadily and end above every frontier-model reference line.}
  \end{minipage}
\end{figure}

The empirical learning trajectories across these metrics are visualized in \Crefrange{fig:training_curve_reward}{fig:training_curve_deal_on_top_buyer_rate}. Across each panel, the blue series denotes performance during training rollouts, while the orange series tracks performance on the held-out test split of assets unseen during training (see \cref{subsec:market_settings}). The horizontal grey dotted and dashed lines indicate the performance of \texttt{GPT-5.4-high-reasoning} and \texttt{DeepSeek-V4-Pro-thinking}, respectively.  Overall, the trajectories show simultaneous improvements in protocol compliance, surplus extraction, and allocation accuracy as reinforcement learning progresses.

\subsection{Episode Outcome}
This subsection analyzes the episode outcomes via the terminal \textit{Reward} (\Cref{fig:training_curve_reward}) and the total transaction volume captured by the \textit{Item Deal Rate} (\Cref{fig:training_curve_per_item_deal_rate}). The \textit{Reward}, defined in equation~\eqref{def:multiitem_reward}, balances constraint adherence with surplus extraction. This metric increases over the course of training, moving from $-0.329$ for the untrained base model to $+0.407$ at the final checkpoint, as also recorded in \Cref{tab:indist3_main}. At the end of training, the trained 30B seller agent achieves a higher reward than frontier architectures with larger parameter scales, including \texttt{GPT-5.4-high-reasoning} ($+0.309$) and \texttt{DeepSeek-V4-Pro-thinking} ($+0.369$).

The \textit{Item Deal Rate} (\Cref{fig:training_curve_per_item_deal_rate}) exhibits a non-monotonic trajectory over the course of optimization. From step 0 to step 20, the rate decreases to a local minimum. This initial decline reflects a behavioral shift toward rejecting low-margin opening offers from buyers rather than concluding agreements immediately to clear inventory. As the agent learns to make effective counter-offers in later turns, the deal rate recovers and stabilizes by the end of training. 

This trajectory demonstrates that maximizing portfolio surplus does not solely depend on increasing the total volume of finalized transactions. Deferring or refusing low-margin agreements preserves inventory capacity for higher-margin opportunities and alternative buyer-product alignments. In comparison, generalist models prioritize transaction frequency over margin optimization; for instance, \texttt{GPT-5.4-high-reasoning} achieves the highest \textit{Item Deal Rate} in the baseline panel but obtains a lower terminal reward due to the acceptance of some suboptimal offers.

\subsection{Constraint Adherence}
Executing multi-item inventory allocations requires the agent to operate within a formalized negotiation protocol. Rather than treating the interaction as open-ended text, establishing a structured protocol ensures that the agent's outputs map to verifiable actions that can be systematically guided via reinforcement learning. These protocol rules encompass syntactic formatting requirements (such as outputting valid, parsable action tokens), critical economic boundaries (such as prohibiting cost violations where a quoted price falls below the private reservation cost $\mathcal{C}_j$), and structural market constraints (such as adhering to the one-to-one buyer-product matching limits).

As a consequence of its compact 30B parameter scale and lack of prior task-specific alignment, the untrained base model initially struggles to maintain these protocol boundaries, exhibiting an \textit{Instruction Violation Rate} of 43.3\%. Because the verifiable reward function applies a strict penalty of $\mathcal{R} = -1$ for any rule breach, it provides a direct negative reward signal that prioritizes protocol compliance during the initial phase of optimization. As shown in \Cref{fig:training_curve_instruction_violation_rate}, the violation rate decreases sharply within the first 15 steps, stabilizing at 1.3\% at convergence. This rapid adjustment demonstrates that the agent successfully internalizes the complex market logic and structural constraints, establishing a stable foundation for strategic optimization.

\subsection{Surplus Extraction}
The \textit{Seller Surplus Extraction Ratio} quantifies pricing efficiency by measuring the proportion of the catalog-wide list-price margin captured by the agent. As shown in \Cref{fig:training_curve_seller_surplus_extraction_ratio}, this metric improves steadily over the course of training. This upward trajectory becomes particularly pronounced after the initial constraint-adherence phase stabilizes, demonstrating that once the agent internalizes the basic protocol rules, its optimization focus shifts toward maximizing economic surplus.

The terminal surplus extraction ratio achieved by the trained seller agent exceeds the performance of the generalist baselines, including \texttt{GPT-5.4-high-reasoning} and \texttt{DeepSeek-V4-Pro-thinking}. While these frontier architectures maintain high transaction volumes, they yield lower overall surplus extraction. This pattern indicates that general-purpose models, which are typically aligned for cooperative dialogue, tend to make substantial economic concessions to counterparties rather than optimizing for long-term portfolio value.

In a setting with a one-deal-per-buyer constraint, finalizing an inefficient transaction carries a high opportunity cost by permanently removing that buyer and asset from the market. By learning to combine selective rejections with calculated counter-proposals, the specialized seller agent successfully navigates this trade-off, capturing a higher proportion of the catalog-wide list-price margin.

\begin{figure}[h]
  \begin{minipage}[t]{0.49\textwidth}
    \centering
    \includegraphics[width=\linewidth]{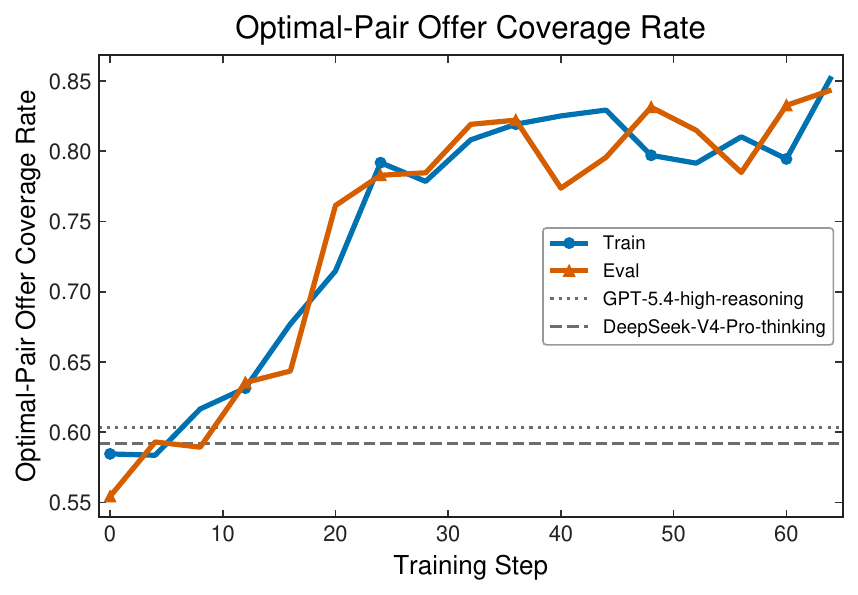}
    \caption{\textbf{Optimal-Pair Offer Coverage Rate.} In-distribution curve across training steps. Experimental setup, series definitions, and benchmarks match Figure~\ref{fig:train_reward}.}
    \label{fig:training_curve_optimal_pair_offer_coverage_rate}
  \end{minipage}\hfill
  \centering
  \begin{minipage}[t]{0.49\textwidth}
    \centering
    \includegraphics[width=\linewidth]{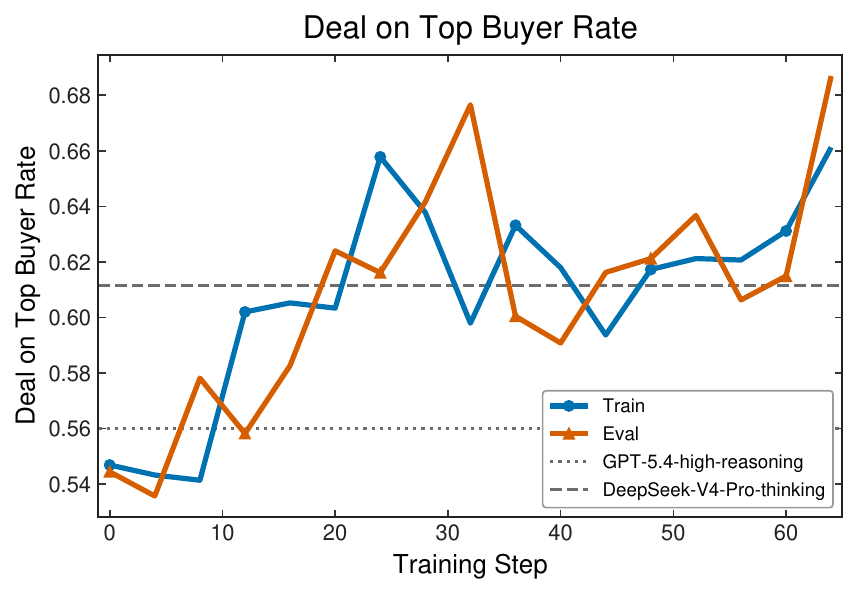}
    \caption{\textbf{Deal on Top Buyer Rate.} In-distribution curve across training steps. Experimental setup, series definitions, and benchmarks match Figure~\ref{fig:train_reward}.}
    \label{fig:training_curve_deal_on_top_buyer_rate}
  \Description{Two line charts against training step. Left, Optimal-Pair Offer Coverage Rate: both series rise substantially above their starting level. Right, Deal on Top Buyer Rate: both series trend upward but stay noisy and never reach 1.0.}
  \end{minipage}
\end{figure}

\subsection{Allocation Quality}
Beyond maximizing immediate transactional surplus, the broader objective of the negotiation agent is to optimize multi-item inventory allocation. This optimization is fundamentally challenged by information asymmetry, as buyers hold distinct, private valuations for substitutable items in the inventory. Consequently, the seller must leverage the dialogue to screen for these valuations and propose high-value matches.

The development of this capability is evaluated using the \textit{Optimal-Pair Offer Coverage Rate} (\Cref{fig:training_curve_optimal_pair_offer_coverage_rate}), which measures how frequently the seller's proposals encompass the ideal allocations determined by the underlying matching oracle (see \Cref{subsec:allocation_metrics}). Over the course of training, this rate increases steadily, exceeding the generalist baseline models in the final stages of training. This upward trend indicates that the agent successfully learns to identify high-value asset pairings through sequential negotiation turns.

This targeted proposal strategy ultimately yields a clear improvement in realized allocation accuracy (\Cref{fig:training_curve_deal_on_top_buyer_rate}). This metric does not stabilize smoothly because buyer valuations are freshly sampled for each episode, introducing inherent variance across training trajectories. Furthermore, because one buyer may hold the highest valuation for multiple items, a rate below $1.0$ does not necessarily imply routing suboptimality under the one-deal-per-buyer constraint. Nonetheless, the performance gains demonstrate that the agent effectively navigates these structural constraints, allowing it to consistently outperform generalist baselines and route assets accurately within the permitted turn limit.

Crucially, these dual improvements emerge without explicit matching incentives. Because the verifiable reward function formalized in \Cref{def:multiitem_reward} measures only the total captured seller surplus and is entirely blind to the hidden oracle matching model, the agent receives no direct reward term penalizing suboptimal target routing or rewarding optimal pairs. The steady improvement in both the \textit{Optimal-Pair Offer Coverage Rate} and the \textit{Deal on Top Buyer Rate} is therefore indirect rather than supervised: better buyer--item assignments raise the terminal seller surplus, so matching is incentivized through the objective even though it is never scored by it. Under the global communication turn limit, the agent discovers that maximizing seller surplus requires accurate preference discovery; it learns to structurally match items to high-valuation buyers because efficient matching drives the maximization of realized seller surplus.

\section{Out-of-Distribution Generalization}
\label{sec:ood}

In this section, we evaluate the performance of the trained seller agent under systematic market shifts to demonstrate that the agent learns generalized negotiation rules rather than overfitting to simulation parameters. Real-world revenue management settings are highly dynamic; a robust agent must maintain stable surplus extraction and matching precision when confronted with unexpected variations in consumer volume, competitive stress, and shifting asset valuations. To establish a rigorous evaluation, we test our trained seller agent across a series of stress-testing scenarios that systematically alter the fundamental layout of the marketplace.

\begin{table}[h]
  \centering
  \small
  \setlength{\tabcolsep}{3pt}
  \renewcommand{\arraystretch}{1.12}
  \begin{tabular}{
  P{0.12\textwidth}
  c
  c
  c
  P{0.14\textwidth}
  P{0.12\textwidth}
  P{0.12\textwidth}
  P{0.14\textwidth}
}
    \toprule
    OOD Environment & Buyers $m$ & Items $n$ & Turns $T$ & Buyer Model & Valuation Structure & Product Category & Core Stress Test \\
    \midrule
    \multicolumn{8}{l}{\textit{In-Distribution Reference}} \\
    \quad & 3 & 3 & 7 & Fixed \text{iter-}60 & Baseline \eqref{eq:budget_gen}  & 12 Train Categories & Training configuration \\
    \hline
    \multicolumn{8}{l}{\textit{Scale Generalization}} \\
    \quad 1. Expanded Standard & 5 & 5 & 12 & \text{iter-}\{50, 60, 70\} & Baseline \eqref{eq:budget_gen} & 12 Train Categories & Scale \& horizon expansion \\
    \quad 2. Expanded Musical & 5 & 5 & 12 &  \text{iter-}\{50, 60, 70\} & Baseline \eqref{eq:budget_gen} & Held-out Musical Instruments & Scale \& unseen price range \\
    \hline
    \multicolumn{8}{l}{\textit{Asymmetric Market Structures}} \\
    \quad 3. Buyer Heavy & 5 & 3 & 12 &  \text{iter-}\{50, 60, 70\} & Baseline \eqref{eq:budget_gen} & 12 Train Categories & High buyer competition \\
    \quad 4. Item Heavy  & 3 & 5 & 7  &  \text{iter-}\{50, 60, 70\} & Baseline \eqref{eq:budget_gen} & 12 Train Categories & High item volume \\
    \hline
    \multicolumn{8}{l}{\textit{Correlated Valuation Distributions}} \\
    \quad 5. Item Correlation  & 5 & 5 & 12 &  \text{iter-}\{50, 60, 70\} & Correlated ($\times \mu_j$)  & 12 Train Categories & Correlated item values \\
    \quad 6. Buyer Correlation & 5 & 5 & 12 &  \text{iter-}\{50, 60, 70\} & Correlated ($\times \gamma_i$) & 12 Train Categories & Correlated buyer valuations \\
    \bottomrule
  \end{tabular}
  \caption{\textbf{Experimental evaluation matrix.} The top row establishes the baseline in-distribution reference configuration. Environments 1--6 represent out-of-distribution (OOD) configurations; all of these environments share the dynamic buyer checkpoint pool $\{\text{iter-}50, \text{iter-}60, \text{iter-}70\}$, while each individual environment introduces a unique structural configuration as specified in the table. For the correlated settings (Environments 5 and 6), the respective valuation multipliers follow $\mu_j, \gamma_i \sim \mathcal{U}[0.5, 2]$.}
  \label{tab:ood_experimental_matrix}
\end{table}

\subsection{Out-of-Distribution Environment}
\label{subsec:ood_environment}

To evaluate whether the seller's learned allocation strategy generalizes beyond its training distribution, we construct six OOD environments by perturbing one or more foundational dimensions of the marketplace. These structural shifts prevent the seller from relying on a memorized valuation distribution, the behavior of a specific buyer model, or a known product catalog and price range. In what follows, we detail each of these distinct dimensions of perturbation.

\paragraph{Buyer-checkpoint variation.} In-distribution training and evaluation fix the buyer agent at the $\text{iter-}60$ checkpoint from \citet{liu2026instructing}. Across all six OOD environments, the buyer agent profile is sampled dynamically per buyer slot per episode, drawing uniformly from a three-checkpoint pool $\{\text{iter-}50,\ \text{iter-}60,\ \text{iter-}70\}$ of the buyer training trajectory. Because every OOD environment draws from this pool, each is a compound shift that alters counterpart behavior in addition to the structural change named in its row. These checkpoints span distinct phases of strategic maturity, with $\text{iter-}70$ representing a more optimized and competitive counterpart than $\text{iter-}60$. These models exhibit divergent tactical behaviors, utilizing varying opening price proposals relative to their private valuations and thereby complicating the seller's strategic probing \citep{liu2026instructing}. Sampling diverse buyer checkpoints ensures the seller's performance is not an artifact of overfitted text coordination patterns memorized against a single fixed counterpart.

\paragraph{Market scale and asymmetry.} While the in-distribution setup restricts the market to a symmetrical configuration of 3 buyers and 3 items over a horizon of 7 total turns, the OOD configurations alter these scale dimensions. Environments 1, 2, 5, and 6 expand the market to a larger layout of 5 buyers and 5 items over an extended horizon of 12 total turns. We also introduce structural supply-demand imbalances by varying the buyer-to-item ratio: Environment 3 evaluates a buyer-heavy market (5 buyers to 3 items over 12 turns) to test behavior under intense buyer competition, whereas Environment 4 evaluates an item-heavy layout (3 buyers to 5 items) characterized by high item volume over a tight 7-turn boundary.

\paragraph{Structured valuation.} The in-distribution valuation matrix draws every entry independently from the same uniform interval (\Cref{subsec:market_settings}). 
The correlated valuation OOD environments introduce structural correlation across items and buyers by scaling the baseline valuation:
$$
\mathcal{B}_{i,j} = 
\begin{cases} 
  r_{i,j} \cdot \mu_j \cdot P_{\text{list},j} & \text{for Environment 5 (Item Correlation),} \\
  r_{i,j} \cdot \gamma_i \cdot P_{\text{list},j} & \text{for Environment 6 (Buyer Correlation),}
\end{cases}
$$
where $r_{i,j} \sim \mathcal{U}[0.40, 1.15]$. Here, $\mu_j$ is a \emph{per-item valuation multiplier} that rescales each item's baseline value across the entire buyer pool, making some items universally more valuable or discounted. Similarly, $\gamma_i$ is a \emph{per-buyer generosity multiplier} that systematically shifts each buyer's overall willingness to pay up or down across all items in the catalog.

During in-distribution training and evaluation, these multipliers are inactive and fixed at $1$, ensuring that all item and buyer valuations are drawn independently from the same distribution. In Env.~5 (Item Correlation), $\mu_j \sim \mathcal{U}[0.5, 2]$ is drawn once per item while $\gamma_i=1$; in Env.~6 (Buyer Correlation), $\gamma_i \sim \mathcal{U}[0.5, 2]$ is drawn once per buyer while $\mu_j=1$. These shifts introduce item-level or buyer-level correlation structure in $\mathcal{B}$ that the seller agent never encountered during training.

\paragraph{Held-out product category.} In-distribution configurations draw item catalogs from the $12$ train/test categories outlined in \Cref{subsec:market_settings}. The corresponding OOD environment (Env.~2 in \Cref{tab:ood_experimental_matrix}) instead draws every catalog from a $13$th category, \textit{Musical Instruments} (\Cref{fig:list_price_distribution}), which was held out completely from all training phases. This category introduces a distinct list-price distribution; because every valuation in an episode is scaled to the item's list price ($\mathcal{B}_{i,j} = r_{i,j} P_{\text{list},j}$), this distribution shift moves the valuations and costs into a range unseen during training, verifying whether the pricing strategy transfers effectively beyond the price scales of the training catalog.

Rather than evaluating these environment perturbations in isolation, we systematically cross-combine these primitive dimensions to construct six distinct out-of-distribution testing environments. As summarized in \Cref{tab:ood_experimental_matrix}, these configurations are grouped into three core stress-testing families:
\begin{itemize}
    \item \textbf{Scale Generalization:} Comprising \textit{Env.~1: Expanded Standard}, which evaluates scale and horizon expansion to a symmetrical market of 5 buyers and 5 items with an expanded $T=12$ communication budget, and \textit{Env.~2: Expanded Musical}, which overlays a scale and unseen price range from the held-out musical instruments catalog onto this expanded horizon.
    \item \textbf{Asymmetric Market Structures:} Stress-testing structural competition imbalances by varying the buyer-to-item ratio, featuring \textit{Env.~3: Buyer Heavy}, which evaluates performance under high buyer competition in a market of 5 buyers and 3 items over $T=12$ turns, and \textit{Env.~4: Item Heavy}, which evaluates performance under high item volume in a market of 3 buyers and 5 items restricted to a tight $T=7$ turn boundary.
    \item \textbf{Correlated Valuation Distributions:} Introducing item-level and buyer-level correlation structure in valuations within the expanded marketplace of 5 buyers and 5 items, comprising \textit{Env.~5: Item Correlation}, which activates the item valuation multiplier $\mu_j$ to yield correlated item values, and \textit{Env.~6: Buyer Correlation}, which activates the buyer generosity multiplier $\gamma_i$ to yield correlated buyer valuations.
\end{itemize}

\subsection{Out-of-Distribution Results}
\label{subsec:ood_results}

\begin{figure}[ht]
  \centering
  \includegraphics[width=0.93\linewidth]{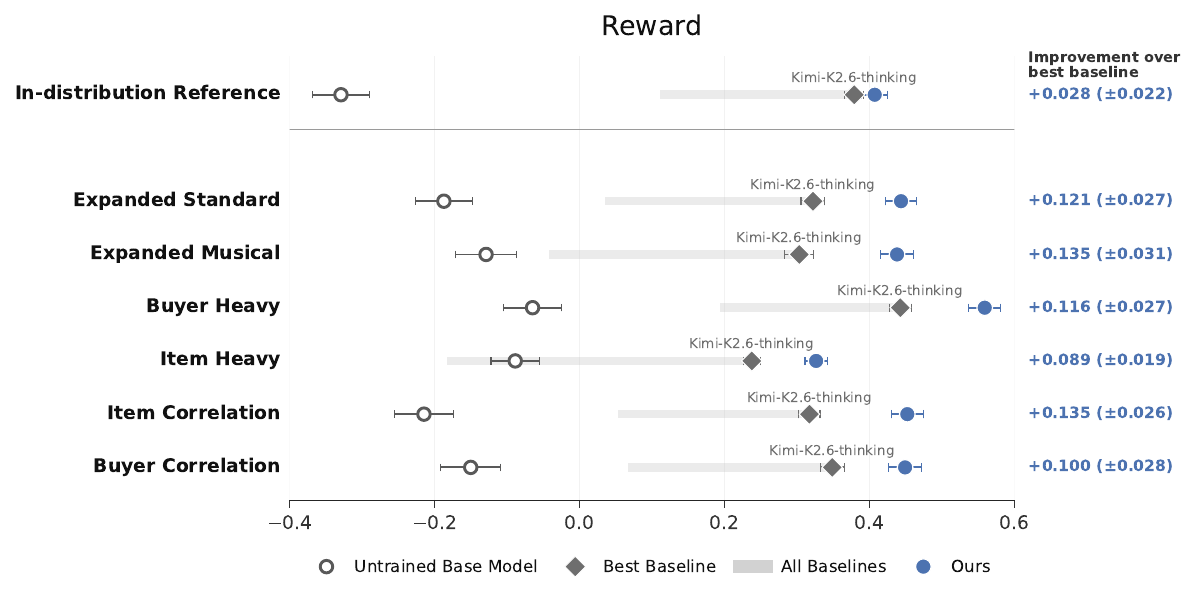}
  \caption{\textbf{Out-of-distribution generalization --- Reward.} Mean episode reward $\mathcal{R}$ of the RL-trained
  seller against the entire frontier-baseline field, one row per market distribution (the in-distribution
  reference plus six out-of-distribution shifts). Grey band: minimum-to-maximum range of the
  eight baselines; grey diamond: best baseline (named); blue circle: trained seller; hollow circle: untrained
  base model. The right column reports the improvement over the best baseline. Error bars denote the standard
  error.}
\label{fig:ood_reward}
\Description{A horizontal forest plot with seven rows: an in-distribution reference row, separated by a rule, above six out-of-distribution rows (Expanded Standard, Expanded Musical, Buyer Heavy, Item Heavy, Item Correlation, Buyer Correlation). Each row plots mean episode reward on a shared axis from $-0.4$ to $0.6$: a hollow circle for the untrained base model, a grey band spanning the minimum-to-maximum range of the eight baselines, a grey diamond for the best baseline, and a filled blue circle for the trained seller, all with standard-error bars. The untrained model is negative in every row. The trained seller lies to the right of the best baseline in all seven rows, and the best baseline is Kimi-K2.6-thinking throughout. The gap is narrowest in the in-distribution row and wider in each out-of-distribution row; the right-hand column lists it as $+0.028$ in-distribution and between $+0.089$ and $+0.135$ across the six shifts.}
  \label{fig:ood_forest_reward}

\end{figure}

\begin{figure}[h]
  \centering
  \includegraphics[width=0.93\linewidth]{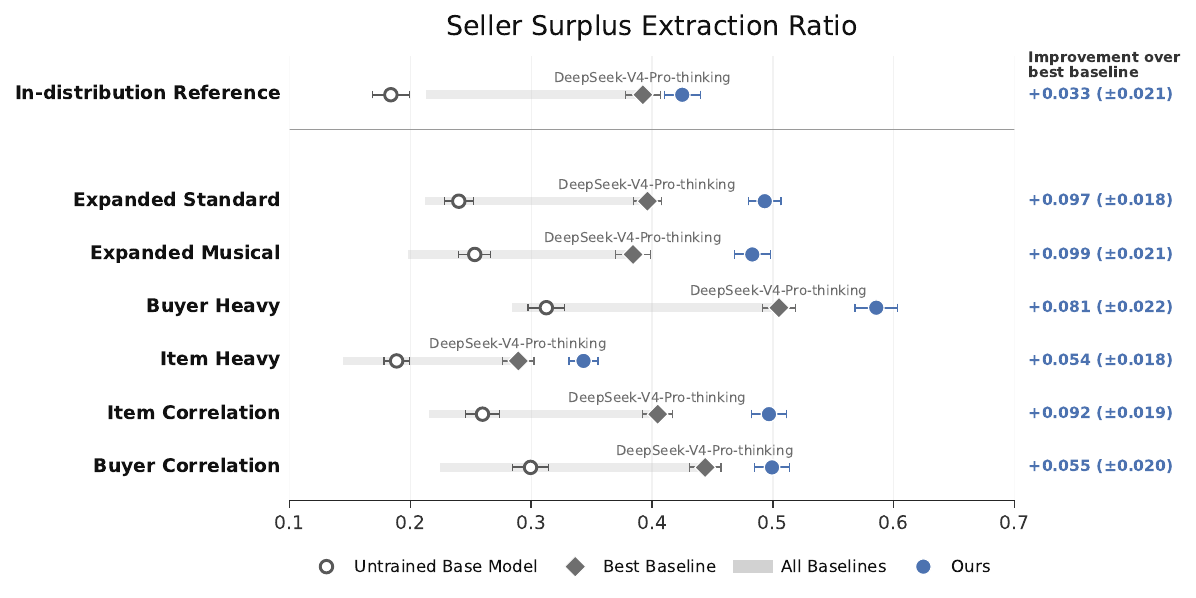}
  \caption{\textbf{Out-of-distribution generalization --- Seller Surplus Extraction Ratio.} The seller's
  share of the catalog-wide list-price margin captured (aggregated over valid episodes), with the same rows
  and marks as \Cref{fig:ood_forest_reward}. Error bars denote the standard error.}
  \label{fig:ood_forest_surplus}
  \Description{A horizontal forest plot with the same seven rows and the same marks as the reward figure, plotting the Seller Surplus Extraction Ratio on an axis from $0.1$ to $0.7$. The untrained base model sits between roughly $0.18$ and $0.35$, the baseline band spans the middle of the axis, and the trained seller lies to the right of the best baseline in every row. DeepSeek-V4-Pro-thinking is the best baseline in all seven rows. The right-hand column reports the improvement as $+0.033$ in-distribution and between $+0.054$ and $+0.099$ across the six out-of-distribution shifts, so the margin is again smallest in-distribution.}

\end{figure}

\begin{figure}[h]
  \centering
  \includegraphics[width=0.93\linewidth]{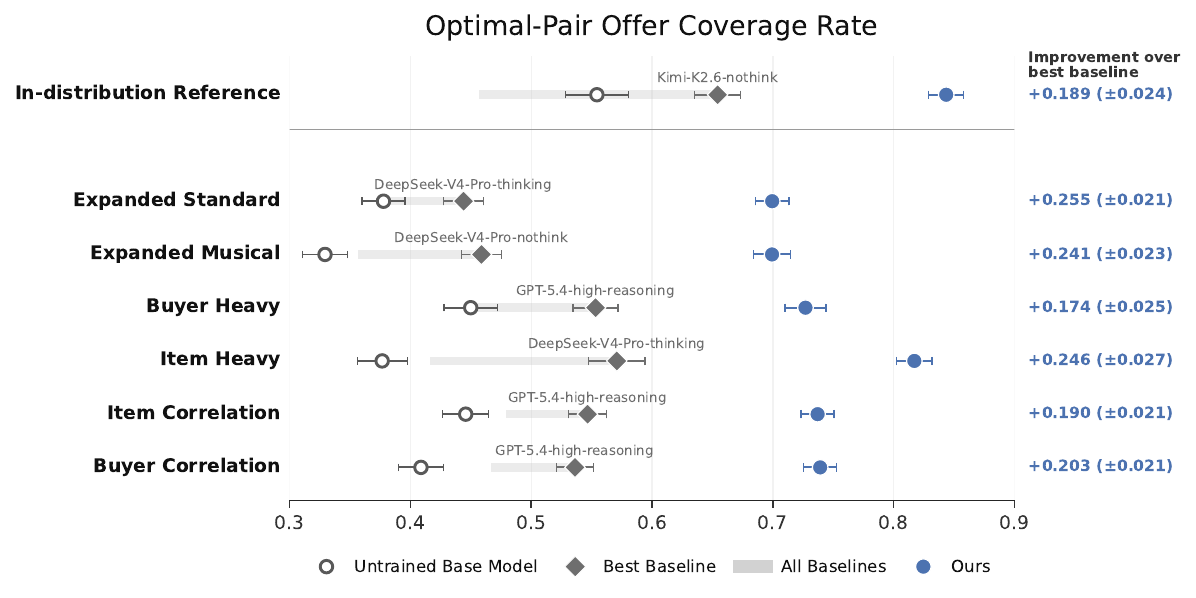}
  \caption{\textbf{Out-of-distribution generalization --- Optimal-Pair Offer Coverage Rate.} How often the
  seller proposes the oracle-optimal buyer--item pairings (aggregated over valid episodes), with the same rows
  and marks as \Cref{fig:ood_forest_reward}. Error bars denote the standard error.}
  \label{fig:ood_forest_coverage}
  \Description{A horizontal forest plot with the same seven rows and the same marks as the reward figure, plotting the Optimal-Pair Offer Coverage Rate on an axis from $0.3$ to $0.9$. The trained seller lies well to the right of the best baseline in every row, and the separation is visibly larger than in the reward and surplus figures. Unlike those two, the best baseline changes across rows: Kimi-K2.6-nothink in-distribution, DeepSeek-V4-Pro-thinking, DeepSeek-V4-Pro-nothink, and GPT-5.4-high-reasoning in three rows. The right-hand column reports improvements between $+0.174$ and $+0.255$, with $+0.189$ in-distribution.}

\end{figure}

We evaluate the seller agent across the six distinct out-of-distribution environments detailed in \Cref{subsec:ood_environment}. These configurations stress-test the model's capacity to generalize under compound shifts that combine buyer-checkpoint variation with changes in market scale, supply--demand structure, valuation structure, and a held-out product category and price regime. To maintain a concise presentation, the detailed performance tables are provided in Appendix~\ref{app:extended_benchmarks}. A unified overview of the key performance metrics across all configurations is presented in \Cref{fig:ood_forest_reward} to \Cref{fig:ood_forest_coverage}.

As illustrated across these three metrics, the trained seller model outperforms the highest-scoring baseline model in all six out-of-distribution environments in terms of reward, surplus extraction, and offer coverage. In the first panel, the reward margin over the top baseline increases across all out-of-distribution environments, ranging from $+0.089$ under high item volume (\textit{Env.~4: Item Heavy}) up to $+0.135$ under an unseen product category (\textit{Env.~2: Expanded Musical}) and under correlated item values (\textit{Env.~5: Item Correlation}). Compared to the in-distribution reference margin ($+0.028$), the trained agent maintains a positive advantage over the strongest baseline within every tested environment. Because the reward normalization has a different attainable ceiling under each market structure, absolute reward levels are not directly comparable across environments; comparisons should therefore be made within each environment against the same baselines.

We emphasize that out-of-distribution refers to a shift from the trained seller’s training environment, and such shifts can simultaneously expand the available economic opportunities and increase the complexity of coordinating decisions across buyers and items. For example, in the buyer-heavy market, a larger buyer-to-item ratio creates more potential high-value matches for a seller with an effective negotiation and matching strategy, while also increasing the number of buyers that must be screened and prioritized and the buyer-item assignments that must be coordinated under the communication budget. The larger within-environment performance gap can therefore arise because the trained seller is better able to exploit the additional opportunities while maintaining effective coordination, whereas the general-purpose baselines may benefit less as the expanded market places greater demands on their ability to screen buyers, allocate attention, and coordinate buyer-item assignments.

\Cref{fig:ood_forest_surplus} and \Cref{fig:ood_forest_coverage} show the corresponding trends for the underlying operational metrics. The trained seller model maintains the highest \textit{Seller Surplus Extraction Ratio} and \textit{Optimal-Pair Offer Coverage Rate} across all six testing environments. For example, in the buyer-heavy configuration (\textit{Env.~3: Buyer Heavy}) and under the buyer valuation shift (\textit{Env.~6: Buyer Correlation}), the trained model achieves a surplus extraction ratio near or above $50\%$ and an optimal allocation coverage exceeding $70\%$, outperforming benchmark models with larger parameter counts. Together, these results suggest that the learned information-gathering, negotiation, and allocation behaviors remain effective across changing market structures, highlighting the value of transferable strategic behaviors for language agents operating in dynamic multi-agent environments.

\section{Discussion}
\label{sec:discussion}

Our findings reveal a tension between conversational agreeableness and explicit economic optimization in resource-constrained markets. The general-purpose baselines we evaluate tend to favor early agreement and higher deal volume at the expense of portfolio-wide seller surplus. This pattern is consistent with prior reports of cooperative or agreeable behavior in language models \citep{ perez2023discovering, sharma2024towards}. In our multi-product setting, a premature, low-margin agreement removes both the buyer and the item from future consideration and may foreclose more profitable buyer-item matches. In contrast, the trained seller develops valuation-screening and buyer-item matching behaviors from a surplus-based verifiable reward. Although the reward does not directly encode a matching heuristic, the agent improves its coverage of high-value buyer-item pairs and more often matches products to higher-valuation buyers.

More broadly, our results illustrate the potential of task-specific post-training for autonomous language agents with persistent objectives. Across the evaluated environments, the trained 30B model outperforms general-purpose frontier models, including models with more than one trillion parameters, on seller surplus extraction and buyer-item allocation quality. Its out-of-distribution performance further shows that the learned behavior transfers across changes in buyer behavior, market scale, supply-demand balance, valuation correlations, and product price ranges. These findings suggest that verifiable-reward training can help language agents coordinate information gathering and allocation across multiple interdependent interactions.

\section{Conclusion}
\label{sec:conclusion}

In this paper, we study multi-item concurrent bargaining under information asymmetry and a global communication budget, where a single seller negotiates a portfolio of substitutable products with multiple buyers who can each purchase at most one item. Our evaluations show that general-purpose LLMs can favor early agreements and high deal volume over portfolio-wide seller surplus. We train the seller using Reinforcement Learning from Verifiable Rewards (RLVR) with a verifiable reward based on seller surplus. Without predefined matching heuristics, training improves buyer valuation screening, buyer-item matching, and seller surplus extraction. The resulting 30B seller matches or outperforms the evaluated general-purpose frontier models, including models with more than one trillion parameters, on both seller surplus extraction and buyer-item allocation quality. The trained seller agent also generalizes across out-of-distribution environments involving new buyer behaviors, market scales, supply-demand imbalances, valuation correlations, and product price ranges. More broadly, these findings highlight the potential of verifiable-reward training for developing language agents capable of long-horizon strategic decision-making in complex, partially observable, multi-agent environments.

\section*{AI Use Statement}

Generative AI tools were used for proofreading and grammar correction and to assist in generating the visual pipeline diagrams in Figures~\ref{fig:market_framework} and \ref{fig:evaluation_framework}. All AI-assisted content was reviewed by the authors, who take responsibility for the final content.

\begin{acks}
This research was supported in part by NSF Awards IIS-2312865 and OAC-2311521, and by compute resources provided through a Tinker Research Grant from Thinking Machines Lab, awarded to Shuze Daniel Liu and Claire Chen.
\end{acks}

\printbibliography

\newpage
\appendix

\section{Notation}
\label{app:notation}

\Cref{tab:notation} collects the notation used throughout the paper, grouped by the part of the framework each symbol belongs to. Full definitions appear at each symbol's point of first use in the main text.

\begin{table}[h]
  \centering
  \footnotesize
  \setlength{\tabcolsep}{3pt}
  \begin{tabular}{@{}lp{2.9cm}p{7.8cm}@{}}
    \toprule
    Symbol & Name & Description \\
    \midrule
    \multicolumn{3}{@{}l@{}}{\textit{Episode parameters and indices (\S\ref{sec:methodology})}} \\
    \quad $n$, $m$ & Item and buyer counts & Number of items and buyer slots per episode. \\
    \quad $j$, $i$ & Item and buyer indices & Item index $j \in \{1, \ldots, n\}$; buyer index $i \in \{1, \ldots, m\}$. \\
    \quad $t$ & Turn index & Seller turn index $t$. \\
    \quad $T$, $K$ & Communication budget and turn limit & Global communication budget ($T = 7$ in training) and per-buyer turn limit ($K = 3$); $T < mK$. \\
    \addlinespace
    \multicolumn{3}{@{}l@{}}{\textit{Prices, costs, and valuations (\S\ref{sec:methodology}, \S\ref{subsec:market_settings}, \S\ref{sec:ood})}} \\
    \quad $P_{\text{list},j}$ & List price & Public list price of item $j$. \\
    \quad $\mathcal{C}_j$ & Seller's cost & Seller's private cost for item $j$; always strictly below $P_{\text{list},j}$. \\
    \quad $\mathcal{B}_{i,j}$, $\mathcal{B}$ & Buyer valuation; valuation matrix & Buyer $i$'s private valuation for item $j$; the full $m \times n$ matrix, never observed by the seller. \\
    \quad $P_{\text{deal},j}$ & Deal price & Agreed price when item $j$ sells; may exceed the list price. \\
    \quad $P_{\text{final},j}$ & Final price & $P_{\text{final},j} := \min(P_{\text{deal},j},\, P_{\text{list},j})$; enters the reward and all surplus metrics. \\
    \quad $\mathcal{U}[a, b]$ & Uniform distribution & Uniform distribution on the interval $[a, b]$. \\
    \quad $r_{i,j}$ & Buyer valuation ratio & $r_{i,j} \sim \mathcal{U}[0.40, 1.15]$, i.i.d.\ across all $(i, j)$ pairs. \\
    \quad $\mu_j$, $\gamma_i$ & Valuation multipliers & Per-item and per-buyer valuation multipliers; both equal $1$ in-distribution and in Envs.~1--4. Env.~5 draws $\mu_j \sim \mathcal{U}[0.5, 2]$ with $\gamma_i = 1$, while Env.~6 draws $\gamma_i \sim \mathcal{U}[0.5, 2]$ with $\mu_j = 1$. \\
    \addlinespace
    \multicolumn{3}{@{}l@{}}{\textit{Reward and seller metrics (\S\ref{subsec:reward}, \Cref{sec:evaluation})}} \\
    \quad $\mathcal{R}$ & Terminal reward & $-1$ on a violation episode, in $[0, 1]$ on a valid episode (equation~\eqref{def:multiitem_reward}). \\
    \quad $S$ & Sold-item set & An episode's set of items that closed a deal; the pool for the Deal on Top Buyer Rate. \\
    \addlinespace
    \multicolumn{3}{@{}l@{}}{\textit{Allocation oracle (\Cref{sec:evaluation})}} \\
    \quad $\mathcal{M}^\star$ & Optimal-matching set & Set of all surplus-maximizing matchings under list-clipped surplus and the one-deal-per-buyer constraint. \\
    \quad $A$, $A(j)$ & Realized allocation & $A(j)$ is the buyer who closed item $j$; $\varnothing$ if item $j$ is unsold. \\
    \bottomrule
  \end{tabular}
  \caption{Notation used throughout the paper, grouped by the part of the framework each symbol belongs to.}
  \label{tab:notation}
\end{table}

\section{Formal Definitions of Evaluation Metrics}
\label{app:backup_metrics}

This section provides the full mathematical formalizations of the allocation metrics introduced in \Cref{sec:evaluation}, anchored on the centralized oracle matching model defined below. Every metric is aggregated over the valid episodes the seller completes without a violation, using the notation of \Cref{tab:notation}.

\subsection{Oracle Matching Model}

The allocation metrics compare the seller's realized allocation against the surplus-maximizing assignment of items to buyers. We define the oracle optimal-matching set
\[
  \mathcal{M}^\star \;=\; \argmax_{M} \sum_{(j,i) \in M} \big(\min(\mathcal{B}_{i,j},\, P_{\text{list},j}) - \mathcal{C}_j\big)_+, \qquad (x)_+ = \max(0, x),
\]
where $M$ ranges over matchings that assign each item to at most one buyer and each buyer to at most one item (a partial, injective map from items to buyers, with $M(j) = \varnothing$ marking item $j$ unpaired). Each pair's surplus is clipped at the list price, so dollars a buyer would pay above the list price do not count; $\mathcal{M}^\star$ is typically non-singleton, and the empty matching is optimal exactly when no positive clipped surplus exists. We write $A$ for the seller's realized allocation, with $A(j)$ the buyer who closed item $j$ ($\varnothing$ if item $j$ is unsold).

\subsection{Primary Allocation Metrics}

\paragraph{Optimal-Pair Offer Coverage Rate.} The fraction of oracle-optimal buyer--item pairs that the seller targets with a \act{SELL} or \act{DEAL} action at least once during the episode, pooled across the valid episodes that admit a non-empty optimal matching:
\[
  \text{Optimal-Pair Offer Coverage Rate} \;=\; \frac{\text{Total Optimal Pairs Offered}}{\text{Total Pairs in Optimal Matching}}.
\]
For each episode we select the optimal matching $M \in \mathcal{M}^\star$ that overlaps most with the seller's proposals; a pair $(i,j) \in M$ counts as offered whenever the seller names item $j$ to buyer $i$ in at least one \act{SELL} or \act{DEAL} action, regardless of whether a deal ultimately closes. The metric therefore scores matching \emph{intent} rather than final execution.

\paragraph{Deal on Top Buyer Rate.} The fraction of successfully sold items that cleared to the buyer holding the highest private valuation for that item, pooled across all sold items of the valid episodes:
\[
  \text{Deal on Top Buyer Rate} \;=\; \frac{\sum_{\text{ep}} \sum_{j \in S} \mathds{1}\big[\, A(j) \in \argmax_i \mathcal{B}_{i,j} \,\big]}{\sum_{\text{ep}} |S|},
\]
where $S$ is the set of items sold in a given episode and $\mathds{1}[\cdot]$ is the indicator function. Ties are permissive: any buyer attaining $\argmax_i \mathcal{B}_{i,j}$ satisfies the indicator. Because one buyer may hold the highest valuation for several substitutable items yet can close at most one under the one-deal-per-buyer constraint, a value below $1$ does not by itself imply a routing error.

\clearpage

\section{In-Distribution Benchmarking Figure}
\label{sec:indist_benchmark_bars}
\begin{figure}[h]
  \centering
  \begin{subfigure}[t]{0.49\textwidth}
    \centering
    \includegraphics[width=\linewidth]{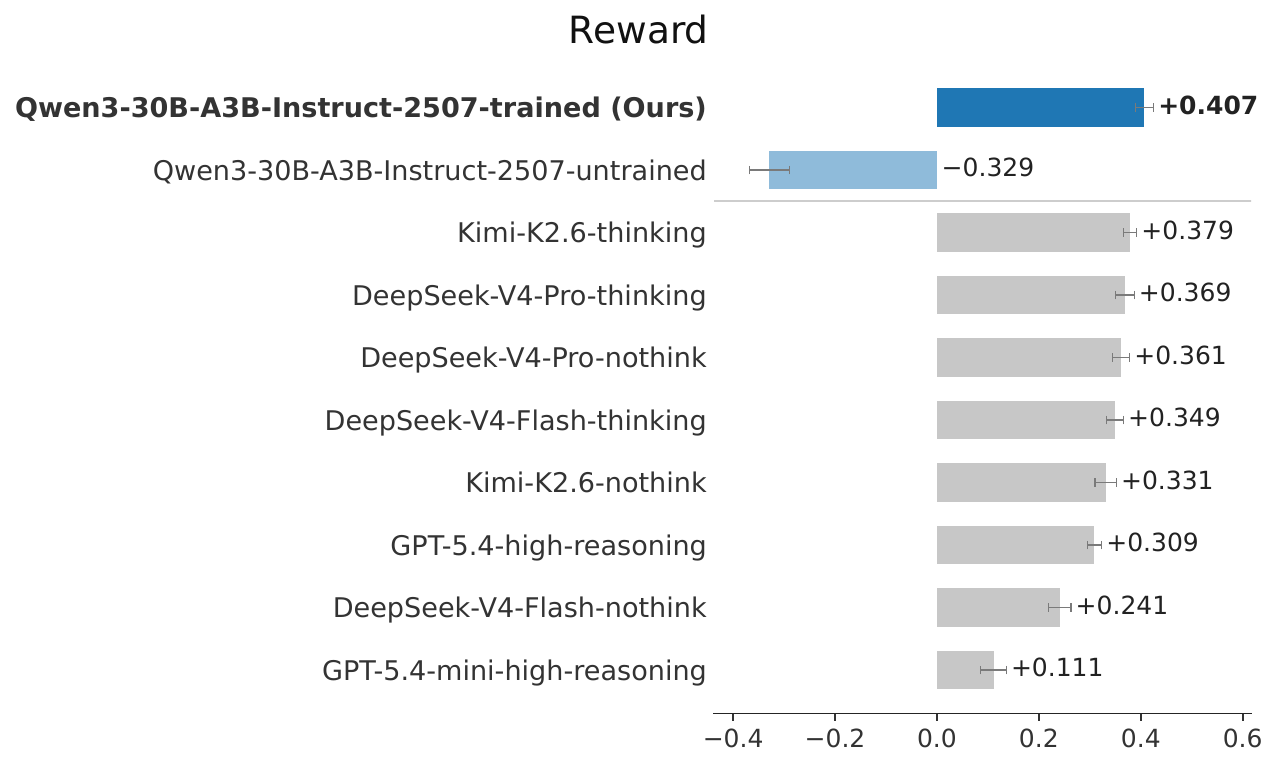}
    \caption{Reward}
    \label{fig:indist_benchmark_bars_reward}
  \end{subfigure}\hfill
  \begin{subfigure}[t]{0.49\textwidth}
    \centering
    \includegraphics[width=\linewidth]{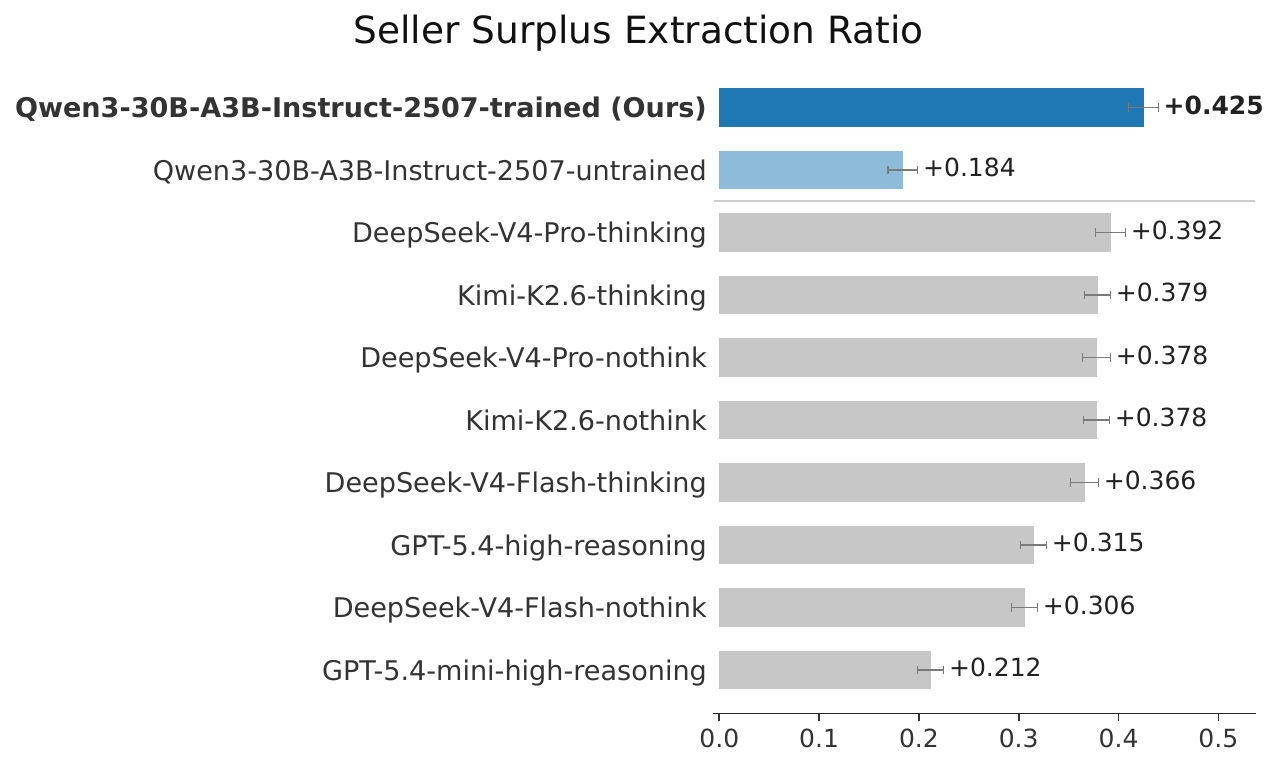}
    \caption{Seller Surplus Extraction Ratio}
    \label{fig:indist_benchmark_bars_surplus}
  \end{subfigure}

  \medskip
  \begin{subfigure}[t]{0.49\textwidth}
    \centering
    \includegraphics[width=\linewidth]{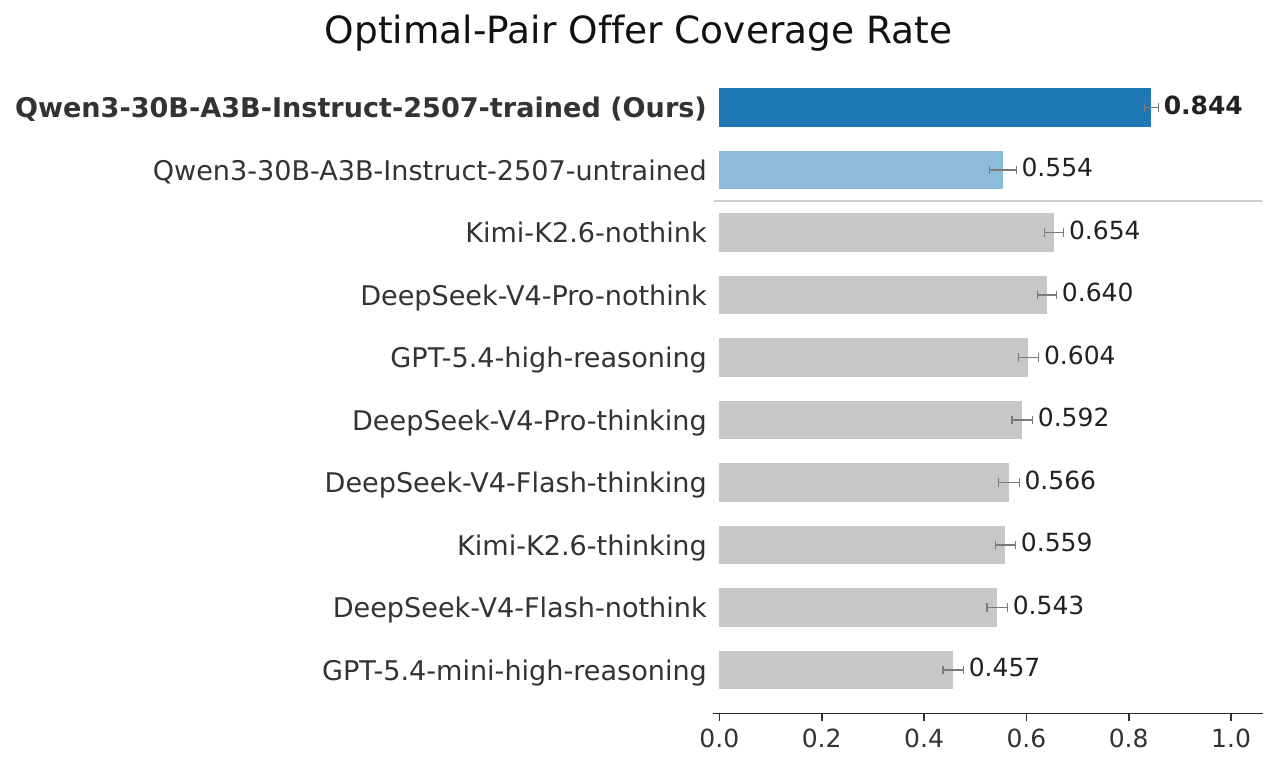}
    \caption{Optimal-Pair Offer Coverage Rate}
    \label{fig:indist_benchmark_bars_coverage}
  \end{subfigure}\hfill
  \begin{subfigure}[t]{0.49\textwidth}
    \centering
    \includegraphics[width=\linewidth]{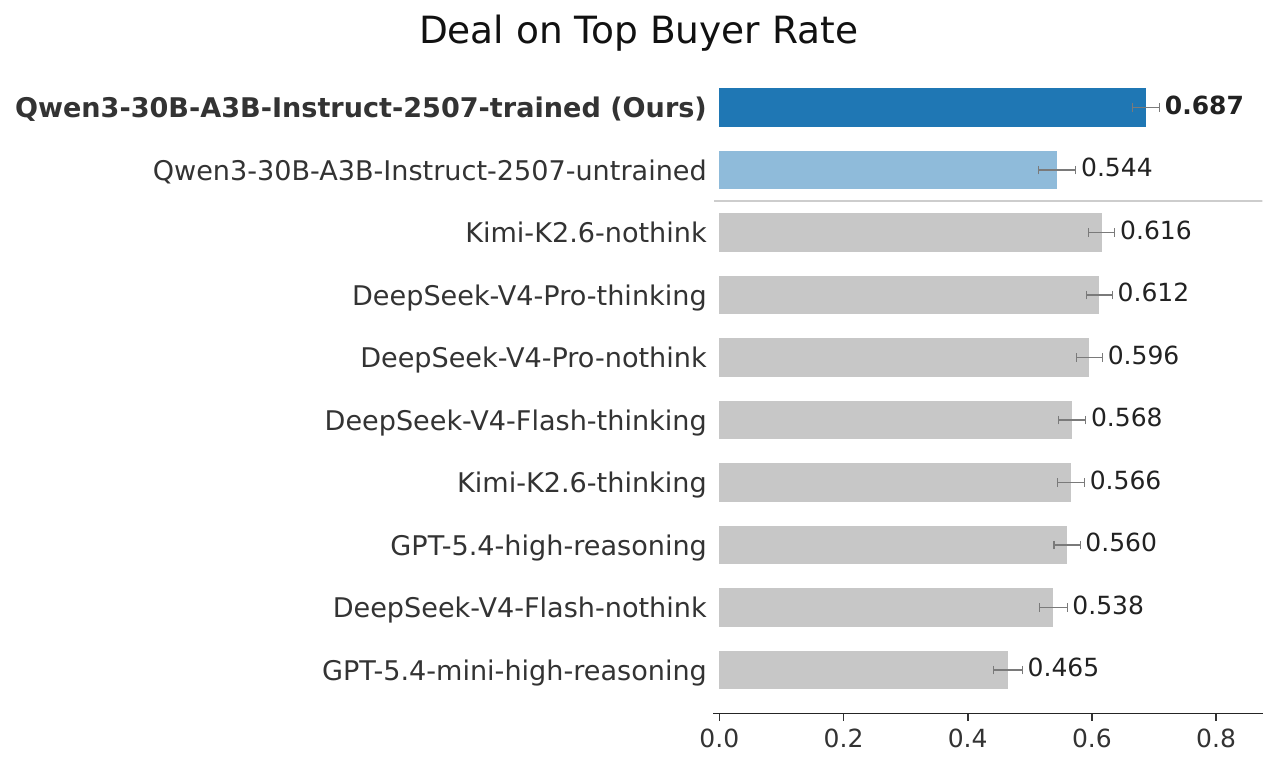}
    \caption{Deal on Top Buyer Rate}
    \label{fig:indist_benchmark_bars_deal_on_top}
  \end{subfigure}
  \caption{\textbf{In-distribution performance benchmarking ($3$ buyers, $3$ items, $T=7$).} Per-metric seller rankings on byte-identical in-distribution held-out instances. The trained model (\textbf{Ours}) and its untrained base appear above the rule, the frontier baselines below. Error bars show the standard error. Values match \Cref{tab:indist3_main}.}
  \label{fig:indist_benchmark_bars}
  \Description{Four horizontal bar charts, one per metric: Reward, Seller Surplus Extraction Ratio, Optimal-Pair Offer Coverage Rate and Deal on Top Buyer Rate. Each chart is a single series with one bar per model, sorted by value. In every panel the trained seller and its untrained base appear above a rule and the eight frontier baselines below it. The trained seller has the longest bar on all four metrics, while the untrained base is the shortest on Reward.}
\end{figure}

This section provides a visual reference complementing the primary empirical benchmarking results discussed in the main text. To clearly illustrate the performance margins across different model architectures, Figure~\ref{fig:indist_benchmark_bars} plots the relative rankings and corresponding standard errors for the entire ten-model panel. The profiles shown below are computed over identical, held-out instances, directly mirroring the exact numerical values reported in Table~\ref{tab:indist3_main}.

\section{Training Implementations}
\label{app:implementation}

This appendix collects the implementation details required to reproduce the training
and evaluation pipeline. \Cref{app:model_details} specifies the seller and buyer models together
with the checkpoints they are initialized from, \Cref{subsec:hyperparams} lists the optimization
and rollout hyperparameters used for the RLVR run and for evaluation, and \Cref{subsec:ood_setup}
gives the configuration of the out-of-distribution environments.

\subsection{Multi-Item Seller Setting --- Model Specifications}
\label{app:model_details}

The only learning agent in this paper is the \emph{seller}; all buyer checkpoints remain frozen. This one-sided training setup differs from equilibrium-learning formulations in which multiple strategic responses are optimized jointly or iteratively, including offline zero-sum, general-sum, and potential-game settings
\citep{zhang2026pessimism, chen2026pessimism, chen2026fast, chen2026offline}. Both the seller and buyers share the same base model, \texttt{Qwen3-30B-A3B-Instruct-2507} \citep{qwen3technicalreport}. During training and in-distribution evaluation, buyers use the iteration-$60$ checkpoint from a bilateral (one-on-one) negotiation RLVR run \citep{liu2026instructing}; OOD buyer checkpoints are specified in Table~\ref{tab:multiitem_models}.

\begin{table}[ht]
\centering
\small
\begin{tabularx}{\textwidth}{@{}>{\raggedright\arraybackslash}p{0.18\textwidth} p{0.34\textwidth} X@{}}
\toprule
Role & Base model & Provenance / checkpoint \\
\midrule

Seller Starting Checkpoint
  & \makecell[l]{\texttt{Qwen/Qwen3-30B-A3B-}\\
                 \texttt{Instruct-2507} \citep{qwen3technicalreport}}
  & Original checkpoint from \citep{qwen3technicalreport}; fine-tuned as the seller. \\

Buyers in Training
  & \makecell[l]{\texttt{Qwen/Qwen3-30B-A3B-}\\
                 \texttt{Instruct-2507}}
  & Frozen buyer checkpoint iteration $60$ (iter-$60$), trained via RLVR on a bilateral negotiation task \citep{liu2026instructing}. \\

Buyers in Evaluation
  & \makecell[l]{\texttt{Qwen/Qwen3-30B-A3B-}\\
                 \texttt{Instruct-2507}}
  & In-distribution evaluation: fixed iter-$60$. Out-of-distribution evaluation: per-buyer uniform over the pool $\{\text{iter-}50,\ \text{iter-}60,\ \text{iter-}70\}$. \\

\bottomrule
\end{tabularx}

\caption{Models used by the multi-item-seller pipeline. The seller is the trained agent; the buyers are frozen checkpoints from \citet{liu2026instructing}.}
\label{tab:multiitem_models}
\end{table}

For benchmark comparisons in the held-out evaluation, the trained seller is contrasted against nine other seller models listed in Table~\ref{tab:model_sources_full}.

\begin{table}[ht]
  \centering
  \small
  \begin{tabular}{lll}
    \toprule
    Model Name  & Parameter Count & Reference \\
    \midrule
    Qwen3-30B-A3B-Instruct-2507-trained (\textbf{Ours}) & 30B & This Work \\
    Qwen3-30B-A3B-Instruct-2507-untrained & 30B & \citet{qwen3technicalreport} \\
    GPT-5.4-high-reasoning & closed-source & \citet{singh2025openai} \\
    GPT-5.4-mini-high-reasoning & closed-source & \citet{singh2025openai} \\
    DeepSeek-V4-Pro-thinking / nothink & 1.6T & \citet{deepseekai2026deepseekv4} \\
    DeepSeek-V4-Flash-thinking / nothink & 284B & \citet{deepseekai2026deepseekv4} \\
    Kimi-K2.6-thinking / nothink & 1T & \citet{team2025kimi} \\
    \bottomrule
  \end{tabular}
  \caption{Seller models evaluated in the held-out benchmark, with total parameter counts. Three open-weight models are run in both a thinking and a no-thinking (\texttt{nothink}) mode, giving a $10$-model panel. DeepSeek-V4-Pro is a mixture-of-experts model with $1.6$T total / $49$B active parameters; DeepSeek-V4-Flash has $284$B total / $13$B active parameters. The buyer is held fixed across the panel (Table~\ref{tab:multiitem_models}).}
  \label{tab:model_sources_full}
\end{table}

\subsection{Training and Evaluation Hyperparameters}
\label{subsec:hyperparams}

The seller is fine-tuned with reinforcement learning via Tinker \citep{tml2025tinker}. The run reported in this paper uses the CISPO loss \citep{chen2025minimax} with clipping thresholds $(0.8, 6.0)$. One \texttt{step} denotes one optimization step over a sampled batch of episodes. Tables~\ref{tab:multiitem_train_hyperparams}, \ref{tab:multiitem_episode_hyperparams}, and~\ref{tab:multiitem_eval_hyperparams} list the hyperparameters that control the configuration.
We did not conduct a systematic hyperparameter search; only the configuration reported in these tables was considered. All settings were fixed before the final experiments, and neither the held-out test set nor the OOD evaluation sets were used for hyperparameter selection.

\begin{table}[h]
  \centering
  \small
  \setlength{\tabcolsep}{4pt}
  \begin{tabular}{@{}ll@{}}
    \toprule
    Setting & Value \\
    \midrule
    loss function          & cispo \\
    CISPO clip low / high  & $0.8\ /\ 6.0$ \\
    Learning rate          & $3\times 10^{-5}$ \\
    KL penalty (kl coef)   & $0$ \\
    batch size             & $64$ \\
    group size             & $8$ \\
    num substeps           & $3$ \\
    epochs                 & $1$ \\
    base environment seed  & $42$ \\
    \bottomrule
  \end{tabular}
  \caption{Optimization hyperparameters for the RL training run.}
  \label{tab:multiitem_train_hyperparams}
\end{table}

\begin{table}[h]
  \centering
  \small
  \setlength{\tabcolsep}{4pt}
  \begin{tabular}{@{}ll@{}}
    \toprule
    Setting & Value \\
    \midrule
    num buyers $m$                  & $3$ \\
    num items $n$                   & $3$ \\
    total turns $T$                 & $7$ \\
    per-buyer turn limit $K$        & $3$ \\
    valuation ratio distribution    & $\mathcal{U}[0.40,\, 1.15]$ \\
    first-offer-noise probability    & $0.5$ \\
    first-offer-noise ratio range    & $[0.25,\, 0.30] \cdot P_{\text{list},j}$ \\
    seller temperature              & $1.0$ \\
    buyer temperature               & $0.7$ \\
    \bottomrule
  \end{tabular}
  \caption{Environment hyperparameters for training. Each buyer valuation entry is $\mathcal{B}_{i,j} = r_{i,j}\,P_{\text{list},j}$ with $r_{i,j} \sim \mathcal{U}[0.40,1.15]$ drawn i.i.d. across all $(i,j)$ pairs. First-offer noise is applied independently to each buyer with probability $0.5$: at episode start, a buyer selected for first-offer noise has its opening offer on every chosen item resampled into $[0.25,0.30] \cdot P_{\text{list},j}$, decoupling the opening from the buyer's private valuation; otherwise the buyer generates its opening offer freely subject to its private valuation constraint.}
  \label{tab:multiitem_episode_hyperparams}
\end{table}

\begin{table}[ht]
  \centering
  \small
  \setlength{\tabcolsep}{4pt}
  \begin{tabular}{@{}ll@{}}
    \toprule
    Setting & Value \\
    \midrule
    seller model              & All models from Table~\ref{tab:model_sources_full} \\
    buyer model               & Qwen3-30B-A3B-Instruct-2507 iter-$60$ (Table~\ref{tab:multiitem_models}) \\
    Test split size           & $240$ held-out clusters \\
    seller temperature (eval) & $0.7$ \\
    buyer temperature (eval)  & $0.7$ \\
    \bottomrule
  \end{tabular}
  \caption{Evaluation hyperparameters for the held-out test split. The $240$ test clusters are the $20\%$ cluster-level hold-out from the $12$ in-distribution categories (\Cref{sec:dataset}); product contexts are unseen during training.}
  \label{tab:multiitem_eval_hyperparams}
\end{table}

\clearpage
\subsection{Out-of-Distribution Evaluation Setup}
\label{subsec:ood_setup}

The out-of-distribution (OOD) result tables report several environments that each perturb the in-distribution setting of \Cref{subsec:market_settings} along one or more axes. The in-distribution reference uses $m=3$ buyers, $n=3$ items, and a total communication budget $T=7$ with a fixed iter-$60$ buyer and the uniform valuation matrix. Across all OOD environments the buyer agent is instead sampled per buyer per episode, uniformly from the pool $\{\text{iter-}50,\ \text{iter-}60,\ \text{iter-}70\}$ of the buyer training run, so the seller never faces a memorized single counterpart. The remaining axes vary as follows: a widened $5$-buyer / $5$-item / $T=12$ environment under the same uniform valuation; the same widened environment with a per-item valuation multiplier $\mu_j \sim \mathcal{U}[0.5,2]$ (some items universally more valued); the same widened environment with a per-buyer generosity multiplier $\gamma_i \sim \mathcal{U}[0.5,2]$ (some buyers universally more willing to pay); a $5$-buyer / $3$-item / $T=12$ environment that scales only the buyer count; a $3$-buyer / $5$-item / $T=7$ environment that scales only the item count; and a held-out Musical Instruments environment that keeps the widened $5$-buyer / $5$-item / $T=12$ shape but draws every catalog from the $13$th category, whose higher list-price range moves the absolute valuation and cost scale outside the training distribution. Table~\ref{tab:ood_eval_regimes} summarizes the environments.

\begin{table}[h]
\centering
\small
\setlength{\tabcolsep}{5pt}
\renewcommand{\arraystretch}{1.15}
\begin{tabular}{@{}lcccll@{}}
\toprule
Environment & Buyers $m$ & Items $n$ & Turns $T$ & Valuation & Catalog \\
\midrule
In-distribution (reference)        & $3$ & $3$ & $7$  & uniform                       & $12$ categories \\
\midrule
Env.~1: Expanded Standard          & $5$ & $5$ & $12$ & uniform                       & $12$ categories \\
Env.~2: Expanded Musical           & $5$ & $5$ & $12$ & uniform                       & Musical Instruments \\
Env.~3: Buyer Heavy                & $5$ & $3$ & $12$ & uniform                       & $12$ categories \\
Env.~4: Item Heavy                 & $3$ & $5$ & $7$  & uniform                       & $12$ categories \\
Env.~5: Item Correlation           & $5$ & $5$ & $12$ & uniform $\times\,\mu_j$       & $12$ categories \\
Env.~6: Buyer Correlation          & $5$ & $5$ & $12$ & uniform $\times\,\gamma_i$    & $12$ categories \\
\bottomrule
\end{tabular}
\caption{Per-episode environment knobs for the in-distribution reference and the out-of-distribution evaluation environments. ``uniform'' denotes the in-distribution valuation $\mathcal{B}_{i,j} = r_{i,j}\,P_{\text{list},j}$ with $r_{i,j} \sim \mathcal{U}[0.40,1.15]$. Env.~5 (Item Correlation) draws $\mu_j \sim \mathcal{U}[0.5,2]$ once per item with $\gamma_i=1$, while Env.~6 (Buyer Correlation) draws $\gamma_i \sim \mathcal{U}[0.5,2]$ once per buyer with $\mu_j=1$. The in-distribution reference uses the fixed iter-$60$ buyer; every OOD environment samples each buyer slot uniformly from the pool $\{\text{iter-}50, \text{iter-}60, \text{iter-}70\}$. First-offer noise (Table~\ref{tab:multiitem_episode_hyperparams}) is applied in every environment.}
\label{tab:ood_eval_regimes}
\end{table}

\section{Dataset Details}
\label{app:dataset}

This section records construction details, additional statistics, a complete example cluster, and schema notes for the substitutable-product dataset of \Cref{sec:dataset}.

\subsection{Construction Details}
\label{subsec:dataset_construction}

The dataset was assembled in three stages. First, a language model proposed candidate narrow product types within each of the $13$ broad Amazon categories, serving as cluster seeds. Second, for every candidate type, the best-selling Amazon products of that type were retrieved through Keepa (the third-party Amazon catalog and price-history service introduced in \Cref{sec:dataset}), restricted to products whose category ancestry matches the candidate's broad category and that pass basic per-product checks (a valid list price, available metadata, an identifiable cost). Third, an automated substitutability review examined each candidate set and kept only products a buyer could genuinely swap for one another---same function, same use case---writing a short rationale for every kept set; candidate types without a substitutable core were discarded. All prices, costs, sales rankings, and product text are a snapshot taken between May 13 and May 17, 2026.

\subsection{Additional Statistics and Category Breakdowns}
\label{subsec:app_additional_stats}

This section provides an expanded empirical look at the dataset profiles across all 13 product categories. \Cref{tab:dataset_per_category} reports the per-category statistics and \Cref{fig:dataset_pie} visualizes the composition. The corpus is balanced by construction: every category contributes exactly $100$ clusters of $5$ products. Clusters are genuinely multi-brand---$4.4$ distinct brands per cluster on average, and $59\%$ of clusters carry $5$ distinct brands---so the seller cannot collapse the catalog into a single-brand pitch. Within a cluster, the median price spread, defined as $s = P_{\max}/P_{\min}$ over the cluster's $5$ list prices, is $1.85\times$ corpus-wide, giving the seller meaningful headroom to steer buyers across price tiers. Median prices also differ sharply across categories---from \$29.99 in Beauty \& Personal Care to \$165.49 in Musical Instruments---an asymmetry the out-of-distribution split below exploits.

\begin{figure}[h]
  \centering
  \includegraphics[width=0.92\linewidth]{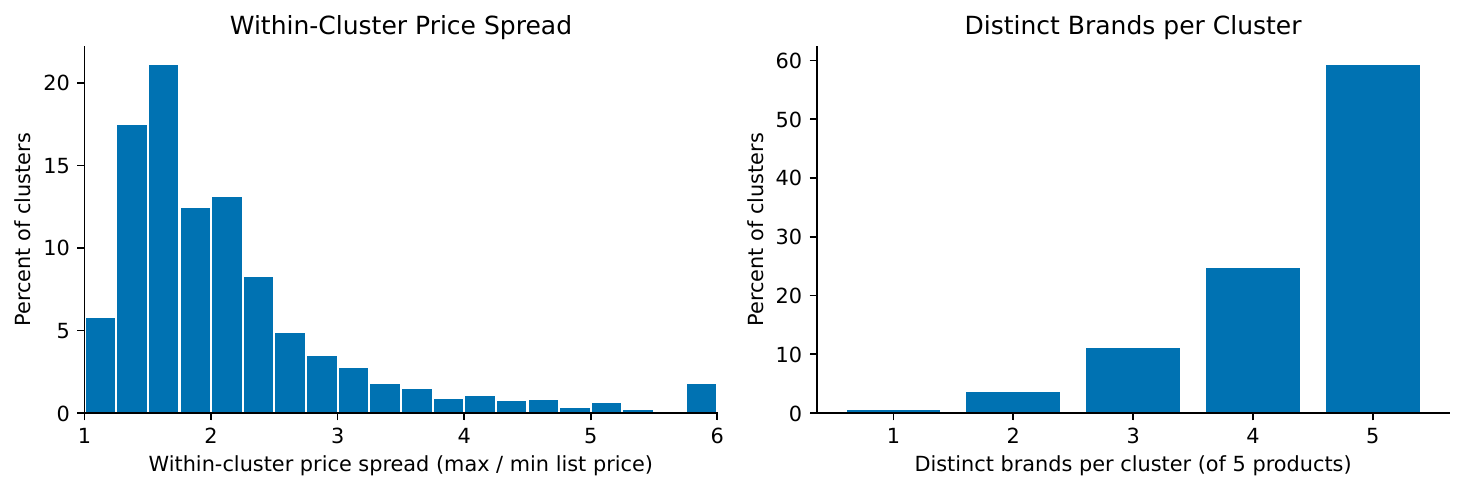}
  \caption{Distributions over the $1{,}300$ clusters of (left) the within-cluster price spread $s = P_{\max}/P_{\min}$, with the $24$ clusters whose spread exceeds $6\times$ folded into the last bin, and (right) the number of distinct brands among each cluster's $5$ products.}
  \label{fig:dataset_spread_brands}
  \Description{Two bar charts over the $1{,}300$ clusters. Left, the within-cluster price spread $P_{\max}/P_{\min}$ is right-skewed: most clusters fall between roughly $1.5\times$ and $2.4\times$, with a sparse tail out to larger spreads and the $24$ clusters above $6\times$ folded into the last bin. Right, the number of distinct brands among each cluster's five products piles up at the top of its range rather than trailing off: about $59\%$ of clusters contain five distinct brands and roughly $24\%$ contain four, while clusters with two or three brands are uncommon.}
\end{figure}

Furthermore, \Cref{fig:dataset_spread_brands} shows the distribution of the within-cluster price spread $s = P_{\max}/P_{\min}$ and of the distinct-brand count across all $1{,}300$ clusters. Spreads concentrate between $1.5\times$ and $2.4\times$ (quartiles $1.50$, $1.85$, $2.37$), with Musical Instruments carrying the widest per-category median spread ($2.40\times$) and Baby the narrowest ($1.61\times$). Brand diversity is high: $59\%$ of clusters carry $5$ distinct brands among their $5$ products and $84\%$ carry at least $4$, so brand-conditioned buyer preferences rarely collapse onto a single pitch.

\begin{table}[t]
  \centering
  \footnotesize
  \setlength{\tabcolsep}{4pt}
  \resizebox{\linewidth}{!}{%
\begin{tabular}{@{}lccccc@{}}
\toprule
Broad Category & Mean Distinct Brands & Median List Price & Median Cluster Min Price & Median Cluster Max Price & Median Price Spread \\
\midrule
Appliances & 4.3 & \$109.98 & \$79.99 & \$159.99 & 1.71$\times$ \\
Automotive & 4.3 & \$45.86 & \$31.49 & \$59.99 & 2.00$\times$ \\
Baby & 4.5 & \$34.99 & \$28.24 & \$48.97 & 1.61$\times$ \\
Beauty \& Personal Care & 4.5 & \$29.99 & \$22.49 & \$43.98 & 1.79$\times$ \\
Clothing, Shoes \& Jewelry & 4.4 & \$36.49 & \$29.95 & \$49.99 & 1.73$\times$ \\
Electronics & 3.9 & \$139.99 & \$99.79 & \$199.99 & 1.95$\times$ \\
Home \& Kitchen & 4.6 & \$46.99 & \$35.99 & \$61.49 & 1.88$\times$ \\
Musical Instruments & 4.1 & \$165.49 & \$88.26 & \$248.84 & 2.40$\times$ \\
Patio, Lawn \& Garden & 4.6 & \$69.99 & \$59.99 & \$99.99 & 1.91$\times$ \\
Pet Supplies & 4.7 & \$33.49 & \$22.74 & \$45.49 & 1.75$\times$ \\
Sports \& Outdoors & 4.4 & \$39.99 & \$32.49 & \$59.99 & 1.81$\times$ \\
Tools \& Home Improvement & 4.3 & \$59.95 & \$39.99 & \$80.00 & 2.00$\times$ \\
Toys \& Games & 4.5 & \$31.99 & \$24.74 & \$43.46 & 1.82$\times$ \\
\midrule
\textbf{Total} & 4.4 & \$47.99 & \$34.08 & \$62.99 & 1.85$\times$ \\
\bottomrule
\end{tabular}
  }
  \caption{Per-category statistics of the substitutable-product dataset. Every category contributes $100$ clusters of $5$ substitutable products ($500$ products per category; $1{,}300$ clusters and $6{,}500$ products in total). Brand counts are means over a category's $100$ clusters; prices are medians; the within-cluster price spread $s = P_{\max}/P_{\min}$ is computed over each cluster's $5$ list prices and reported as the per-category median.}
  \label{tab:dataset_per_category}
\end{table}

\begin{figure}[!t]
\centering
\begin{minipage}{0.50\textwidth}
    \centering
    \includegraphics[width=\linewidth]{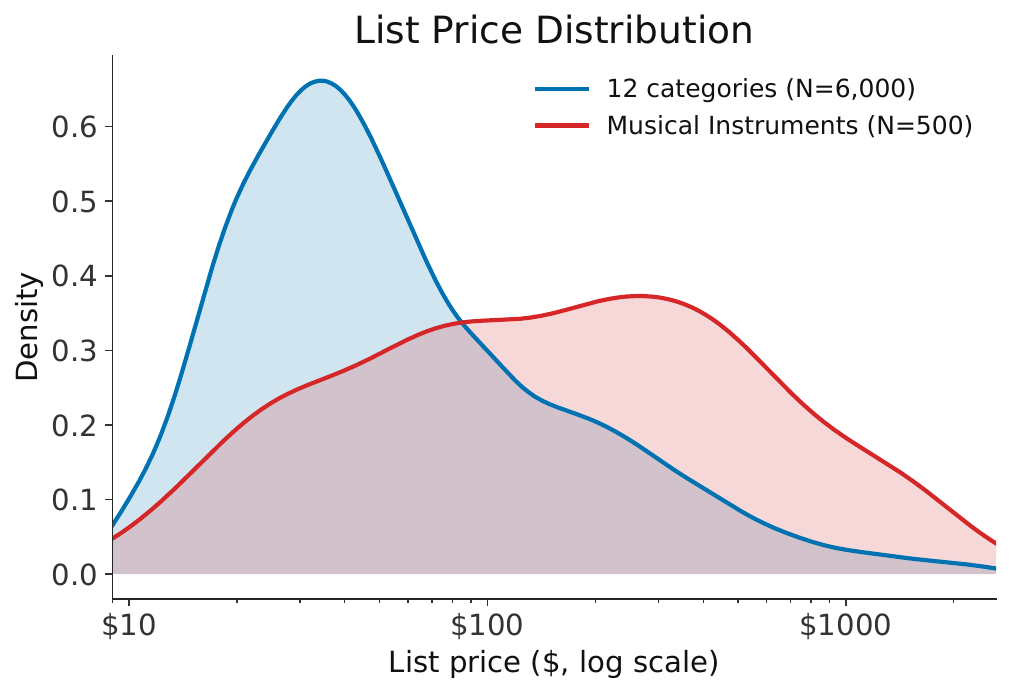}
\end{minipage}
\hfill
\begin{minipage}{0.45\textwidth}
  \caption{List price distribution of the dataset's $6{,}500$ products on a log scale: Musical Instruments ($N{=}500$, red) versus the $12$ remaining categories pooled ($N{=}6{,}000$, blue). Musical Instruments concentrates in a visibly higher price range, motivating its selection as the held-out out-of-distribution category.}
    \label{fig:list_price_distribution}
  \Description{Two overlaid filled density curves of product list prices on a logarithmic price axis. The blue curve is the $12$ pooled in-distribution categories ($N=6{,}000$); the red curve is Musical Instruments ($N=500$). The Musical Instruments density is shifted toward substantially higher prices, with most of its mass to the right of the pooled curve.}

\end{minipage}
\end{figure}

\subsection{Example Cluster}
\label{subsec:dataset_example_cluster}

The boxes below reproduce the dataset record behind the qualitative example of Appendix~\ref{app:transcripts}: the video-recorder cluster (\texttt{Video Recorders}, Electronics)---one box with the cluster-level fields, followed by one box per product, covering the three items of the episode catalog (\Cref{subsec:episode_information}) and two further products of the cluster. The product features are truncated to their first $200$ characters for display. The lowest all-time price---the lowest price at which the product was ever sold on Amazon---is the product's seller cost $\mathcal{C}_j$ (\Cref{sec:dataset}).

\begin{chatmsg}[grey]{Cluster Information}[breakable]
  \small
  Broad Category: Electronics\\
  Narrow Product Type: Video Recorders\\
  Leaf Category ID: \texttt{14241611}\\
  Product ASINs: \texttt{B0749F32T1}, \texttt{B08LSRT2JB}, \texttt{B06WP6Q898}, \texttt{B01DELNN82}, \texttt{B0DB1XN58W}\\
  Number of Products: 5\\
  Minimum List Price: \$55.99\\
  Maximum List Price: \$419.99\\
  Distinct Brands: 4\\
  Seller Type Counts: \texttt{third\_party\_only} (3), \texttt{amazon} (2)
\end{chatmsg}

\begin{chatmsg}[grey]{Product 1: ANNKE}[breakable]
  \small
  ASIN: \texttt{B0749F32T1}\\
  Title: ``ANNKE 3K Lite 8 Channel Hybrid 5-in-1 Security Digital Video Recorder, H.265+ Surveillance DVR Supports 8CH Analog and 2CH 6MP IP Cameras, Remote Access, AI Human\&Vehicle Detection, 1TB Hard Drive''\\
  Features: ``5-in-1 Hybrid DVR -- This expandable hybrid DVR combines DVR, NVR and HVR features. It supports up to 8 cameras, including 720P/960H/1080P/5MP (20fps) analog HD-TVI, CVI, AHD, CVBS, plus 2 extra IP cam ...''\\
  Brand: ANNKE\\
  List Price: \$169.99\\
  Lowest Price All-Time: \$55.99\\
  Seller Type: \texttt{third\_party\_only}\\
  Narrow Product Type: Video Recorders\\
  Broad Category: Electronics
\end{chatmsg}

\begin{chatmsg}[grey]{Product 2: Lorex}[breakable]
  \small
  ASIN: \texttt{B08LSRT2JB}\\
  Title: ``Lorex 4K Ultra HD 8-Channel Fusion Series PoE Network Video Recorder with 2TB Storage, Smart Motion Detection and Smart Home Compatibility''\\
  Features: ``4K ULTRA HD - Record and view impressive 4K video resolution with a cutting-edge recorder, providing you with superior detail and definition. LOREX FUSION - As part of the Lorex Fusion collection, you ...''\\
  Brand: Lorex\\
  List Price: \$249.99\\
  Lowest Price All-Time: \$100.00\\
  Seller Type: \texttt{third\_party\_only}\\
  Narrow Product Type: Video Recorders\\
  Broad Category: Electronics
\end{chatmsg}

\begin{chatmsg}[grey]{Product 3: ZOSI}[breakable]
  \small
  ASIN: \texttt{B06WP6Q898}\\
  Title: ``ZOSI H.265+ 8 Channel 5MP Lite 1080p CCTV DVR Recorder Without HDD, Hybrid Capability 4-in-1(Analog/AHD/TVI/CVI) Surveillance DVR for Security Camera, Remote Access, Human Vehicle Detect, Alert Push''\\
  Features: ``(1). [ AI Human \& Vehicle Detection for Accurate Alerts] \_\_ Equipped with advanced AI technology, this DVR system intelligently detects humans and vehicles, reducing false alarms caused by animals or ...''\\
  Brand: ZOSI\\
  List Price: \$55.99\\
  Lowest Price All-Time: \$42.07\\
  Seller Type: \texttt{amazon}\\
  Narrow Product Type: Video Recorders\\
  Broad Category: Electronics
\end{chatmsg}

\begin{chatmsg}[grey]{Product 4: REOLINK}[breakable]
  \small
  ASIN: \texttt{B01DELNN82}\\
  Title: ``REOLINK 16CH Network Video Recorder for Home Security Camera System,Only Work with 16MP/12MP/4K/8MP/5MP/4MP The Same Brand PoE IP Cam,24/7 Recording to Pre-Installed 4TB Hard Drive,RLN16-410''\\
  Features: ``ONLY WORK WITH REOLINK IP CAMERAS: Work perfectly with all Reolink PoE 16MP/12MP/4K/5MP/4MP cameras, such as Duo 3 PoE, RLC-1212A, RLC-1224A, RLC-823A 16X, Trackmix PoE, Duo 2 PoE, 811A, 810A, 820A, 5 ...''\\
  Brand: REOLINK\\
  List Price: \$419.99\\
  Lowest Price All-Time: \$271.99\\
  Seller Type: \texttt{amazon}\\
  Narrow Product Type: Video Recorders\\
  Broad Category: Electronics
\end{chatmsg}

\begin{chatmsg}[grey]{Product 5: REOLINK}[breakable]
  \small
  ASIN: \texttt{B0DB1XN58W}\\
  Title: ``REOLINK Home Hub Pro, Centralized Control for up to 24 REOLINK Security Cameras, up to 16TB HDD Storage, 2TB HDD Included, AES-128 Encryption, 16MP Streaming, Wi-Fi 6, 120dB Siren. No Monthly Fees''\\
  Features: ``All-In-One Control \& Storage Center: Centralized storage \& alarm management for all Reolink cams - wired, wireless, PoE \& doorbells (excluding 4G devices) on the portable App. Effortlessly monitor up ...''\\
  Brand: REOLINK\\
  List Price: \$339.99\\
  Lowest Price All-Time: \$179.99\\
  Seller Type: \texttt{third\_party\_only}\\
  Narrow Product Type: Video Recorders\\
  Broad Category: Electronics
\end{chatmsg}

\subsection{Dataset Structure and Details}
\label{subsec:dataset_schema}

Each cluster row contains basic market details, including the broad product category, the specific product type, unique product IDs, the number of distinct brands, the price range, and a short explanation of why the items can be substituted for one another. Within these clusters, individual product records track product titles, brands, public list prices, the lowest historical price (used as the seller's cost), and where the product is sold.

The dataset also includes two text fields: a long product description and a bulleted list of features. While long descriptions are only available for about a third of the products, the bulleted feature lists are included for every item. Because these feature lists are always present, the negotiation system uses them as the main text to show product details during our tests.

\section{Prompt Templates}
\label{app:prompts}

This section reproduces, verbatim, the system prompts that define the two roles of the multi-item negotiation. Both roles are played by the same base model, \texttt{Qwen3-30B-A3B-Instruct-2507} \citep{qwen3technicalreport}: the seller is the trained checkpoint; buyer slots use the frozen iteration-$60$ checkpoint of a bilateral-negotiation RLVR run \citep{liu2026instructing} (\Cref{subsec:agent_models}) during training and in-distribution evaluation, while OOD evaluation uses the checkpoint pool described in \Cref{subsec:ood_setup}. The two prompts establish the XML action grammar, the action space (\act{SELL}/\act{DEAL}/\act{QUIT} for the seller, \act{BUY}/\act{DEAL}/\act{QUIT} for the buyer), and the information partition that separates each agent's private reasoning (\texttt{<negotiation\_thought>}) from its public message (\texttt{<talk>}) and formal move (\texttt{<action>}). The seller prompt is a \texttt{.format()} template whose five placeholders are substituted per episode; the buyer prompt is a fixed string. Each agent's per-episode context---catalog, valuations, and standing state---is supplied through the user turns and is not shown here.

\subsection{Seller System Prompts}

The seller's system prompt is the mechanics-only template below. Its five placeholders---\texttt{\{m\}}, \texttt{\{n\}}, \texttt{\{K\}}, \texttt{\{T\}}, and the seller-facing \texttt{\{catalog\_block\}}---are substituted at runtime with the per-episode constants and the catalog listing (one line per item, each carrying the item's public list price and the seller's private cost; \Cref{subsec:market_settings}).

\begin{chatmsg}[grey]{System Prompt for Seller}[breakable]
  \small
  You are a seller negotiating an n-item catalog with m buyers in parallel.

  \medskip
  \# Symbols
  \begin{itemize}\itemsep0pt
    \item $i \in \{1, \ldots, m\}$ --- buyer index.
    \item $j \in \{1, \ldots, n\}$ --- item index in the catalog.
    \item \texttt{cost\_j} --- your private reservation cost for item j (never sell below this).
    \item \texttt{list\_price\_j} --- public list price for item j (buyers can see this).
    \item \texttt{p\_deal,j} --- the agreed price at which item j is sold (defined only when item j closes; MAY exceed \texttt{list\_price\_j}).
    \item \texttt{p\_final,j = min(p\_deal,j, list\_price\_j)}: the list-clipped price that enters the reward numerator.
    \item \texttt{B\_\{i,j\}} --- buyer i's private per-item budget for item j (you never see this).
  \end{itemize}

  Your goal is to maximize the catalog-wide reward:

  \medskip
  \texttt{R = sum\_j (p\_final,j - cost\_j) *} $\mathds{1}$\texttt{[item j sold] / sum\_j (list\_price\_j - cost\_j)}

  \medskip
  $R \in [0, 1]$ on a valid episode; $R = -1$ on any seller violation. Selling above list is allowed, but dollars above \texttt{list\_price\_j} do not count (numerator clips at list).

  \medskip
  A seller violation is any of:
  \begin{itemize}\itemsep0pt
    \item[(i)] a priced \act{SELL} or \act{DEAL} at \texttt{\$M < cost\_j} for some item j;
    \item[(ii)] \texttt{<target>} naming a buyer whose status is not active;
    \item[(iii)] \act{SELL} listing an item j that has already closed in a prior deal;
    \item[(iv)] \texttt{<target>} missing, unparseable, or out of range;
    \item[(v)] the reply does not parse into the four XML tags below.
  \end{itemize}

  \medskip
  \# Output format

  Every reply MUST be exactly four XML tags, in this order (\texttt{<negotiation\_thought>}, \texttt{<target>}, \texttt{<talk>}, \texttt{<action>}):
  \begin{itemize}\itemsep0pt
    \item \texttt{<negotiation\_thought>} your hidden reasoning; not sent to buyers. \texttt{</negotiation\_thought>}
    \item \texttt{<target>Buyer i</target>} \quad (i in 1..m, currently active)
    \item \texttt{<talk>}short message; visible only to the Targeted buyer\texttt{</talk>}
    \item \texttt{<action>}EXACTLY ONE of the three action forms below\texttt{</action>}
  \end{itemize}

  Where the action inside \texttt{<action>...</action>} is exactly one of:
  \begin{itemize}\itemsep0pt
    \item \act{SELL} \texttt{[<codename\_1, \$M\_1>, <codename\_2, \$M\_2>, ...]} --- pitch any subset of available items, one price each.
    \item \act{DEAL} \texttt{codename\_j \$M} --- accept the Targeted buyer's standing \act{BUY} \texttt{codename\_j \$M} (price must match).
    \item \act{QUIT} --- end the episode.
  \end{itemize}

  \medskip
  \# Episode constants

  \texttt{m = \{m\}} buyers \quad \texttt{n = \{n\}} items \\
  \texttt{K = \{K\}} per-buyer message cap \quad \texttt{T = \{T\}} total seller-message budget \quad (\texttt{T < m*K})

  \medskip
  \# Item catalog (private --- buyers see only list prices, never your cost)

  \texttt{\{catalog\_block\}}

  \medskip
  \# Conversation mechanics
  \begin{itemize}\itemsep0pt
    \item Turn 1: each buyer sends an opening offer naming a subset of items at proposed prices. Items not opened on turn 1 remain pitchable via \act{SELL}.
    \item Later turns: the buyer you Target replies; others stay silent.
    \item A buyer sees ONLY your \texttt{<talk>} and \texttt{<action>} --- never \texttt{<negotiation\_thought>}, \texttt{<target>}, or any other buyer's dialogue.
    \item Buyers' budgets \texttt{B\_\{i,j\}} are private (see Symbols).
    \item Each buyer can close at most one item. After their \act{DEAL}, they become satisfied and cannot be Targeted again.
    \item \act{DEAL} only closes against a standing \act{BUY} from the Targeted buyer --- the item and price must match an outstanding \act{BUY} from them.
    \item An item sold to one buyer is removed from the catalog; listing it in a later \act{SELL} is a violation.
    \item The most recent user message ends with \texttt{=== Standing offers ===}; that block is the freshest source of truth.
  \end{itemize}

  \medskip
  You must limit your reply under 400 tokens for all 4 parts combined.
\end{chatmsg}

\subsection{Buyer System Prompts}

During training and in-distribution evaluation, every buyer slot is filled by the same base model as the seller, \texttt{Qwen3-30B-A3B-Instruct-2507}, frozen at the iteration-$60$ checkpoint of a bilateral-negotiation RLVR run \citep{liu2026instructing}. Across the out-of-distribution evaluations, each buyer slot is instead sampled per episode from the checkpoint pool $\{\text{iter-}50, \text{iter-}60, \text{iter-}70\}$ of the same buyer training trajectory. In both cases only the seller's parameters update during training. The buyer system prompt is the fixed string below, with a token-budget rider appended (last paragraph). It carries no placeholders: the per-episode valuation vector, the buyer-facing item board, and the optional first-offer block are delivered through the buyer's user turns.

\begin{chatmsg}[grey]{System Prompt for Buyer}[breakable]
  \small
  You are a buyer negotiating with one seller against an n-item catalog of substitutable items. You may close at most ONE item; once you \act{DEAL}, the negotiation ends for you.

  \medskip
  \# Output format

  Every reply MUST be exactly three XML tags, in this order:
  \begin{itemize}\itemsep0pt
    \item \texttt{<negotiation\_thought>} your hidden reasoning; not sent to the seller. \texttt{</negotiation\_thought>}
    \item \texttt{<talk>}short message; visible to the seller\texttt{</talk>}
    \item \texttt{<action>}EXACTLY ONE of the three action forms below\texttt{</action>}
  \end{itemize}

  Where the action inside \texttt{<action>...</action>} is exactly one of:
  \begin{itemize}\itemsep0pt
    \item \act{BUY} \texttt{[<codename\_1, \$M\_1>, <codename\_2, \$M\_2>, ...]} --- propose buying any subset (1..n) of currently-available items, with one price per item. Each price must be at most your private budget for that item (\texttt{\$M\_j <= B\_\{i,j\}}); the env rejects an over-budget bid.
    \item \act{DEAL} \texttt{codename\_j \$M} --- accept the seller's standing \act{SELL} \texttt{codename\_j \$M} for that item at the exact price \texttt{\$M} they offered. Singleton --- closes exactly one item.
    \item \act{QUIT} --- end your participation in the episode.
  \end{itemize}

  \medskip
  \# Conversation mechanics
  \begin{itemize}\itemsep0pt
    \item You cannot propose to BUY an item that has already been sold. The user message includes a ``Available / Sold'' board telling you which codenames are valid targets.
  \end{itemize}

  \medskip
  \# Reward (terminal only)

  For each item you might close, your bargained ratio is:

  \medskip
  \texttt{rho = (B\_\{i,j\} - p\_deal\_j) / (B\_\{i,j\} - cost\_j)}

  \medskip
  where \texttt{p\_deal\_j} is the price you actually pay. A lower \texttt{p\_deal\_j} is better. You can only close one item this episode.

  \medskip
  Constraint: each reply you produce must be at most 300 tokens in total (Thought, Talk, and Action combined).
\end{chatmsg}

\noindent
The optional first-offer block, when present, is appended to a buyer's first user turn and implements the first-offer noise of \Cref{subsec:market_settings}: at episode start each buyer independently, with probability $0.5$, has its opening offer on every chosen item resampled into $[0.25, 0.30] \cdot P_{\text{list},j}$ independently of its true valuation, while otherwise the buyer opens freely within its private valuation bound. When active for a buyer, the block pins those opening prices for that buyer's first reply only; from its second reply onward the buyer bids freely within its per-item budget.

\section{Extended Benchmarking Results}
\label{app:extended_benchmarks}

This section presents the complete evaluation tables for \Cref{sec:benchmark} and \Cref{sec:ood}. The in-distribution evaluation reports sample means alongside their corresponding standard errors ($\text{mean} \pm \text{standard error}$) computed over the full evaluation episode corpus. For each out-of-distribution environment we provide a point-estimate table matching the main-text layout together with a companion table carrying the corresponding standard errors. The point-estimate tables for Envs.~1--6 are \Cref{tab:ood_uniform_main,tab:ood_uniform_mi_main,tab:ood_5b3i_main,tab:ood_3b5i_main,tab:ood_itemmult_main,tab:ood_buyergen_main}, respectively. The model evaluation panel and metric column structures mirror the main-text layouts, with top-performing values highlighted in bold and second-best values underlined.

All seller metrics reported here are estimated from direct episode rollouts rather than through off-policy estimators or optimized data-collection policies; the latter address complementary questions of variance and sample efficiency in reinforcement-learning policy evaluation
\citep{liu2024efficient, liu2024doubly, chen2024efficient, liu2024efficientmul, liu2025efficientrobust, chen2026robust}.

\begin{table}[h]
  \centering
  \small
  \setlength{\tabcolsep}{4pt}
  \resizebox{\linewidth}{!}{%
\begin{tabular}{lccccccc}
\toprule
Model & Params & Reward $\uparrow$ & \makecell{Seller Surplus\\Extraction\\Ratio} & \makecell{Item\\Deal\\Rate} & \makecell{Deal on\\Top Buyer\\Rate} & \makecell{Optimal-Pair\\Offer Coverage\\Rate} & \makecell{Instruction\\Violation\\Rate} \\
\midrule
\rowcolor{gray!10} Qwen3-30B-A3B-Instruct-2507-trained (\textbf{Ours}) & 30B & $\mathbf{+0.407 \pm 0.018}$ & $\mathbf{+0.425 \pm 0.015}$ & $64.7\% \pm 1.6\%$ & $\mathbf{0.687 \pm 0.022}$ & $\mathbf{0.844 \pm 0.014}$ & $1.3\% \pm 0.7\%$ \\
\rowcolor{gray!10} Qwen3-30B-A3B-Instruct-2507-untrained & 30B & $-0.329 \pm 0.039$ & $+0.184 \pm 0.015$ & $66.2\% \pm 2.4\%$ & $0.544 \pm 0.030$ & $0.554 \pm 0.026$ & $43.3\% \pm 3.2\%$ \\
\midrule
Kimi-K2.6-thinking & 1T & $\underline{+0.379 \pm 0.013}$ & $+0.379 \pm 0.013$ & $73.0\% \pm 1.5\%$ & $0.566 \pm 0.022$ & $0.559 \pm 0.020$ & $\mathbf{0.0\% \pm 0.0\%}$ \\
DeepSeek-V4-Pro-thinking & 1.6T & $+0.369 \pm 0.018$ & $\underline{+0.392 \pm 0.015}$ & $73.4\% \pm 1.6\%$ & $0.612 \pm 0.021$ & $0.592 \pm 0.020$ & $1.7\% \pm 0.8\%$ \\
DeepSeek-V4-Pro-nothink & 1.6T & $+0.361 \pm 0.017$ & $+0.378 \pm 0.014$ & $\underline{76.2\% \pm 1.6\%}$ & $0.596 \pm 0.021$ & $0.640 \pm 0.019$ & $1.3\% \pm 0.7\%$ \\
DeepSeek-V4-Flash-thinking & 284B & $+0.349 \pm 0.017$ & $+0.366 \pm 0.014$ & $71.0\% \pm 1.7\%$ & $0.568 \pm 0.022$ & $0.566 \pm 0.020$ & $1.3\% \pm 0.7\%$ \\
Kimi-K2.6-nothink & 1T & $+0.331 \pm 0.021$ & $+0.378 \pm 0.013$ & $75.0\% \pm 1.5\%$ & $\underline{0.616 \pm 0.021}$ & $\underline{0.654 \pm 0.019}$ & $3.4\% \pm 1.2\%$ \\
GPT-5.4-high-reasoning & closed-source & $+0.309 \pm 0.014$ & $+0.315 \pm 0.013$ & $\mathbf{78.1\% \pm 1.4\%}$ & $0.560 \pm 0.021$ & $0.604 \pm 0.020$ & $\underline{0.4\% \pm 0.4\%}$ \\
DeepSeek-V4-Flash-nothink & 284B & $+0.241 \pm 0.022$ & $+0.306 \pm 0.013$ & $73.4\% \pm 1.6\%$ & $0.538 \pm 0.022$ & $0.543 \pm 0.020$ & $5.0\% \pm 1.4\%$ \\
GPT-5.4-mini-high-reasoning & closed-source & $+0.111 \pm 0.025$ & $+0.212 \pm 0.013$ & $72.3\% \pm 1.8\%$ & $0.465 \pm 0.023$ & $0.457 \pm 0.020$ & $8.4\% \pm 1.8\%$ \\
\bottomrule
\end{tabular}
  }
  \caption{\textbf{In-distribution held-out evaluation, full results} ($3$ buyers, $3$ items, $T=7$).}
  \label{tab:indist3_appendix}
\end{table}

\begin{table}[h]
  \centering
  \small
  \setlength{\tabcolsep}{4pt}
  \resizebox{\linewidth}{!}{%
\begin{tabular}{lccccccc}
\toprule
Model & Params & Reward $\uparrow$ & \makecell{Seller Surplus\\Extraction\\Ratio} & \makecell{Item\\Deal\\Rate} & \makecell{Deal on\\Top Buyer\\Rate} & \makecell{Optimal-Pair\\Offer Coverage\\Rate} & \makecell{Instruction\\Violation\\Rate} \\
\midrule
\rowcolor{gray!10} Qwen3-30B-A3B-Instruct-2507-trained (\textbf{Ours}) & 30B & $\mathbf{+0.444}$ & $\mathbf{+0.493}$ & $70.8\%$ & $\mathbf{0.594}$ & $\mathbf{0.699}$ & $3.3\%$ \\
\rowcolor{gray!10} Qwen3-30B-A3B-Instruct-2507-untrained & 30B & $-0.187$ & $+0.240$ & $74.1\%$ & $0.384$ & $0.378$ & $34.5\%$ \\
\midrule
Kimi-K2.6-thinking & 1T & $\underline{+0.322}$ & $+0.345$ & $75.9\%$ & $0.398$ & $0.402$ & $\underline{1.7\%}$ \\
DeepSeek-V4-Pro-nothink & 1.6T & $+0.308$ & $+0.347$ & $81.5\%$ & $0.417$ & $0.426$ & $2.9\%$ \\
DeepSeek-V4-Flash-thinking & 284B & $+0.268$ & $+0.353$ & $77.1\%$ & $0.388$ & $0.398$ & $6.3\%$ \\
GPT-5.4-high-reasoning & closed-source & $+0.265$ & $+0.265$ & $\mathbf{82.8\%}$ & $0.363$ & $0.439$ & $\mathbf{0.0\%}$ \\
DeepSeek-V4-Flash-nothink & 284B & $+0.204$ & $+0.326$ & $78.0\%$ & $0.392$ & $0.422$ & $9.2\%$ \\
Kimi-K2.6-nothink & 1T & $+0.140$ & $+0.328$ & $\underline{82.0\%}$ & $\underline{0.426}$ & $0.420$ & $14.2\%$ \\
DeepSeek-V4-Pro-thinking & 1.6T & $+0.133$ & $\underline{+0.396}$ & $79.0\%$ & $0.418$ & $\underline{0.444}$ & $18.8\%$ \\
GPT-5.4-mini-high-reasoning & closed-source & $+0.035$ & $+0.212$ & $78.3\%$ & $0.354$ & $0.381$ & $14.6\%$ \\
\bottomrule
\end{tabular}
  }
  \caption{\textbf{Out-of-distribution Env.~1, Expanded Standard ($5$ buyers, $5$ items, $T=12$, uniform valuation)} with the diverse buyer pool. Baselines sorted by \textit{Reward} (descending); best per column in bold, second-best underlined (\textit{Instruction Violation Rate} is lower-is-better). Metric definitions in \Cref{sec:evaluation}. Standard errors in \Cref{tab:ood_uniform_appendix}.}
  \label{tab:ood_uniform_main}
\end{table}

\begin{table}[h]
  \centering
  \small
  \setlength{\tabcolsep}{4pt}
  \resizebox{\linewidth}{!}{%
\begin{tabular}{lccccccc}
\toprule
Model & Params & Reward $\uparrow$ & \makecell{Seller Surplus\\Extraction\\Ratio} & \makecell{Item\\Deal\\Rate} & \makecell{Deal on\\Top Buyer\\Rate} & \makecell{Optimal-Pair\\Offer Coverage\\Rate} & \makecell{Instruction\\Violation\\Rate} \\
\midrule
\rowcolor{gray!10} Qwen3-30B-A3B-Instruct-2507-trained (\textbf{Ours}) & 30B & $\mathbf{+0.444 \pm 0.022}$ & $\mathbf{+0.493 \pm 0.013}$ & $70.8\% \pm 1.3\%$ & $\mathbf{0.594 \pm 0.017}$ & $\mathbf{0.699 \pm 0.014}$ & $3.3\% \pm 1.2\%$ \\
\rowcolor{gray!10} Qwen3-30B-A3B-Instruct-2507-untrained & 30B & $-0.187 \pm 0.039$ & $+0.240 \pm 0.012$ & $74.1\% \pm 1.7\%$ & $0.384 \pm 0.020$ & $0.378 \pm 0.018$ & $34.5\% \pm 3.1\%$ \\
\midrule
Kimi-K2.6-thinking & 1T & $\underline{+0.322 \pm 0.016}$ & $+0.345 \pm 0.012$ & $75.9\% \pm 1.4\%$ & $0.398 \pm 0.016$ & $0.402 \pm 0.015$ & $\underline{1.7\% \pm 0.8\%}$ \\
DeepSeek-V4-Pro-nothink & 1.6T & $+0.308 \pm 0.018$ & $+0.347 \pm 0.011$ & $81.5\% \pm 1.1\%$ & $0.417 \pm 0.016$ & $0.426 \pm 0.015$ & $2.9\% \pm 1.1\%$ \\
DeepSeek-V4-Flash-thinking & 284B & $+0.268 \pm 0.024$ & $+0.353 \pm 0.011$ & $77.1\% \pm 1.3\%$ & $0.388 \pm 0.017$ & $0.398 \pm 0.015$ & $6.3\% \pm 1.6\%$ \\
GPT-5.4-high-reasoning & closed-source & $+0.265 \pm 0.010$ & $+0.265 \pm 0.010$ & $\mathbf{82.8\% \pm 1.1\%}$ & $0.363 \pm 0.016$ & $0.439 \pm 0.015$ & $\mathbf{0.0\% \pm 0.0\%}$ \\
DeepSeek-V4-Flash-nothink & 284B & $+0.204 \pm 0.027$ & $+0.326 \pm 0.012$ & $78.0\% \pm 1.2\%$ & $0.392 \pm 0.017$ & $0.422 \pm 0.015$ & $9.2\% \pm 1.9\%$ \\
Kimi-K2.6-nothink & 1T & $+0.140 \pm 0.032$ & $+0.328 \pm 0.011$ & $\underline{82.0\% \pm 1.1\%}$ & $\underline{0.426 \pm 0.017}$ & $0.420 \pm 0.016$ & $14.2\% \pm 2.3\%$ \\
DeepSeek-V4-Pro-thinking & 1.6T & $+0.133 \pm 0.037$ & $\underline{+0.396 \pm 0.012}$ & $79.0\% \pm 1.3\%$ & $0.418 \pm 0.018$ & $\underline{0.444 \pm 0.016}$ & $18.8\% \pm 2.5\%$ \\
GPT-5.4-mini-high-reasoning & closed-source & $+0.035 \pm 0.029$ & $+0.212 \pm 0.010$ & $78.3\% \pm 1.5\%$ & $0.354 \pm 0.017$ & $0.381 \pm 0.016$ & $14.6\% \pm 2.3\%$ \\
\bottomrule
\end{tabular}
  }
  \caption{\textbf{Out-of-distribution ($5$ buyers, $5$ items, $T=12$, uniform valuation), full results.}}
  \label{tab:ood_uniform_appendix}
\end{table}

\begin{table}[h]
  \centering
  \small
  \setlength{\tabcolsep}{4pt}
  \resizebox{\linewidth}{!}{%
\begin{tabular}{lccccccc}
\toprule
Model & Params & Reward $\uparrow$ & \makecell{Seller Surplus\\Extraction\\Ratio} & \makecell{Item\\Deal\\Rate} & \makecell{Deal on\\Top Buyer\\Rate} & \makecell{Optimal-Pair\\Offer Coverage\\Rate} & \makecell{Instruction\\Violation\\Rate} \\
\midrule
\rowcolor{gray!10} Qwen3-30B-A3B-Instruct-2507-trained (\textbf{Ours}) & 30B & $\mathbf{+0.438}$ & $\mathbf{+0.483}$ & $69.5\%$ & $\mathbf{0.550}$ & $\mathbf{0.699}$ & $3.0\%$ \\
\rowcolor{gray!10} Qwen3-30B-A3B-Instruct-2507-untrained & 30B & $-0.129$ & $+0.253$ & $75.4\%$ & $0.349$ & $0.329$ & $30.5\%$ \\
\midrule
Kimi-K2.6-thinking & 1T & $\underline{+0.303}$ & $+0.337$ & $74.8\%$ & $0.418$ & $0.406$ & $\underline{2.5\%}$ \\
DeepSeek-V4-Pro-nothink & 1.6T & $+0.281$ & $+0.371$ & $\mathbf{81.9\%}$ & $0.433$ & $\underline{0.459}$ & $6.5\%$ \\
GPT-5.4-high-reasoning & closed-source & $+0.260$ & $+0.260$ & $\underline{81.5\%}$ & $0.378$ & $0.458$ & $\mathbf{0.0\%}$ \\
DeepSeek-V4-Flash-nothink & 284B & $+0.238$ & $+0.310$ & $78.2\%$ & $0.367$ & $0.419$ & $5.5\%$ \\
DeepSeek-V4-Flash-thinking & 284B & $+0.208$ & $+0.327$ & $74.1\%$ & $0.384$ & $0.409$ & $9.0\%$ \\
Kimi-K2.6-nothink & 1T & $+0.148$ & $+0.312$ & $81.3\%$ & $0.366$ & $0.370$ & $12.5\%$ \\
DeepSeek-V4-Pro-thinking & 1.6T & $+0.141$ & $\underline{+0.384}$ & $77.7\%$ & $\underline{0.444}$ & $0.426$ & $17.6\%$ \\
GPT-5.4-mini-high-reasoning & closed-source & $-0.042$ & $+0.198$ & $73.1\%$ & $0.366$ & $0.357$ & $20.0\%$ \\
\bottomrule
\end{tabular}
  }
  \caption{\textbf{Out-of-distribution Env.~2, Expanded Musical (held-out Musical Instruments category)} ($5$ buyers, $5$ items, $T=12$, uniform valuation). Baselines sorted by \textit{Reward} (descending); best per column in bold, second-best underlined (\textit{Instruction Violation Rate} is lower-is-better). Metric definitions in \Cref{sec:evaluation}. Standard errors in \Cref{tab:ood_uniform_mi_appendix}.}
  \label{tab:ood_uniform_mi_main}
\end{table}

\begin{table}[h]
  \centering
  \small
  \setlength{\tabcolsep}{4pt}
  \resizebox{\linewidth}{!}{%
\begin{tabular}{lccccccc}
\toprule
Model & Params & Reward $\uparrow$ & \makecell{Seller Surplus\\Extraction\\Ratio} & \makecell{Item\\Deal\\Rate} & \makecell{Deal on\\Top Buyer\\Rate} & \makecell{Optimal-Pair\\Offer Coverage\\Rate} & \makecell{Instruction\\Violation\\Rate} \\
\midrule
\rowcolor{gray!10} Qwen3-30B-A3B-Instruct-2507-trained (\textbf{Ours}) & 30B & $\mathbf{+0.438 \pm 0.023}$ & $\mathbf{+0.483 \pm 0.015}$ & $69.5\% \pm 1.5\%$ & $\mathbf{0.550 \pm 0.019}$ & $\mathbf{0.699 \pm 0.015}$ & $3.0\% \pm 1.2\%$ \\
\rowcolor{gray!10} Qwen3-30B-A3B-Instruct-2507-untrained & 30B & $-0.129 \pm 0.042$ & $+0.253 \pm 0.013$ & $75.4\% \pm 1.7\%$ & $0.349 \pm 0.021$ & $0.329 \pm 0.018$ & $30.5\% \pm 3.3\%$ \\
\midrule
Kimi-K2.6-thinking & 1T & $\underline{+0.303 \pm 0.020}$ & $+0.337 \pm 0.014$ & $74.8\% \pm 1.6\%$ & $0.418 \pm 0.018$ & $0.406 \pm 0.016$ & $\underline{2.5\% \pm 1.1\%}$ \\
DeepSeek-V4-Pro-nothink & 1.6T & $+0.281 \pm 0.027$ & $+0.371 \pm 0.013$ & $\mathbf{81.9\% \pm 1.2\%}$ & $0.433 \pm 0.018$ & $\underline{0.459 \pm 0.017}$ & $6.5\% \pm 1.7\%$ \\
GPT-5.4-high-reasoning & closed-source & $+0.260 \pm 0.012$ & $+0.260 \pm 0.012$ & $\underline{81.5\% \pm 1.2\%}$ & $0.378 \pm 0.017$ & $0.458 \pm 0.017$ & $\mathbf{0.0\% \pm 0.0\%}$ \\
DeepSeek-V4-Flash-nothink & 284B & $+0.238 \pm 0.025$ & $+0.310 \pm 0.013$ & $78.2\% \pm 1.3\%$ & $0.367 \pm 0.018$ & $0.419 \pm 0.017$ & $5.5\% \pm 1.6\%$ \\
DeepSeek-V4-Flash-thinking & 284B & $+0.208 \pm 0.029$ & $+0.327 \pm 0.013$ & $74.1\% \pm 1.4\%$ & $0.384 \pm 0.019$ & $0.409 \pm 0.017$ & $9.0\% \pm 2.0\%$ \\
Kimi-K2.6-nothink & 1T & $+0.148 \pm 0.033$ & $+0.312 \pm 0.013$ & $81.3\% \pm 1.3\%$ & $0.366 \pm 0.018$ & $0.370 \pm 0.017$ & $12.5\% \pm 2.3\%$ \\
DeepSeek-V4-Pro-thinking & 1.6T & $+0.141 \pm 0.039$ & $\underline{+0.384 \pm 0.015}$ & $77.7\% \pm 1.5\%$ & $\underline{0.444 \pm 0.020}$ & $0.426 \pm 0.018$ & $17.6\% \pm 2.7\%$ \\
GPT-5.4-mini-high-reasoning & closed-source & $-0.042 \pm 0.035$ & $+0.198 \pm 0.012$ & $73.1\% \pm 1.6\%$ & $0.366 \pm 0.020$ & $0.357 \pm 0.017$ & $20.0\% \pm 2.8\%$ \\
\bottomrule
\end{tabular}
  }
  \caption{\textbf{Out-of-distribution on held-out Musical Instruments ($5$/$5$/$T=12$), full results.}}
  \label{tab:ood_uniform_mi_appendix}
\end{table}

\begin{table}[h]
  \centering
  \small
  \setlength{\tabcolsep}{4pt}
  \resizebox{\linewidth}{!}{%
\begin{tabular}{lccccccc}
\toprule
Model & Params & Reward $\uparrow$ & \makecell{Seller Surplus\\Extraction\\Ratio} & \makecell{Item\\Deal\\Rate} & \makecell{Deal on\\Top Buyer\\Rate} & \makecell{Optimal-Pair\\Offer Coverage\\Rate} & \makecell{Instruction\\Violation\\Rate} \\
\midrule
\rowcolor{gray!10} Qwen3-30B-A3B-Instruct-2507-trained (\textbf{Ours}) & 30B & $\mathbf{+0.559}$ & $\mathbf{+0.586}$ & $80.6\%$ & $\mathbf{0.534}$ & $\mathbf{0.727}$ & $1.7\%$ \\
\rowcolor{gray!10} Qwen3-30B-A3B-Instruct-2507-untrained & 30B & $-0.065$ & $+0.313$ & $89.5\%$ & $0.318$ & $0.450$ & $28.7\%$ \\
\midrule
Kimi-K2.6-thinking & 1T & $\underline{+0.443}$ & $+0.449$ & $90.7\%$ & $0.403$ & $0.508$ & $\underline{0.4\%}$ \\
GPT-5.4-high-reasoning & closed-source & $+0.400$ & $+0.400$ & $\mathbf{95.0\%}$ & $0.412$ & $\underline{0.553}$ & $\mathbf{0.0\%}$ \\
DeepSeek-V4-Pro-thinking & 1.6T & $+0.367$ & $\underline{+0.505}$ & $93.3\%$ & $0.420$ & $0.534$ & $9.2\%$ \\
DeepSeek-V4-Pro-nothink & 1.6T & $+0.337$ & $+0.414$ & $\underline{94.6\%}$ & $0.394$ & $0.504$ & $5.4\%$ \\
Kimi-K2.6-nothink & 1T & $+0.322$ & $+0.429$ & $\mathbf{95.0\%}$ & $\underline{0.422}$ & $0.517$ & $7.5\%$ \\
DeepSeek-V4-Flash-thinking & 284B & $+0.298$ & $+0.435$ & $92.8\%$ & $0.397$ & $0.465$ & $9.6\%$ \\
DeepSeek-V4-Flash-nothink & 284B & $+0.287$ & $+0.373$ & $93.9\%$ & $0.383$ & $0.479$ & $6.2\%$ \\
GPT-5.4-mini-high-reasoning & closed-source & $+0.193$ & $+0.284$ & $92.5\%$ & $0.365$ & $0.447$ & $7.1\%$ \\
\bottomrule
\end{tabular}
  }
  \caption{\textbf{Out-of-distribution Env.~3, Buyer Heavy ($5$ buyers, $3$ items, $T=12$)}, uniform valuation. Baselines sorted by \textit{Reward} (descending); best per column in bold, second-best underlined (\textit{Instruction Violation Rate} is lower-is-better). Metric definitions in \Cref{sec:evaluation}. Standard errors in \Cref{tab:ood_5b3i_appendix}.}
  \label{tab:ood_5b3i_main}
\end{table}

\begin{table}[h]
  \centering
  \small
  \setlength{\tabcolsep}{4pt}
  \resizebox{\linewidth}{!}{%
\begin{tabular}{lccccccc}
\toprule
Model & Params & Reward $\uparrow$ & \makecell{Seller Surplus\\Extraction\\Ratio} & \makecell{Item\\Deal\\Rate} & \makecell{Deal on\\Top Buyer\\Rate} & \makecell{Optimal-Pair\\Offer Coverage\\Rate} & \makecell{Instruction\\Violation\\Rate} \\
\midrule
\rowcolor{gray!10} Qwen3-30B-A3B-Instruct-2507-trained (\textbf{Ours}) & 30B & $\mathbf{+0.559 \pm 0.022}$ & $\mathbf{+0.586 \pm 0.018}$ & $80.6\% \pm 1.5\%$ & $\mathbf{0.534 \pm 0.021}$ & $\mathbf{0.727 \pm 0.017}$ & $1.7\% \pm 0.8\%$ \\
\rowcolor{gray!10} Qwen3-30B-A3B-Instruct-2507-untrained & 30B & $-0.065 \pm 0.040$ & $+0.313 \pm 0.015$ & $89.5\% \pm 1.6\%$ & $0.318 \pm 0.022$ & $0.450 \pm 0.022$ & $28.7\% \pm 2.9\%$ \\
\midrule
Kimi-K2.6-thinking & 1T & $\underline{+0.443 \pm 0.015}$ & $+0.449 \pm 0.014$ & $90.7\% \pm 1.3\%$ & $0.403 \pm 0.019$ & $0.508 \pm 0.019$ & $\underline{0.4\% \pm 0.4\%}$ \\
GPT-5.4-high-reasoning & closed-source & $+0.400 \pm 0.014$ & $+0.400 \pm 0.014$ & $\mathbf{95.0\% \pm 0.9\%}$ & $0.412 \pm 0.019$ & $\underline{0.553 \pm 0.019}$ & $\mathbf{0.0\% \pm 0.0\%}$ \\
DeepSeek-V4-Pro-thinking & 1.6T & $+0.367 \pm 0.031$ & $\underline{+0.505 \pm 0.014}$ & $93.3\% \pm 1.0\%$ & $0.420 \pm 0.020$ & $0.534 \pm 0.020$ & $9.2\% \pm 1.9\%$ \\
DeepSeek-V4-Pro-nothink & 1.6T & $+0.337 \pm 0.025$ & $+0.414 \pm 0.015$ & $\underline{94.6\% \pm 1.0\%}$ & $0.394 \pm 0.019$ & $0.504 \pm 0.019$ & $5.4\% \pm 1.5\%$ \\
Kimi-K2.6-nothink & 1T & $+0.322 \pm 0.027$ & $+0.429 \pm 0.014$ & $\mathbf{95.0\% \pm 0.9\%}$ & $\underline{0.422 \pm 0.020}$ & $0.517 \pm 0.019$ & $7.5\% \pm 1.7\%$ \\
DeepSeek-V4-Flash-thinking & 284B & $+0.298 \pm 0.031$ & $+0.435 \pm 0.015$ & $92.8\% \pm 1.1\%$ & $0.397 \pm 0.020$ & $0.465 \pm 0.020$ & $9.6\% \pm 1.9\%$ \\
DeepSeek-V4-Flash-nothink & 284B & $+0.287 \pm 0.025$ & $+0.373 \pm 0.014$ & $93.9\% \pm 1.0\%$ & $0.383 \pm 0.019$ & $0.479 \pm 0.019$ & $6.2\% \pm 1.6\%$ \\
GPT-5.4-mini-high-reasoning & closed-source & $+0.193 \pm 0.025$ & $+0.284 \pm 0.014$ & $92.5\% \pm 1.2\%$ & $0.365 \pm 0.019$ & $0.447 \pm 0.019$ & $7.1\% \pm 1.7\%$ \\
\bottomrule
\end{tabular}
  }
  \caption{\textbf{Out-of-distribution ($5$ buyers, $3$ items, $T=12$), full results.}}
  \label{tab:ood_5b3i_appendix}
\end{table}

\begin{table}[h]
  \centering
  \small
  \setlength{\tabcolsep}{4pt}
  \resizebox{\linewidth}{!}{%
\begin{tabular}{lccccccc}
\toprule
Model & Params & Reward $\uparrow$ & \makecell{Seller Surplus\\Extraction\\Ratio} & \makecell{Item\\Deal\\Rate} & \makecell{Deal on\\Top Buyer\\Rate} & \makecell{Optimal-Pair\\Offer Coverage\\Rate} & \makecell{Instruction\\Violation\\Rate} \\
\midrule
\rowcolor{gray!10} Qwen3-30B-A3B-Instruct-2507-trained (\textbf{Ours}) & 30B & $\mathbf{+0.326}$ & $\mathbf{+0.343}$ & $46.4\%$ & $\mathbf{0.728}$ & $\mathbf{0.817}$ & $1.3\%$ \\
\rowcolor{gray!10} Qwen3-30B-A3B-Instruct-2507-untrained & 30B & $-0.089$ & $+0.189$ & $46.8\%$ & $0.568$ & $0.377$ & $23.3\%$ \\
\midrule
Kimi-K2.6-thinking & 1T & $\underline{+0.238}$ & $+0.248$ & $49.6\%$ & $0.573$ & $0.465$ & $\underline{0.8\%}$ \\
DeepSeek-V4-Pro-nothink & 1.6T & $+0.213$ & $+0.239$ & $\mathbf{51.7\%}$ & $0.580$ & $0.520$ & $2.1\%$ \\
Kimi-K2.6-nothink & 1T & $+0.211$ & $+0.253$ & $\underline{51.4\%}$ & $0.596$ & $0.527$ & $3.3\%$ \\
GPT-5.4-high-reasoning & closed-source & $+0.195$ & $+0.195$ & $50.4\%$ & $0.513$ & $0.503$ & $\mathbf{0.0\%}$ \\
DeepSeek-V4-Flash-nothink & 284B & $+0.182$ & $+0.234$ & $49.6\%$ & $0.546$ & $0.476$ & $4.2\%$ \\
DeepSeek-V4-Flash-thinking & 284B & $+0.097$ & $+0.248$ & $47.6\%$ & $0.536$ & $0.472$ & $12.1\%$ \\
GPT-5.4-mini-high-reasoning & closed-source & $+0.049$ & $+0.144$ & $48.7\%$ & $0.478$ & $0.416$ & $8.3\%$ \\
DeepSeek-V4-Pro-thinking & 1.6T & $-0.183$ & $\underline{+0.289}$ & $50.3\%$ & $\underline{0.610}$ & $\underline{0.571}$ & $36.7\%$ \\
\bottomrule
\end{tabular}
  }
  \caption{\textbf{Out-of-distribution Env.~4, Item Heavy ($3$ buyers, $5$ items, $T=7$)}, uniform valuation. Baselines sorted by \textit{Reward} (descending); best per column in bold, second-best underlined (\textit{Instruction Violation Rate} is lower-is-better). Metric definitions in \Cref{sec:evaluation}. Standard errors in \Cref{tab:ood_3b5i_appendix}.}
  \label{tab:ood_3b5i_main}
\end{table}

\begin{table}[h]
  \centering
  \small
  \setlength{\tabcolsep}{4pt}
  \resizebox{\linewidth}{!}{%
\begin{tabular}{lccccccc}
\toprule
Model & Params & Reward $\uparrow$ & \makecell{Seller Surplus\\Extraction\\Ratio} & \makecell{Item\\Deal\\Rate} & \makecell{Deal on\\Top Buyer\\Rate} & \makecell{Optimal-Pair\\Offer Coverage\\Rate} & \makecell{Instruction\\Violation\\Rate} \\
\midrule
\rowcolor{gray!10} Qwen3-30B-A3B-Instruct-2507-trained (\textbf{Ours}) & 30B & $\mathbf{+0.326 \pm 0.015}$ & $\mathbf{+0.343 \pm 0.012}$ & $46.4\% \pm 1.0\%$ & $\mathbf{0.728 \pm 0.019}$ & $\mathbf{0.817 \pm 0.015}$ & $1.3\% \pm 0.7\%$ \\
\rowcolor{gray!10} Qwen3-30B-A3B-Instruct-2507-untrained & 30B & $-0.089 \pm 0.034$ & $+0.189 \pm 0.011$ & $46.8\% \pm 1.2\%$ & $0.568 \pm 0.024$ & $0.377 \pm 0.021$ & $23.3\% \pm 2.7\%$ \\
\midrule
Kimi-K2.6-thinking & 1T & $\underline{+0.238 \pm 0.012}$ & $+0.248 \pm 0.009$ & $49.6\% \pm 0.9\%$ & $0.573 \pm 0.020$ & $0.465 \pm 0.019$ & $\underline{0.8\% \pm 0.6\%}$ \\
DeepSeek-V4-Pro-nothink & 1.6T & $+0.213 \pm 0.015$ & $+0.239 \pm 0.010$ & $\mathbf{51.7\% \pm 0.8\%}$ & $0.580 \pm 0.020$ & $0.520 \pm 0.019$ & $2.1\% \pm 0.9\%$ \\
Kimi-K2.6-nothink & 1T & $+0.211 \pm 0.017$ & $+0.253 \pm 0.009$ & $\underline{51.4\% \pm 0.9\%}$ & $0.596 \pm 0.020$ & $0.527 \pm 0.019$ & $3.3\% \pm 1.2\%$ \\
GPT-5.4-high-reasoning & closed-source & $+0.195 \pm 0.009$ & $+0.195 \pm 0.009$ & $50.4\% \pm 0.9\%$ & $0.513 \pm 0.020$ & $0.503 \pm 0.019$ & $\mathbf{0.0\% \pm 0.0\%}$ \\
DeepSeek-V4-Flash-nothink & 284B & $+0.182 \pm 0.019$ & $+0.234 \pm 0.010$ & $49.6\% \pm 0.9\%$ & $0.546 \pm 0.021$ & $0.476 \pm 0.019$ & $4.2\% \pm 1.3\%$ \\
DeepSeek-V4-Flash-thinking & 284B & $+0.097 \pm 0.028$ & $+0.248 \pm 0.011$ & $47.6\% \pm 1.1\%$ & $0.536 \pm 0.022$ & $0.472 \pm 0.020$ & $12.1\% \pm 2.1\%$ \\
GPT-5.4-mini-high-reasoning & closed-source & $+0.049 \pm 0.022$ & $+0.144 \pm 0.009$ & $48.7\% \pm 1.0\%$ & $0.478 \pm 0.022$ & $0.416 \pm 0.019$ & $8.3\% \pm 1.8\%$ \\
DeepSeek-V4-Pro-thinking & 1.6T & $-0.183 \pm 0.041$ & $\underline{+0.289 \pm 0.013}$ & $50.3\% \pm 1.1\%$ & $\underline{0.610 \pm 0.025}$ & $\underline{0.571 \pm 0.023}$ & $36.7\% \pm 3.1\%$ \\
\bottomrule
\end{tabular}
  }
  \caption{\textbf{Out-of-distribution ($3$ buyers, $5$ items, $T=7$), full results.}}
  \label{tab:ood_3b5i_appendix}
\end{table}

\begin{table}[h]
  \centering
  \small
  \setlength{\tabcolsep}{4pt}
  \resizebox{\linewidth}{!}{%
\begin{tabular}{lccccccc}
\toprule
Model & Params & Reward $\uparrow$ & \makecell{Seller Surplus\\Extraction\\Ratio} & \makecell{Item\\Deal\\Rate} & \makecell{Deal on\\Top Buyer\\Rate} & \makecell{Optimal-Pair\\Offer Coverage\\Rate} & \makecell{Instruction\\Violation\\Rate} \\
\midrule
\rowcolor{gray!10} Qwen3-30B-A3B-Instruct-2507-trained (\textbf{Ours}) & 30B & $\mathbf{+0.452}$ & $\mathbf{+0.497}$ & $69.9\%$ & $\mathbf{0.366}$ & $\mathbf{0.737}$ & $3.0\%$ \\
\rowcolor{gray!10} Qwen3-30B-A3B-Instruct-2507-untrained & 30B & $-0.215$ & $+0.260$ & $74.4\%$ & $0.276$ & $0.446$ & $37.7\%$ \\
\midrule
Kimi-K2.6-thinking & 1T & $\underline{+0.317}$ & $+0.334$ & $73.6\%$ & $0.289$ & $0.481$ & $\underline{1.3\%}$ \\
DeepSeek-V4-Pro-nothink & 1.6T & $+0.293$ & $+0.337$ & $77.8\%$ & $\underline{0.313}$ & $0.515$ & $3.3\%$ \\
GPT-5.4-high-reasoning & closed-source & $+0.272$ & $+0.278$ & $\mathbf{79.8\%}$ & $0.311$ & $\underline{0.547}$ & $\mathbf{0.4\%}$ \\
DeepSeek-V4-Pro-thinking & 1.6T & $+0.251$ & $\underline{+0.405}$ & $75.9\%$ & $0.288$ & $0.519$ & $10.9\%$ \\
DeepSeek-V4-Flash-nothink & 284B & $+0.187$ & $+0.319$ & $78.3\%$ & $0.280$ & $0.490$ & $10.0\%$ \\
Kimi-K2.6-nothink & 1T & $+0.176$ & $+0.338$ & $\underline{79.0\%}$ & $0.289$ & $0.493$ & $12.1\%$ \\
DeepSeek-V4-Flash-thinking & 284B & $+0.088$ & $+0.339$ & $75.0\%$ & $0.285$ & $0.498$ & $18.8\%$ \\
GPT-5.4-mini-high-reasoning & closed-source & $+0.053$ & $+0.215$ & $75.6\%$ & $0.276$ & $0.479$ & $13.3\%$ \\
\bottomrule
\end{tabular}
  }
  \caption{\textbf{Out-of-distribution Env.~5, Item Correlation: per-item valuation multiplier} $\mu_j \sim \mathcal{U}[0.5,2]$ ($5$ buyers, $5$ items, $T=12$). Baselines sorted by \textit{Reward} (descending); best per column in bold, second-best underlined (\textit{Instruction Violation Rate} is lower-is-better). Metric definitions in \Cref{sec:evaluation}. Standard errors in \Cref{tab:ood_itemmult_appendix}.}
  \label{tab:ood_itemmult_main}
\end{table}

\begin{table}[h]
  \centering
  \small
  \setlength{\tabcolsep}{4pt}
  \resizebox{\linewidth}{!}{%
\begin{tabular}{lccccccc}
\toprule
Model & Params & Reward $\uparrow$ & \makecell{Seller Surplus\\Extraction\\Ratio} & \makecell{Item\\Deal\\Rate} & \makecell{Deal on\\Top Buyer\\Rate} & \makecell{Optimal-Pair\\Offer Coverage\\Rate} & \makecell{Instruction\\Violation\\Rate} \\
\midrule
\rowcolor{gray!10} Qwen3-30B-A3B-Instruct-2507-trained (\textbf{Ours}) & 30B & $\mathbf{+0.452 \pm 0.022}$ & $\mathbf{+0.497 \pm 0.015}$ & $69.9\% \pm 1.3\%$ & $\mathbf{0.366 \pm 0.017}$ & $\mathbf{0.737 \pm 0.014}$ & $3.0\% \pm 1.1\%$ \\
\rowcolor{gray!10} Qwen3-30B-A3B-Instruct-2507-untrained & 30B & $-0.215 \pm 0.040$ & $+0.260 \pm 0.014$ & $74.4\% \pm 1.8\%$ & $0.276 \pm 0.019$ & $0.446 \pm 0.019$ & $37.7\% \pm 3.1\%$ \\
\midrule
Kimi-K2.6-thinking & 1T & $\underline{+0.317 \pm 0.015}$ & $+0.334 \pm 0.011$ & $73.6\% \pm 1.5\%$ & $0.289 \pm 0.016$ & $0.481 \pm 0.015$ & $\underline{1.3\% \pm 0.7\%}$ \\
DeepSeek-V4-Pro-nothink & 1.6T & $+0.293 \pm 0.019$ & $+0.337 \pm 0.011$ & $77.8\% \pm 1.2\%$ & $\underline{0.313 \pm 0.015}$ & $0.515 \pm 0.015$ & $3.3\% \pm 1.2\%$ \\
GPT-5.4-high-reasoning & closed-source & $+0.272 \pm 0.012$ & $+0.278 \pm 0.011$ & $\mathbf{79.8\% \pm 1.3\%}$ & $0.311 \pm 0.015$ & $\underline{0.547 \pm 0.016}$ & $\mathbf{0.4\% \pm 0.4\%}$ \\
DeepSeek-V4-Pro-thinking & 1.6T & $+0.251 \pm 0.030$ & $\underline{+0.405 \pm 0.012}$ & $75.9\% \pm 1.3\%$ & $0.288 \pm 0.016$ & $0.519 \pm 0.016$ & $10.9\% \pm 2.0\%$ \\
DeepSeek-V4-Flash-nothink & 284B & $+0.187 \pm 0.028$ & $+0.319 \pm 0.012$ & $78.3\% \pm 1.3\%$ & $0.280 \pm 0.015$ & $0.490 \pm 0.016$ & $10.0\% \pm 1.9\%$ \\
Kimi-K2.6-nothink & 1T & $+0.176 \pm 0.030$ & $+0.338 \pm 0.012$ & $\underline{79.0\% \pm 1.2\%}$ & $0.289 \pm 0.016$ & $0.493 \pm 0.016$ & $12.1\% \pm 2.1\%$ \\
DeepSeek-V4-Flash-thinking & 284B & $+0.088 \pm 0.035$ & $+0.339 \pm 0.013$ & $75.0\% \pm 1.4\%$ & $0.285 \pm 0.017$ & $0.498 \pm 0.017$ & $18.8\% \pm 2.5\%$ \\
GPT-5.4-mini-high-reasoning & closed-source & $+0.053 \pm 0.028$ & $+0.215 \pm 0.010$ & $75.6\% \pm 1.4\%$ & $0.276 \pm 0.016$ & $0.479 \pm 0.016$ & $13.3\% \pm 2.2\%$ \\
\bottomrule
\end{tabular}
  }
  \caption{\textbf{Out-of-distribution with per-item valuation multiplier $\mu_j \sim \mathcal{U}[0.5,2]$, full results.}}
  \label{tab:ood_itemmult_appendix}
\end{table}

\begin{table}[h]
  \centering
  \small
  \setlength{\tabcolsep}{4pt}
  \resizebox{\linewidth}{!}{%
\begin{tabular}{lccccccc}
\toprule
Model & Params & Reward $\uparrow$ & \makecell{Seller Surplus\\Extraction\\Ratio} & \makecell{Item\\Deal\\Rate} & \makecell{Deal on\\Top Buyer\\Rate} & \makecell{Optimal-Pair\\Offer Coverage\\Rate} & \makecell{Instruction\\Violation\\Rate} \\
\midrule
\rowcolor{gray!10} Qwen3-30B-A3B-Instruct-2507-trained (\textbf{Ours}) & 30B & $\mathbf{+0.449}$ & $\mathbf{+0.499}$ & $70.7\%$ & $\mathbf{0.372}$ & $\mathbf{0.739}$ & $3.3\%$ \\
\rowcolor{gray!10} Qwen3-30B-A3B-Instruct-2507-untrained & 30B & $-0.150$ & $+0.299$ & $74.1\%$ & $0.289$ & $0.409$ & $34.6\%$ \\
\midrule
Kimi-K2.6-thinking & 1T & $\underline{+0.349}$ & $+0.372$ & $74.5\%$ & $\underline{0.292}$ & $0.500$ & $\underline{1.7\%}$ \\
DeepSeek-V4-Pro-nothink & 1.6T & $+0.304$ & $+0.361$ & $\underline{80.0\%}$ & $0.263$ & $0.482$ & $4.2\%$ \\
GPT-5.4-high-reasoning & closed-source & $+0.297$ & $+0.297$ & $79.5\%$ & $0.283$ & $\underline{0.536}$ & $\mathbf{0.0\%}$ \\
DeepSeek-V4-Pro-thinking & 1.6T & $+0.293$ & $\underline{+0.444}$ & $79.8\%$ & $0.274$ & $0.512$ & $10.5\%$ \\
DeepSeek-V4-Flash-nothink & 284B & $+0.271$ & $+0.344$ & $79.1\%$ & $0.274$ & $0.485$ & $5.4\%$ \\
DeepSeek-V4-Flash-thinking & 284B & $+0.266$ & $+0.394$ & $78.0\%$ & $0.258$ & $0.489$ & $9.2\%$ \\
Kimi-K2.6-nothink & 1T & $+0.184$ & $+0.359$ & $\mathbf{81.8\%}$ & $0.253$ & $0.505$ & $12.9\%$ \\
GPT-5.4-mini-high-reasoning & closed-source & $+0.066$ & $+0.224$ & $75.9\%$ & $0.260$ & $0.467$ & $12.9\%$ \\
\bottomrule
\end{tabular}
  }
  \caption{\textbf{Out-of-distribution Env.~6, Buyer Correlation: per-buyer generosity multiplier} $\gamma_i \sim \mathcal{U}[0.5,2]$ ($5$ buyers, $5$ items, $T=12$). Baselines sorted by \textit{Reward} (descending); best per column in bold, second-best underlined (\textit{Instruction Violation Rate} is lower-is-better). Metric definitions in \Cref{sec:evaluation}. Standard errors in \Cref{tab:ood_buyergen_appendix}.}
  \label{tab:ood_buyergen_main}
\end{table}

\begin{table}[h]
  \centering
  \small
  \setlength{\tabcolsep}{4pt}
  \resizebox{\linewidth}{!}{%
\begin{tabular}{lccccccc}
\toprule
Model & Params & Reward $\uparrow$ & \makecell{Seller Surplus\\Extraction\\Ratio} & \makecell{Item\\Deal\\Rate} & \makecell{Deal on\\Top Buyer\\Rate} & \makecell{Optimal-Pair\\Offer Coverage\\Rate} & \makecell{Instruction\\Violation\\Rate} \\
\midrule
\rowcolor{gray!10} Qwen3-30B-A3B-Instruct-2507-trained (\textbf{Ours}) & 30B & $\mathbf{+0.449 \pm 0.022}$ & $\mathbf{+0.499 \pm 0.015}$ & $70.7\% \pm 1.4\%$ & $\mathbf{0.372 \pm 0.017}$ & $\mathbf{0.739 \pm 0.014}$ & $3.3\% \pm 1.2\%$ \\
\rowcolor{gray!10} Qwen3-30B-A3B-Instruct-2507-untrained & 30B & $-0.150 \pm 0.041$ & $+0.299 \pm 0.015$ & $74.1\% \pm 1.6\%$ & $0.289 \pm 0.019$ & $0.409 \pm 0.019$ & $34.6\% \pm 3.1\%$ \\
\midrule
Kimi-K2.6-thinking & 1T & $\underline{+0.349 \pm 0.017}$ & $+0.372 \pm 0.012$ & $74.5\% \pm 1.4\%$ & $\underline{0.292 \pm 0.015}$ & $0.500 \pm 0.016$ & $\underline{1.7\% \pm 0.8\%}$ \\
DeepSeek-V4-Pro-nothink & 1.6T & $+0.304 \pm 0.021$ & $+0.361 \pm 0.011$ & $\underline{80.0\% \pm 1.2\%}$ & $0.263 \pm 0.015$ & $0.482 \pm 0.016$ & $4.2\% \pm 1.3\%$ \\
GPT-5.4-high-reasoning & closed-source & $+0.297 \pm 0.012$ & $+0.297 \pm 0.012$ & $79.5\% \pm 1.2\%$ & $0.283 \pm 0.015$ & $\underline{0.536 \pm 0.016}$ & $\mathbf{0.0\% \pm 0.0\%}$ \\
DeepSeek-V4-Pro-thinking & 1.6T & $+0.293 \pm 0.031$ & $\underline{+0.444 \pm 0.013}$ & $79.8\% \pm 1.2\%$ & $0.274 \pm 0.015$ & $0.512 \pm 0.016$ & $10.5\% \pm 2.0\%$ \\
DeepSeek-V4-Flash-nothink & 284B & $+0.271 \pm 0.023$ & $+0.344 \pm 0.012$ & $79.1\% \pm 1.3\%$ & $0.274 \pm 0.015$ & $0.485 \pm 0.016$ & $5.4\% \pm 1.5\%$ \\
DeepSeek-V4-Flash-thinking & 284B & $+0.266 \pm 0.028$ & $+0.394 \pm 0.013$ & $78.0\% \pm 1.2\%$ & $0.258 \pm 0.015$ & $0.489 \pm 0.016$ & $9.2\% \pm 1.9\%$ \\
Kimi-K2.6-nothink & 1T & $+0.184 \pm 0.031$ & $+0.359 \pm 0.012$ & $\mathbf{81.8\% \pm 1.2\%}$ & $0.253 \pm 0.015$ & $0.505 \pm 0.016$ & $12.9\% \pm 2.2\%$ \\
GPT-5.4-mini-high-reasoning & closed-source & $+0.066 \pm 0.028$ & $+0.224 \pm 0.011$ & $75.9\% \pm 1.4\%$ & $0.260 \pm 0.016$ & $0.467 \pm 0.016$ & $12.9\% \pm 2.2\%$ \\
\bottomrule
\end{tabular}
  }
  \caption{\textbf{Out-of-distribution with per-buyer generosity multiplier $\gamma_i \sim \mathcal{U}[0.5,2]$, full results.}}
  \label{tab:ood_buyergen_appendix}
\end{table}

\section{Qualitative Example: Untrained versus Trained Seller on an Identical Instance}
\label{app:transcripts}

This section presents one evaluation instance played end-to-end by two sellers: the untrained base model (\Cref{subsec:failures}) and a trained checkpoint (\Cref{subsec:good_examples}); \Cref{fig:dialogue-excerpt-first-two-turns} in the main text excerpts the first two seller turns of the trained episode. The two episodes share the instance exactly---the same three-item catalog, the same seller costs, and the same hidden valuation matrix $\mathcal{B}$---while the buyer turns are independently sampled rollouts, so the buyers' messages differ across the two transcripts. In this episode no buyer's opening offers were resampled by the first-offer noise. Transcript conventions: \emph{Talk} and \emph{Action} lines are quoted verbatim from the episode records; the seller's \emph{Thought} is quoted verbatim here (the main-text figure shows an abridged version); buyer \emph{Thought} fields are omitted; in the valuation tables, bold marks each item's highest valuation. The seller's \emph{Thought} is private---each buyer sees only the \emph{Talk} and \emph{Action} of the messages targeted at it.

\subsection{Episode Information}
\label{subsec:episode_information}

The seller-side episode constants and item catalog, as the seller prompt presents them (the full dataset records of this catalog appear in \Cref{subsec:dataset_example_cluster}):

\begin{chatmsg}[grey]{Per-episode constants and item catalog (verbatim seller-prompt content)}[breakable]
  \scriptsize
  \texttt{m = 3\ \ buyers\ \ \ \ \ \ n = 3\ \ items}\\
  \texttt{K = 3\ \ per-buyer message cap\ \ \ \ \ \ T = 7\ \ total seller-message budget\ \ (T < m*K)}\\
  \\
  \texttt{\# Item catalog (private — buyers see only list prices, never your cost)}\\
  \texttt{\ \ - video\_recorders\_1: ANNKE 3K Lite 8 Channel Hybrid 5-in-1 Security Digital Video Recorder, H.265+... — list \$169.99, your cost \$55.99}\\
  \texttt{\ \ - video\_recorders\_2: Lorex 4K Ultra HD 8-Channel Fusion Series PoE Network Video Recorder with 2TB... — list \$249.99, your cost \$100.00}\\
  \texttt{\ \ - video\_recorders\_3: ZOSI H.265+ 8 Channel 5MP Lite 1080p CCTV DVR Recorder Without HDD, Hybrid Ca... — list \$55.99, your cost \$42.07}
\end{chatmsg}

The catalog economics of the instance and the realized allocations of the two episodes:

\begin{table}[h]
  \centering
  \footnotesize
  \begin{tabular}{@{}lccc@{}}
    \toprule
     & Item 1 (ANNKE) & Item 2 (Lorex) & Item 3 (ZOSI) \\
    \midrule
    List price $P_{\text{list},j}$ & \$169.99 & \$249.99 & \$55.99 \\
    Seller cost $\mathcal{C}_j$ & \$55.99 & \$100.00 & \$42.07 \\
    \midrule
    Untrained outcome (\Cref{subsec:failures}) & Buyer 2 @ \$100.00 & unsold & Buyer 1 @ \$45.66 \\
    Trained outcome (\Cref{subsec:good_examples}) & Buyer 1 @ \$140.00 & Buyer 2 @ \$240.00 & Buyer 3 @ \$43.99 \\
    \bottomrule
  \end{tabular}
  \caption{Catalog economics and realized allocations for the appendix instance (the hidden valuation matrix appears with each transcript, \Cref{subsec:failures} and \Cref{subsec:good_examples}). Buyer~2 holds the top valuation on \emph{both} high-value items but can close at most one deal, so the surplus-maximizing assignment routes Item~1 to Buyer~1, Item~2 to Buyer~2, and Item~3 to Buyer~3---exactly the allocation the trained seller realizes. The untrained seller instead spends Buyer~1 on the cheapest item and Buyer~2 on the mid item, after which no remaining buyer can afford the highest-margin Lorex.}
  \label{tab:appendix_ground_truth}
\end{table}

\subsection{Untrained Baseline}
\label{subsec:failures}

\begin{center}\small
{\bfseries\color{blue!60!black} Hidden buyer valuations $\mathcal{B}_{i,j}$}\\[0.4em]
  \begin{tabular}{@{}lrrr@{}}
    \toprule
     & Item 1 & Item 2 & Item 3 \\
    \midrule
    Buyer 1 & \$147.60 & \$193.57 & \textbf{\$45.66} \\
    Buyer 2 & \textbf{\$193.18} & \textbf{\$252.12} & \$22.60 \\
    Buyer 3 & \$122.97 & \$149.76 & \$43.99 \\
    \bottomrule
  \end{tabular}
\end{center}

\begingroup
\small
\setlength{\parskip}{0pt}


\noindent\textbf{Turn 1 --- all three buyers open on the cheapest item.}
\begin{chatmsg}[blue]{Buyer 1}[breakable]
  \textbf{Talk:} Let me propose \$45.66 for video\_recorders\_3.\\
  \textbf{Action:} \act{BUY} video\_recorders\_3 \$45.66
\end{chatmsg}
\begin{chatmsg}[blue]{Buyer 2}[breakable]
  \textbf{Talk:} I propose \$10.00 for video\_recorders\_3.\\
  \textbf{Action:} \act{BUY} video\_recorders\_3 \$10.00
\end{chatmsg}
\begin{chatmsg}[blue]{Buyer 3}[breakable]
  \textbf{Talk:} Let’s start with a proposal for the third item. How about \$35 for video\_recorders\_3?\\
  \textbf{Action:} \act{BUY} video\_recorders\_3 \$35.00
\end{chatmsg}
\begin{chatmsg}[green]{Seller (untrained) --- turn 1}[breakable]
  \textbf{Thought:} Buyer 1's offer of \$45.66 for video\_recorders\_3 is above cost (\$42.07) and close to list price (\$55.99), making it the most viable candidate for immediate closure. Buyer 2’s \$10 offer is below cost and thus unacceptable. Buyer 3’s \$35 offer is below cost and also unacceptable. Thus, only Buyer 1’s offer can be accepted. Since all three buyers are active and have bid on the same item, I must act carefully to avoid violating rules. I will close the deal with Buyer 1 on video\_recorders\_3 at \$45.66 and then redirect the others to the remaining items.\\
  \textbf{Target:} Buyer 1\\
  \textbf{Talk:} Thank you for your offer. I accept your proposal of \$45.66 for video\_recorders\_3. The deal is closed.\\
  \textbf{Action:} \act{DEAL} video\_recorders\_3 \$45.66
\end{chatmsg}
\begin{chatmsg}[green]{Seller (untrained) --- turn 1}[breakable]
  \textbf{Thought:} Buyer 1 is satisfied. Buyer 2 and Buyer 3 both offered below cost for video\_recorders\_3, which is unprocessable. I must now pitch the remaining items to them. Buyer 2’s low offer suggests they may be budget-conscious; Buyer 3’s offer was more reasonable but still below cost. Since Buyer 2 has a weak position and no successful bid yet, I should target them first to maximize the chance of a sale. I will pitch video\_recorders\_1, which has a healthy margin (list \$169.99, cost \$55.99), at \$100 — well below list but above cost and a reasonable midpoint. This may hook Buyer 2, who seems eager. Buyer 3 can be approached later if needed.\\
  \textbf{Target:} Buyer 2\\
  \textbf{Talk:} I can offer video\_recorders\_1 at \$100. Would you be interested?\\
  \textbf{Action:} \act{SELL} video\_recorders\_1 \$100.00
\end{chatmsg}

\noindent\textbf{Turn 2.}
\begin{chatmsg}[blue]{Buyer 2}[breakable]
  \textbf{Talk:} Yes, I accept your offer of \$100 for video\_recorders\_1.\\
  \textbf{Action:} \act{DEAL} video\_recorders\_1 \$100.00
\end{chatmsg}
\begin{chatmsg}[green]{Seller (untrained) --- turn 2}[breakable]
  \textbf{Thought:} Buyer 1 and Buyer 2 are satisfied. Only Buyer 3 remains active and has not closed any item. They offered \$35 for video\_recorders\_3, which is below cost (\$42.07), so no deal was possible. However, they are still interested and may be open to other items. I should now pitch video\_recorders\_2 — the remaining item — which has a high list price (\$249.99) and a solid margin (cost \$100.00). I’ll propose \$150, a moderate price that’s well above cost and below list, to test willingness. Since Buyer 3 is the only remaining buyer and has a history of making low bids, this could signal value. If accepted, I maximize revenue; if not, I may need to drop it further, but I must avoid below-cost offers. Proceed with one final pitch.\\
  \textbf{Target:} Buyer 3\\
  \textbf{Talk:} I can offer video\_recorders\_2 at \$150. Would you like to close this deal?\\
  \textbf{Action:} \act{SELL} video\_recorders\_2 \$150.00
\end{chatmsg}

\noindent\textbf{Turn 3 --- the last buyer exits; the Lorex goes unsold.}
\begin{chatmsg}[blue]{Buyer 3}[breakable]
  \textbf{Talk:} I can't accept \$150 — it exceeds my budget. I'll pass.\\
  \textbf{Action:} \act{QUIT}
\end{chatmsg}

\endgroup

\medskip
\noindent\textbf{Outcome.} The untrained seller accepts Buyer~1's opening bid on the cheapest item immediately (\$45.66, a \$3.59 margin), then sells Item~1 to Buyer~2 at a self-proposed \$100---although Buyer~2's hidden valuation for the Lorex (Item~2) was \$252.12. With both buyers \emph{satisfied}, the only remaining buyer, Buyer~3, cannot afford the Lorex (a \$149.76 valuation against the \$150 quote) and quits: the highest-margin item goes unsold. By \eqref{def:multiitem_reward},
$\mathcal{R} = \dfrac{(100 - 55.99) + (45.66 - 42.07)}{(169.99 - 55.99) + (249.99 - 100) + (55.99 - 42.07)} = \dfrac{47.60}{277.91} = 0.171$.

\subsection{Trained Seller}
\label{subsec:good_examples}

\begin{center}\small
{\bfseries\color{blue!60!black} Hidden buyer valuations $\mathcal{B}_{i,j}$}\\[0.4em]
  \begin{tabular}{@{}lrrr@{}}
    \toprule
     & Item 1 & Item 2 & Item 3 \\
    \midrule
    Buyer 1 & \$147.60 & \$193.57 & \textbf{\$45.66} \\
    Buyer 2 & \textbf{\$193.18} & \textbf{\$252.12} & \$22.60 \\
    Buyer 3 & \$122.97 & \$149.76 & \$43.99 \\
    \bottomrule
  \end{tabular}
\end{center}

\begingroup
\small
\setlength{\parskip}{0pt}


\noindent\textbf{Turn 1 --- all three buyers open on the cheapest item.}
\begin{chatmsg}[blue]{Buyer 1}[breakable]
  \textbf{Talk:} How about \$40 for video\_recorders\_3?\\
  \textbf{Action:} \act{BUY} video\_recorders\_3 \$40.00
\end{chatmsg}
\begin{chatmsg}[blue]{Buyer 2}[breakable]
  \textbf{Talk:} I'd like to propose \$15 for video\_recorders\_3 — is that acceptable?\\
  \textbf{Action:} \act{BUY} video\_recorders\_3 \$15.00
\end{chatmsg}
\begin{chatmsg}[blue]{Buyer 3}[breakable]
  \textbf{Talk:} Let’s start with video\_recorders\_3. I propose \$35. Is that acceptable?\\
  \textbf{Action:} \act{BUY} video\_recorders\_3 \$35.00
\end{chatmsg}
\begin{chatmsg}[green]{Seller (trained) --- turn 1}[breakable]
  \textbf{Thought:} Buyer 1's offer of \$40 for item 3 is below my cost (\$42.07) — violation if accepted. Buyer 2's \$15 is also below cost — invalid. Buyer 3's \$35 is still below cost. None of the opening [BUY] offers for video\_recorders\_3 are acceptable. I must reject all three for item 3. However, I can pitch the other two items (1 and 2) to any buyer. Since all buyers are active and none have closed an item yet, I should initiate negotiations on items 1 and 2. Buyer 1 offered the lowest for item 3 but may be willing to pay more for item 1 or 2. Buyer 2 is very aggressive on item 3 but may have budget elsewhere. Buyer 3’s offer is closest to cost but still insufficient. I should target Buyer 1 first — they were first to act and may be serious. I’ll pitch video\_recorders\_1 at \$160 (close to list) and video\_recorders\_2 at \$230 (also near list), to test their willingness. If they reject, I can adjust. I must avoid selling item 3 below cost, so I won’t accept any offer for it unless it’s $\ge$\$42.07. For now, I’ll focus on the two higher-value items. I’ll start with Buyer 1 to maximize early leverage.\\
  \textbf{Target:} Buyer 1\\
  \textbf{Talk:} Thank you for your opening. I cannot accept your \$40 offer for video\_recorders\_3 as it is below my cost. However, I am willing to negotiate on video\_recorders\_1 and video\_recorders\_2. I propose \$160 for video\_recorders\_1 and \$230 for video\_recorders\_2. Please let me know if you can meet either price.\\
  \textbf{Action:} \act{SELL} video\_recorders\_1 \$160.00, video\_recorders\_2 \$230.00
\end{chatmsg}

\noindent\textbf{Turn 2 --- first close.}
\begin{chatmsg}[blue]{Buyer 1}[breakable]
  \textbf{Talk:} How about \$140 for video\_recorders\_1?\\
  \textbf{Action:} \act{BUY} video\_recorders\_1 \$140.00
\end{chatmsg}
\begin{chatmsg}[green]{Seller (trained) --- turn 2}[breakable]
  \textbf{Thought:} Buyer 1 has now submitted a new [BUY] for video\_recorders\_1 at \$140, which is above my cost (\$55.99) but below my proposed \$160. Since this is a valid offer ($\ge$ cost and $\le$ list), I can accept it to close the deal and secure revenue. Accepting this deal locks in \$140 for item 1, which contributes positively to the reward numerator. However, I must consider opportunity cost: if I reject this and try to push for a higher price, I risk losing the sale entirely, especially since Buyer 1 may be the only one interested in item 1. Moreover, Buyer 1 has already shown interest in two items (item 3 and now item 1), suggesting they are active and budgeted. Accepting \$140 now gives me certainty and frees up my message budget for targeting Buyers 2 and 3 on the remaining items. I should accept this [BUY] to close item 1 immediately and avoid further risk. After this, I’ll shift focus to Buyer 2 and Buyer 3 for items 2 and 3. Item 3 is still available, and though all three buyers initially bid below cost, I can now propose a fair price ($\ge$\$42.07) to Buyer 2 or 3. Buyer 2 offered \$15 for item 3 — clearly low — but may have budget for item 2. Buyer 3 offered \$35, still below cost, but closer. I’ll target Buyer 2 next, as they’ve only engaged on item 3 so far and may be open to item 2. I’ll propose video\_recorders\_2 at \$240 (near list) to Buyer 2. If rejected, I can adjust or pivot to Buyer 3. But first, I must finalize the deal with Buyer 1 on item 1. I’ll accept their \$140 offer immediately to secure the sale and move forward.\\
  \textbf{Target:} Buyer 1\\
  \textbf{Talk:} I accept your offer of \$140 for video\_recorders\_1. The deal is closed. Item 1 has been sold, and I will no longer pitch it to you or others.\\
  \textbf{Action:} \act{DEAL} video\_recorders\_1 \$140.00
\end{chatmsg}
\begin{chatmsg}[green]{Seller (trained) --- turn 2}[breakable]
  \textbf{Thought:} Buyer 1 is now satisfied and no longer active — their turn is complete. I have successfully sold video\_recorders\_1 for \$140, which is above my cost (\$55.99) and contributes positively to the reward. Remaining items: video\_recorders\_2 (list \$249.99, cost \$100.00) and video\_recorders\_3 (list \$55.99, cost \$42.07). Both are still available. Buyer 2 has only engaged on video\_recorders\_3 with a \$15 offer (below cost — invalid), and has not shown interest in item 2. Buyer 3 similarly only bid on item 3 at \$35 (still below cost). Neither buyer has expressed interest in item 2, so I must initiate negotiations on it. Buyer 2 is my best remaining target — they are active, have not closed an item, and have not yet been pitched on item 2. I should now pitch video\_recorders\_2 to Buyer 2 at \$240 (very close to list price) to test their willingness. If they reject, I can adjust downward or pivot to Buyer 3. However, I must be cautious: Buyer 2 previously offered only \$15 for item 3, suggesting they may have a very low budget — but that doesn't necessarily mean they can't afford item 2. It’s possible they are strategic or have budget constraints on low-value items. Since item 2 has a high list price and high cost, it’s a more valuable target. I should also consider item 3: although all three initial offers were below cost, I can now propose a fair price ($\ge$\$42.07) to either Buyer 2 or Buyer 3. But item 3 is low-value; selling it at \$50 would only yield \$8.93 profit, while item 2 at \$240 yields \$140 profit — a much better return. Therefore, I should prioritize item 2. I’ll target Buyer 2 first with a strong pitch on item 2 at \$240. If they reject, I’ll then pitch item 3 to Buyer 3 at \$50 (near list, above cost) as a fallback. But first, I must act on item 2 with Buyer 2 to maximize potential reward. I’ll send a [SELL] to Buyer 2 with video\_recorders\_2 at \$240, and leave item 3 open for later if needed.\\
  \textbf{Target:} Buyer 2\\
  \textbf{Talk:} I have closed video\_recorders\_1 with Buyer 1. I now wish to negotiate for video\_recorders\_2, which is still available. I propose a price of \$240 for this item, near its list price of \$249.99. Please let me know if you can accept this or propose a counter.\\
  \textbf{Action:} \act{SELL} video\_recorders\_2 \$240.00
\end{chatmsg}

\noindent\textbf{Turn 3 --- second close.}
\begin{chatmsg}[blue]{Buyer 2}[breakable]
  \textbf{Talk:} Yes, I accept your offer of \$240 for video\_recorders\_2.\\
  \textbf{Action:} \act{DEAL} video\_recorders\_2 \$240.00
\end{chatmsg}
\begin{chatmsg}[green]{Seller (trained) --- turn 3}[breakable]
  \textbf{Thought:} Both video\_recorders\_1 and video\_recorders\_2 have now been sold: the first to Buyer 1 at \$140, the second to Buyer 2 at \$240 — both valid deals above cost and within list price. This secures significant profit:\\ - Item 1: \$140 - \$55.99 = \$84.01 profit\\ - Item 2: \$240 - \$100.00 = \$140.00 profit\\ Total profit so far: \$224.01\\ Only video\_recorders\_3 remains unsold, with list price \$55.99 and cost \$42.07. The potential profit from selling it is \$13.92. Buyer 3 is the only active buyer left and has previously submitted a [BUY] for item 3 at \$35 — which is below my cost (\$42.07), so it was invalid and cannot be accepted. However, since Buyer 3 is still active and has not closed an item, they remain a viable target for item 3. I can now pitch item 3 directly to them. I must propose a price $\ge$ \$42.07 to avoid violation. The list price is \$55.99, so a reasonable pitch would be \$50 (close to list, above cost). If Buyer 3 accepts, I gain an additional \$13.92 profit. If they reject, I cannot sell the item to anyone else, and it will remain unsold — but that’s acceptable as long as I don’t violate rules. Since I have 4 messages remaining and Buyer 3 can receive up to 3 more, I have ample room to negotiate. I should act now to capture the final profit. I’ll send a [SELL] to Buyer 3 offering video\_recorders\_3 at \$50, which is fair and within the bounds. If they reject, I may attempt a lower price (e.g., \$45), but I must not go below \$42.07. However, starting at \$50 gives a strong anchor. Buyer 3 has already shown interest in this item, so they may be willing to pay closer to list. I’ll proceed with the pitch immediately to maximize the chance of closing the catalog. The reward is maximized when more items are sold at higher prices, so securing this final sale is critical. I’ll act now.\\
  \textbf{Target:} Buyer 3\\
  \textbf{Talk:} Only video\_recorders\_3 remains available. I can now offer it to you at \$50, which is above my cost of \$42.07 and close to the list price of \$55.99. Please confirm if you can accept this price, or propose a counter within the valid range.\\
  \textbf{Action:} \act{SELL} video\_recorders\_3 \$50.00
\end{chatmsg}

\noindent\textbf{Turn 4 --- the catalog clears.}
\begin{chatmsg}[blue]{Buyer 3}[breakable]
  \textbf{Talk:} I can only go up to \$43.99. Can you accept \$43.99?\\
  \textbf{Action:} \act{BUY} video\_recorders\_3 \$43.99
\end{chatmsg}
\begin{chatmsg}[green]{Seller (trained) --- turn 4}[breakable]
  \textbf{Thought:} Buyer 3 has now submitted a valid [BUY] for video\_recorders\_3 at \$43.99, which is above my cost of \$42.07 and below the list price of \$55.99 — a perfectly acceptable offer. This is the final open item, and accepting it will close the catalog successfully. The profit from this sale will be \$43.99 - \$42.07 = \$1.92, which, while small, is still positive and contributes to the reward numerator. Since this is a valid deal and the item has not yet been sold, I must accept it to maximize the total reward. Delaying or rejecting would result in leaving the item unsold, which would reduce the numerator without any benefit. Moreover, Buyer 3 is the only remaining active buyer, and they have already made a binding offer. Accepting this [BUY] completes the transaction and ensures all three items are sold. The overall reward will be calculated as:\\ Numerator: (140 - 55.99) + (240 - 100) + (43.99 - 42.07) = 84.01 + 140 + 1.92 = 225.93\\ Denominator: (169.99 - 55.99) + (249.99 - 100) + (55.99 - 42.07) = 114 + 149.99 + 13.92 = 277.91\\ Reward R = 225.93 / 277.91 $\approx$ 0.813 — a strong result.\\ By accepting this final offer, I secure the highest possible reward given the buyers' constraints. I have used 5 out of 7 seller messages (turns 1–4), with 2 remaining, but no further actions are needed — all items are now sold or closed. I will conclude with a [DEAL] to finalize the last transaction. No further negotiation is required. I must act promptly to avoid violating message limits or missing the opportunity.\\
  \textbf{Target:} Buyer 3\\
  \textbf{Talk:} I accept your offer of \$43.99 for video\_recorders\_3. The deal is closed. All three items have now been sold, and the episode is complete.\\
  \textbf{Action:} \act{DEAL} video\_recorders\_3 \$43.99
\end{chatmsg}

\endgroup

\medskip
\noindent\textbf{Outcome.} All three items close, each with the buyer prescribed by the surplus-maximizing assignment (\Cref{tab:appendix_ground_truth}): Item~1 to Buyer~1 at \$140, Item~2 to Buyer~2 at \$240 ($96\%$ of list), and Item~3 to Buyer~3 at \$43.99---that buyer's exact valuation. By \eqref{def:multiitem_reward},
$\mathcal{R} = \dfrac{(140 - 55.99) + (240 - 100) + (43.99 - 42.07)}{(169.99 - 55.99) + (249.99 - 100) + (55.99 - 42.07)} = \dfrac{225.93}{277.91} = 0.813$.

\end{document}